\documentclass[twoside,11pt,final]{article}

\usepackage{jmlr2e}

\usepackage{color,ifdraft,calc}
\usepackage[T1]{fontenc}

\usepackage{amsmath,amsfonts,amssymb}
\let\qedhere\relax %
\usepackage{tikz,pgfplots}
\usetikzlibrary{backgrounds,matrix,calc,decorations.pathreplacing}
\usepackage{wrapfig}

\usepackage[final]{hyperref}  %% wants to come last...

\makeatletter{}%!TEX root = article.tex

\newcommand{\Use}[1]{}

\newcommand{\toi}[1]{{#1\rightarrow\infty}}
\newcommand{\rep}[3][,]{#2_1#1\ldots#1#2_{#3}}
\newcommand{\repx}[3][,]{{\def\x{1}#2}#1\ldots#1{\def\x{{#3}}#2}}
\newcommand{\iseq}[2]{#2_1 #1 #2_{2} #1 \ldots}
\newcommand{\qqtext}[1]{\qquad\text{#1}\qquad}

\newcommand{\bl}{\left\langle}
\newcommand{\br}{\right\rangle}
\def\<#1>{\bl\;#1\;\br}

\newcommand{\1}{^{-1}}

\let\phi\varphi
\let\epsilon\varepsilon

\newcommand{\tand}{\text{ and }}
\newcommand{\tor}{\text{ or }}

\newcommand{\then}{\;\Longrightarrow\;}
\newcommand{\deff}{\;:\Longleftrightarrow\,}

\newcommand{\set}[2][\big]{#1\{\,#2\,#1\}}

\newcommand{\nolabel}[1]{\notag\relax}

\newcommand{\newcl}[1]{\expandafter\def\csname cl#1\endcsname{\mathcal{#1}}}
\newcommand{\newbf}[1]{\expandafter\def\csname bf#1\endcsname{\mathbf{#1}}}
\newcommand{\newds}[1]{\expandafter\def\csname ds#1\endcsname{\mathbb{#1}}}
\newcommand{\newfr}[1]{\expandafter\def\csname fr#1\endcsname{\mathfrak{#1}}}
\newcommand{\newmathop}[1]{\expandafter\DeclareMathOperator\csname #1\endcsname{#1}}
\newcommand{\newaxiom}[1]{\expandafter\def\csname #1\endcsname{\textnormal{\textsc{#1}}}}

\newds{N}
\newds{R}
\newds{Z}
\newds{P}
\newds{C}
\newds{B}
\newds{T}
\newds{D}
\newds{G}
\newds{Q}
\newds{F}

\newcl{A}
\newcl{B}
\newcl{C}
\newcl{D}
\newcl{E}
\newcl{H}
\newcl{L}
\newcl{P}
\newcl{F}
\newcl{G}
\newcl{K}
\newcl{M}
\newcl{I}
\newcl{L}
\newcl{N}
\newcl{M}
\newcl{Q}
\newcl{S}
\newcl{T}

\newbf{R}
\newbf{m}
\newbf{s}
\newbf{A}
\newbf{B}
\newbf{C}
\newbf{w}
\newbf{a}
\newbf{b}
\newbf{M}
\newbf{D}
\newbf{v}
\newbf{c}
\newbf{d}
\newbf{Q}
\newbf{q}
\newbf{y}
\newbf{g}
\newbf{z}
\newbf{i}

\newmathop{Tr}
\newmathop{diag}
\newmathop{Var}
\newmathop{Cov}
\newmathop{diam}
\newmathop{supp}
\newmathop{id}
\newmathop{Outer}
\newmathop{Inner}
\newmathop{ISA}
\newmathop{roots}
\newmathop{leaves}

\newmathop{Ground}

\newaxiom{Structured}
\newaxiom{ScaleInvariance}
\newaxiom{BaseMeasureClustering}
\newaxiom{DisjointAdditivity}
\newaxiom{BaseAdditivity}
\newaxiom{Continuity}

\let\Gr\dsG

\newfr{a}
\newfr{b}
\newfr{c}
\newfr{d}
\newfr{e}

\renewcommand{\a}{\fra}
\renewcommand{\b}{\frb}
\renewcommand{\c}{\frc}
\renewcommand{\d}{\frd}

\newcommand{\ca}{\alpha}
\newcommand{\cb}{\beta}
\newcommand{\cc}{\gamma}

\newcommand{\cl}{c}

\newcommand{\clAll}{(\clA,\clQ,\orthoG)}

\newcommand{\clAdyad}{\clA_{\text{Dyad}}}
\newcommand{\clAbox}{\clA_{\text{Box}}}

\newcommand{\clAconv}{\clA_{\text{Conv}}}

\newcommand{\clCdyad}{\clC_{\text{Dyad}}}

\newcommand{\clCconv}{\clC_{\text{Conv}}}

\newcommand{\clAdyadT}{\clA_{\text{Dyad}}^\tau}

\newcommand{\clAconvT}{\clA_{\text{Conv}}^\tau}

\newcommand{\circcirc}{\circ\kern-.6ex\circ}

\newcommand{\ortho}[1]{\perp_{#1}}
\newcommand{\orthoG}{\perp}

\newcommand{\northo}[1]{\mathrel{\mbox{$\circcirc_{#1}$}}}
\newcommand{\northoG}{\mathrel{\mbox{$\circcirc$}}}
\newcommand{\orthocup}[1][\orthoD]{\mathop{\overset{#1}\cup}}
\newcommand{\orthocupG}{\orthocup[\orthoG]}

\newcommand{\newortho}[2]{
  \expandafter\def\csname ortho#1\endcsname{\ortho{#2}}
  \expandafter\def\csname northo#1\endcsname{\northo{#2}}
  \expandafter\def\csname orthocup#1\endcsname{\orthocup[\ortho{#2}]}
}

\newortho{D}{\emptyset}
\newortho{L}{\ell}
\newortho{A}{\clA}
\newortho{T}{\tau}
\newortho{One}{1}
\newortho{Two}{2}

\newcommand{\indep}{\perp\kern-1.2ex\perp}
\newcommand{\Portho}[1]{\mathrel{\mbox{$\indep_{#1}$}}}
\let\nPortho\northo

\newcommand{\newPortho}[2]{
  \expandafter\def\csname Portho#1\endcsname{\Portho{#2}}
  \expandafter\def\csname nPortho#1\endcsname{\nPortho{#2}}
}

\newPortho{G}{}
\newPortho{P}{P}
\newPortho{Q}{Q}
\newPortho{T}{-\tau}
\newPortho{O}{\circ}
\newPortho{Pp}{P'}
\newPortho{One}{1}
\newPortho{Two}{2}

\newcommand{\R}[1]{\big|_{#1}}
\newcommand{\Rge}[1]{\R{\supset #1}}\newcommand{\Rg}[1]{\R{\supsetneqq #1}}
\newcommand{\Rle}[1]{\R{\subset #1}}\newcommand{\Rl}[1]{\R{\subsetneqq #1}}

\newcommand{\dotcup}{\mathop{\dot\cup}}

\newcommand{\ada}[1]{\noindent\emph{(#1).}} 
\newcommand{\qed}{\hfill\rule{7pt}{7pt}}

\newcommand{\refreshQED}{\gdef\qedhere{\gdef\qedhere{\relax}%
\ifmmode\tag*{\qed}\else\qed\fi}\let\myqedhere\qedhere}

\newenvironment{proofof}[1]{\refreshQED\noindent{\bf Proof of #1:}}{\qedhere\medskip}

\newenvironment{tikzpictureC}[1][]{\begin{center}\begin{tikzpicture}[#1]}
{\end{tikzpicture}\end{center}}

\tikzset{smooth/.style={out=0,in=180}}
\tikzset{cluster set/.style={very thick,blue}}
\newcommand{\drawset}[4][cluster set]{\draw[#1] (#3,#2)--(#4,#2) ;}
\newcommand{\drawsetCC}[4][cluster set]{\drawset[#1]{#2}{#3}{#4} \node[#1] at (#3,#2) {\tiny\relax[} node[#1] at (#4,#2) {\tiny]} ;}
\newcommand{\drawsetOC}[4][cluster set]{\drawset[#1]{#2}{#3}{#4} \node[#1] at (#3,#2) {\,\tiny(} node[#1] at (#4,#2) {\tiny]} ;}
\newcommand{\drawsetCO}[4][cluster set]{\drawset[#1]{#2}{#3}{#4} \node[#1] at (#3,#2) {\tiny\relax[} node[#1] at (#4,#2) {\tiny)\,} ;}
\newcommand{\drawsetOO}[4][cluster set]{\drawset[#1]{#2}{#3}{#4} \node[#1] at (#3,#2) {\tiny(} node[#1] at (#4,#2) {\tiny)} ;}
\newcommand{\drawlevel}[5][thick,anchor=north,blue]{\draw[thick,anchor=north,blue,#1]
	(#2,0) -- (#2,#4) -- (#3,#4) node[midway] {#5} -- (#3,0) ;}
\newcommand{\captionadd}[2][]{\caption[#1]{#1 #2}}

\newcommand{\theP}{\draw[smooth] (0,0) to ++(1.5,3.5) to ++(1,-1.5) to ++(1,1)
	to ++(1,-2) to ++(1,1) node[anchor=south east] {$P$} to ++(1,-2)  ;}

\usetikzlibrary {positioning} 
\tikzstyle {style_knoten}=[rectangle, fill=gray!25, draw, solid, sharp corners, text width=1.3 cm] 
\tikzstyle {style_arrow}=[->, line width=0.05cm]

\newtheorem{axiom}{Axiom}

\newcommand{\begriff}[1]{\textbf{\color{blue}#1}\index{#1}}

\newenvironment{description.defi}{\let\origitem\item\relax\def\item[##1]{\origitem[##1:]}
\begin{description}}{\end{description}}
\renewenvironment{description.defi}{\let\origitem\item\relax\renewcommand{\item}[1][]{\origitem \textbf{##1}:}
\begin{enumerate}}{\end{enumerate}}

\newlength{\myleftmargin}
\newlength{\fixboxwidth}
\jmlrheading{1}{2015}{1-48}{4/00}{10/00}{Philipp Thomann, Ingo Steinwart, and Nico Schmid}

\ShortHeadings{Towards an Axiomatic Approach to Hierarchical Clustering of Measures}{Thomann, Steinwart, and
Schmid}
\firstpageno{1}

\begin{document}

\title{Towards an Axiomatic Approach to\\ Hierarchical Clustering of Measures}

\author{\name Philipp Thomann \email philipp.thomann@mathematik.uni-stuttgart.de
       \\
       \name Ingo Steinwart \email ingo.steinwart@mathematik.uni-stuttgart.de\\
       \name Nico Schmid \email nico.schmid@mathematik.uni-stuttgart.de\\
       \addr Institute for Stochastics and Applications \\
       University of Stuttgart, Germany}

\editor{Vladimir Vapnik, Alexander Gammerman, and Vladimir Vovk}

\maketitle

\begin{abstract}%   <- trailing '%' for backward compatibility of .sty file
We propose %axioms for hierarchical clustering of probability measures.
some axioms for hierarchical clustering of probability measures
and investigate their ramifications.
The basic idea is to let the user stipulate the clusters for some elementary measures.
This is done without the need of any notion of metric, similarity or dissimilarity.
Our main results then show that for each suitable choice of user-defined clustering on elementary measures 
we obtain a unique notion of clustering on a large set of distributions
satisfying a set of additivity and continuity axioms.
We illustrate the developed theory by numerous examples including some with and some without a density.
\end{abstract}


%\clearpage
%\tableofcontents

\makeatletter{}%!TEX root = article.tex

\section{Introduction}

\makeatletter{}%!TEX root = article.tex

Clustering is one of the most basic tools to investigate unsupervised data:
finding groups in data.
%It is the continuation of the human need to understand and compartmentalize the world.
Its applications reach from categorization of news articles
over medical imaging to crime analysis.
For this reason, a wealth of algorithms have been proposed, among the best-known being:
$k$-means \citep{Macqueen-1967},
linkage \citep{Ward-1963,Sibson-1973,Defays-1977},
cluster tree \citep{Stuetzle-2003},
DBSCAN \citep{EsterKriegelSanderXu-1996},
spectral clustering \citep{DonathHoffmann-1973,Luxburg-2007},
and
expectation-maximization for generative models \citep{DempsterLairdRubin-1977}.
For more information and research on clustering we refer the reader to
%\citet{JardineSibson-1971}, \citet{Hartigan-1975}, and \citet{KaufmanRousseeuw-1990}
\citet{JardineSibson-1971,Hartigan-1975,KaufmanRousseeuw-1990,Mirkin-2005,GanMaWu-2007,Kogan-2007,BenDavid-2015,Menardi-2015}
and the references therein.

%\medskip

However, each \emph{ansatz} has its own implicit or explicit definition
of what clustering is.
Indeed for $k$-means it is a particular Voronoi partition,
for \citet[Section~11.13]{Hartigan-1975} it is the collection of connected components of a density level set,
and for generative models it is the decomposition of mixed measures into the parts.
 \citet{Stuetzle-2003} stipulates a grouping around the modes of a density, while
\citet{Chacon-2014} uses gradient-flows.
Thus, there is no universally accepted definition.
%The scientific method though separates the formulation of a problem from its solution:
%Setting the rules of the game first, avoids bias of the result.

A good notion of clustering certainly needs to address the inherent random variability in data.
This can be achieved by notions of clusterings for infinite sample regimes
or complete knowledge scenarios---as \citet{LuxburgBenDavid-2005} put it.
Such an approach has various advantages: one can talk about ground-truth,
can compare alternative clustering algorithms
(empirically, theoretically, or in a combination of both by using artificial data),
and can define and establish consistency and learning rates.
Defining clusters as the connected components of density level sets satisfies all of these
requirements.
Yet it seems to be slightly \emph{ad-hoc} and it will always be debatable,
whether thin bridges should connect components,
and whether close components should really be separated.
Similar concerns may be raised for other infinite sample notions of clusterings
such as \citet{Stuetzle-2003} and \citet{Chacon-2014}.

\medskip

In this work we address these and other issues by asking ourselves:
%by proposing some axioms for clustering of measures
%and investigating their ramifications.
\emph{What does the set of clustering functions look like?
What can defining properties---or axioms---of clustering functions be and what are their ramifications?
Given such defining properties, are there functions fulfilling these?
How many are there?
Can a fruitful theory be developed? And finally, for which distributions do we obtain a clustering and for which not?}

These questions have led us to an axiomatic approach.
The basic idea is to let the user stipulate the clusters for some elementary measures.
Here, his choice does not need to rely on a 
metric or another pointwise notion of similarity
though---only basic shapes for geometry and a separation relation
have to be specified.
Our main results then show that for each suitable choice we obtain a unique notion of clustering
satisfying a set of additivity and continuity axioms on a large set of measures.
These will be motivated in Section~\ref{sub.spirit.ansatz} and are defined in
Axioms~\ref{axiom.base.clustering}, \ref{axiom.clustering}, and~\ref{axiom.continuous.clustering}.
The major technical achievement of this work is Theorem~\ref{thm.uniqueness}:
it establishes criteria (c.f.\ Definition~\ref{defi.adapted}) to ensure a unique limit structure, which in turn makes it possible 
to define a unique additive and continuous clustering in Theorem \ref{thm.extension.axioms}.
Furthermore in Section~\ref{sub.density} we explain how this framework is linked to density based clustering,
and in the examples of Section~\ref{sub.examples.hausdorff} we investigate
the consequences in the setting of mixed Hausdorff dimensions.

\makeatletter{}%!TEX root = article.tex

\subsection{Related Work}

Some axioms for clustering have been proposed and investigated, but
to our knowledge, all approaches concern clustering of finite data.
\citet{JardineSibson-1971} were probably the first to consider axioms for hierarchical clusterings:
these are maps of sets of dissimilarity matrices to sets of e.g.\ ultrametric matrices.
Given such sets they obtain continuity and uniqueness of such a map using several axioms.
This setting was used by \citet{JanowitzWille-1995} to classify clusterings that are equivariant
for all monotone transformations of the values of the distance matrix.
Later, \citet{PuzichaHofmannBuhmann-1998} investigate axioms for cost functions
of data-partitionings and then obtain clustering functions as optimizers of such cost functions.
They consider as well a hierarchical version,
marking the last axiomatic treatment of that case until today.
More recently, \cite{Kleinberg-2003} put forward an impossibility result.
He gives three axioms and shows that any (non-hierarchical) clustering of distance matrices
can fulfill at most two of them.
\citet{ZadehBenDavid-2009} remedy the impossibility by restricting to $k$-partitions,
and they use minimum spanning trees to characterize different clustering functions.
A completely different setting is \citet{Meila-2005} where
an arsenal of axioms is given for distances of clustering partitions.
They characterize some distances (variation of information, classification error metric)
using different subsets of their axioms.
%
%However, to our knowledge, the work until now only considered finite samples of data points $D=(\rep xn)$.
%In order to get consistency results there has to be some notion
%of clustering of an unknown underlying target measure $P$ of \emph{infinite sample}
%or \emph{complete knowledge} as was declared interesting by \citet{LuxburgBenDavid-2005}

One of the reviewers brought clustering of discrete data to our attention.
As far as we understand,
consensus clustering \citep{Mirkin-1975,DayMcMorris-2003}
and additive clustering \citep{ShepardArabie-1979,Mirkin-1987}
are popular in social studies clustering communities.
What we call additive clustering in this work is something completely different though.
Still, application of our notions to clustering of discrete structures warrants further research.

%%% Not yet cited:
%\citet{BandeltDress-1992}
%\citet{ChenVanNess-1994}

%\citet{JanowitzWille-1995} should is using the setting as in \citet{JardineSibson-1971}
%\citet{Mirkin-1975}
%\citet{DayMcMorris-2003}
%\citet{ShepardArabie-1979}
%\citet{Mirkin-1987}

%\bigskip
%\noindent
%This work is organized in the following way.
%In the remainder of this introduction we give a birds view of the theory.
%In Section~\ref{section.additive.clustering} we introduce clustering bases
%and the axioms for additive clustering which yield
%uniqueness for a class of simple measures, c.f.\ Theorem~\ref{thm.clustering.of.simple.measures}.
%Section~\ref{section.continuous.clustering} introduces several
%new notions: isomonotone sequences, subadditivity, and $P$-adapted simple measures.
%These culminate in the main result Theorem~\ref{thm.uniqueness}: uniqueness of clustering for most measures.
%The section finishes with a comparison of $\cl(f\,d\mu)$
%to the forest of connected components of density level sets of $f$ (Theorem~\ref{thm.clustering.density.level.sets}).
%Finally Section~\ref{section.examples} uses examples to give
%a better understanding of how to use the theory developed in the sections before.
%It starts by giving instruments to construct nice clustering bases
%and several examples of densities in dimensions $1$ and $2$ are then examined.
%The section is completed by considering aggregations of heterogeneous Hausdorff dimensions.
%% : When can be a mixture of incomparable measures be clustered.
%The proofs begin in Section~\ref{sec:proofs}.

\makeatletter{}%!TEX root = article.tex

\subsection{Spirit of Our Ansatz}\label{sub.spirit.ansatz}

Let us now give a brief description of our approach. To this end assume for 
simplicity that we wish to find a hierarchical clustering for certain
distributions on $\dsR^d$. We denote the   set of such distributions by $\clP$.
Then a clustering is simply a map $c$ that assigns every $P\in \clP$ to
a   collection $c(P)$  of non-empty events. Since we are interested 
in hierarchical clustering, $c(P)$ will always be a forest, i.e.~we have 
\begin{equation}\label{axiom:forest}
  A,A'\in c(P) \then A\orthoG A' \tor A\subset A'\tor A\supset A'.
\end{equation}
Here $A\orthoG A'$ means \emph{sufficiently distinct}, i.e.~$A\cap A'=\emptyset$ or something stronger (cf.~Definition~\ref{defi.ortho}.
Following the idea that eventually one needs to store and process the clustering $c(P)$ on a computer, 
our first axiom assumes that  $c(P)$ is 
\emph{finite}. 
For a distribution with a continuous density
the level set forest, i.e.~the collection of all connected components of density level sets, will therefore \emph{not} 
be viewed as a clustering. For densities with finitely many modes, however, 
this level set forest consists of long chains interrupted 
by 
finitely many branchings. In this case,   the most relevant information for clustering is certainly represented
at the branchings and not in the intermediate chains.
Based on this observation, our second   clustering axiom postulates that $c(P)$ does not contain chains.
More precisely, if $s(F)$ denotes the forest that is obtained by replacing each chain in the forest $F$ by 
the maximal element of the chain, our \begriff{structured forest axiom} demands that 
\begin{equation}\label{axiom:structured}
   s(c(P)) = c(P)\, .
\end{equation}
To simplify notations we further extend the clustering to the cone defined by $\clP$
by setting 
\begin{equation}\label{axiom:scale-invariance}
   c(\alpha P) := c(P)
\end{equation}
for all $\alpha>0$ and $P\in \clP$. 
Equivalently we can view $\clP$ as a collection of finite non-trivial  measures and $c$ as a map on $\clP$ 
such that for $\alpha>0$ and $P\in \clP$ we have $\alpha P\in\clP$ and  $c(\alpha P) = c(P)$.
It is needless to say that this extended view on clusterings does not change the nature of a clustering.

Our next two axioms are based on the observation that there do not only exist distributions for which 
the ``right notion'' of a clustering is debatable but there are also distributions for which everybody would 
agree about the clustering. For example, if $P$ is the uniform distribution on a Euclidean ball $B$, then 
certainly everybody would set $c(P) = \{B\}$.
Clearly, other such examples are possible, too, and therefore we view the determination of distributions 
with such simple clusterings as a \emph{design decision}. More precisely, we assume that 
we have a collection $\clA$ of closed sets, called  
\begriff{base sets}
and a family $\clQ=\{Q_A\}_{A\in\clA}\subset\clP$ called \begriff{base measures}
with the property $A=\supp Q_A$ for all $A\in\clA$. Now, our \begriff{base measure axiom}
stipulates
\begin{equation}\label{axiom:trivial}
   c(Q_A) = \{A\}.
\end{equation}
It is not surprising that different choices of $\clA$, $\clQ$, and $\orthoG$ may lead to different 
clusterings. In particular we will see that larger classes $\clA$ usually result in more distributions
for which we can construct a clustering satisfying all our clustering axioms. On the other hand, taking a larger class
$\clA$ means that more agreement needs to be sought about the distributions having a trivial clustering \eqref{axiom:trivial}.
For this reason the choice of $\clA$ can be viewed as  a trade-off.

\begin{figure}[ptbh]
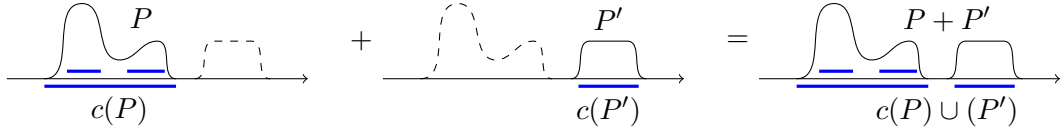

\begin{tikzpictureC}

\begin{scope}
\draw[->] (.5,0) -- (4.5,0) node[anchor=west] {};
%\draw[->] (1,0) -- (1,1) node[anchor=south] {$dP / dx$};
\draw[smooth] (1,0) to (1.5,1) to (2,.25) to (2.5,.5) to (2.75,0)
	;%
\draw[smooth,dashed] (3,0) to (3.25,.5) to (3.75,.5) to (4,0);
\drawset{0.1}{2.1}{2.6}
\drawset{0.1}{1.3}{1.75}
\drawset{-0.1}{1}{2.75}
%\drawset[cluster set,red]{-.1}{3.1}{3.9}

\node[anchor=west] at (2,.8) {$P$};
\node[anchor=north] at (2,-.1) {$c(P)$};
\end{scope}

\node[anchor=east] at (5.5,.5) {$+$};

\begin{scope}[xshift=5cm]
\draw[->] (.5,0) -- (4.5,0) node[anchor=west] {};
%\draw[->] (1,0) -- (1,1) node[anchor=south] {$dP / dx$};
\draw[smooth,dashed] (1,0) to (1.5,1) to (2,.25) to (2.5,.5) to (2.75,0)
	;%
\draw[smooth] (3,0) to (3.25,.5) to (3.75,.5) to (4,0);
\drawset[cluster set]{-.1}{3.1}{3.9}

\node[] at (3.5,.8) {$P'$};
\node[anchor=north] at (3.5,-.1) {$c(P')$};
\end{scope}

\node[anchor=east] at (10.5,.5) {$=$};

\begin{scope}[xshift=10cm]
\draw[->] (.5,0) -- (4.5,0) node[anchor=west] {};
%\draw[->] (1,0) -- (1,1) node[anchor=south] {$dP / dx$};
\draw[smooth] (1,0) to (1.5,1) to (2,.25) to (2.5,.5) to (2.75,0)
	;%
\draw[smooth] (3,0) to (3.25,.5) to (3.75,.5) to (4,0);
\drawset{0.1}{2.1}{2.6}
\drawset{0.1}{1.3}{1.75}
\drawset{-0.1}{1}{2.75}
\drawset{-.1}{3.1}{3.9}

\node[anchor=center] at (3,.8) {$P+P'$};
\node[anchor=north] at (3,-.1) {$c(P)\cup(P')$};
\end{scope}

%\drawboundingboxGrid
\end{tikzpictureC}
\vspace*{-3ex}
\caption{Example of disjoint additivity for two distributions having a density.}\label{fig.spirit.disjoint.additivity}
\end{figure}

Axiom \eqref{axiom:trivial} only describes distributions that have  a trivial clustering. However, there are also 
distributions for which everybody would agree on a non-trivial clustering. For example, if $P$ 
is the uniform distribution on two well separated Euclidean balls $B_1$ and $B_2$, then the ``natural''
clustering would be $c(P) = \{B_1,B_2\}$. Our \begriff{disjoint additivity axiom}
generalizes this observation by postulating 
\begin{equation}\label{axiom:disjoint-additivity}
  \supp P_1 \orthoG \supp P_2 \then c(P_1+P_2)=c(P_1) \cup c(P_2)\, .
\end{equation}
In other words, if $P$ consists of two spatially well separated sources $P_1$ and $P_2$, 
the clustering of $P$ should reflect this spatial separation, 
see also Figure~\ref{fig.spirit.disjoint.additivity}.
Moreover note this axiom formalizes
the vague term ``spatially well separated''  with the help of the relation $\orthoG$, which, like 
$\clA$ and  $\clQ$ is a design parameter that usually influences the nature of the  clustering.

\begin{figure}[bht]
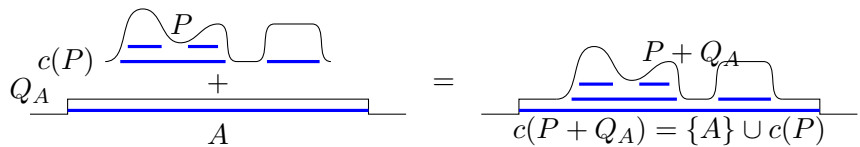

\begin{tikzpictureC}

\draw (0,0) -| (.5,.2) -- +(4,0) |- (5,0);
\drawset{.05}{.5}{4.5}
\node[anchor=south] at (0,0) {$Q_A$};
\node[anchor=north] at (2.5,0) {$A$};

\node at (2.5,.4) {$+$} ;

\begin{scope}[yshift=.7cm]
\draw[smooth] (1,0) to (1.5,.7) to (2,.25) to (2.5,.5) to (2.75,0)
	to (3,0) to (3.25,.5) to (3.75,.5) to (4,0);
\drawset{0.2}{2.1}{2.5}
\drawset{0.2}{1.3}{1.75}
\drawset{0}{1.2}{2.6}
\drawset{0}{3.15}{3.85}

\node[anchor=east] at (2.3,.5) {$P$};
\node[anchor=east] at (1,0) {$c(P)$};
\end{scope}

\node at (5.5,.4) {$=$} ;

\begin{scope}[xshift=6cm]
%%%% The sum:

\draw (0,0) -| (.5,.2) -- (1,.2)  (4,.2) -- +(.5,0) |- (5,0);
\drawset{0.05}{.5}{4.5}
%\draw (0,0) to[smooth] (1,.4)  +(3,0) to[smooth] (5,0);
%\node[anchor=south] at (0,0) {$Q_A$};
%\node[anchor=north] at (2.5,0) {$A$};

\begin{scope}[yshift=.2cm]
\draw[smooth] (1,0) to (1.5,.7) to (2,.25) to (2.5,.5) to (2.75,0)
	to (3,0) to (3.25,.5) to (3.75,.5) to (4,0);
\drawset{0.2}{2.1}{2.5}
\drawset{0.2}{1.3}{1.75}
\drawset{0}{1.2}{2.6}
\drawset{0}{3.15}{3.85}

\node[anchor=west] at (2,.6) {$P+Q_A$};
\end{scope}
\node[anchor=north] at (2.5,0.1) {$c(P+Q_A) = \{A\} \cup c(P)$};
\end{scope}

\end{tikzpictureC}
\vspace*{-3ex}
\caption{Example of base additivity.\label{fig.pour.noise}}
\end{figure}

The axioms \eqref{axiom:trivial} and \eqref{axiom:disjoint-additivity} only described the horizontal
behaviour of clusterings, i.e.~the depth of the clustering forest is not affected by 
\eqref{axiom:trivial} and \eqref{axiom:disjoint-additivity}. Our second additivity axiom
addresses this. To motivate it, assume that we have a $P\in \clP$ and a base measure $Q_A$,
e.g.~a uniform distribution on $A$,
such that $\supp P\subset A$. Then adding $Q_A$ to $P$ can be viewed as 
 pouring uniform noise  over $P$. Intuitively, this uniform noise should not affect the
internal and possibly delicate clustering of $P$ but only its roots, see also 
Figure~\ref{fig.pour.noise}.
Our \begriff{base additivity axiom} formalizes this intuition by stipulating
\begin{equation}\label{axiom:base-additivity}
   \supp P \subset A \then 
  c(P+Q_A) 
  = s\bigl(c(P) \cup \{A\}\bigr).
\end{equation}
Here the structure operation $s(\,\cdot\,)$ is applied on the right-hand side to avoid a conflict with the 
structured forest axiom \eqref{axiom:structured}.
Also note that it is this very axiom that directs our theory towards hierarchical clustering, since 
it controls the vertical growth of clusterings under a simple operation.

Any clustering satisfying  the axioms \eqref{axiom:forest} to \eqref{axiom:base-additivity}  
will be called an \begriff{additive clustering}.
Now the first, and rather simple part of our theory
shows that under some mild technical assumptions there is a \emph{unique} additive clustering on the 
set of  \begriff{simple measures on forests}
\[
   \clS(\clA) := \bigg\{\,
    \sum_{A\in F} \ca_A Q_A\mid \text{$F\subset\clA$ is a forest and $\ca_A>0$ for all $A\in F$}
  \,\bigg\}\, .
\]
Moreover, for $P\in \clS(\clA)$ there is a unique representation $P=  \sum_{A\in F} \ca_A Q_A$  and 
the additive  clustering is given by $c(P) = s(F)$.

Unfortunately, the set $\clS(\clA)$ of simple measures, 
on which the uniqueness holds, is usually rather small.
Consequently, additive clusterings on large collections $\clP$
are far from being uniquely 
determined. Intuitively, we may hope to address this issue if we additionally impose some sort of continuity on 
the clusterings, i.e.~an implication of the form 
\begin{equation}\label{axiom:continuity}
   P_n \to P \then c(P_n) \to c(P)\, .
\end{equation}
Indeed, having an implication of the form \eqref{axiom:continuity}, it is straightforward to 
show that the clustering is not only uniquely determined on $\clS(\clA)$ but actually on 
the ``closure'' of $\clS(\clA)$.
% 
%  
% 
% Clearly, to make this precise, we need to exactly describe what 
% the two indicated convergences on the left and the right mean.
To find a formalization of \eqref{axiom:continuity}, we first note that 
from a user perspective,  $c(P_n) \to c(P)$ usually describes a \emph{desired}
type of convergence. Following this idea,  $ P_n \to P$ then describes a \emph{sufficient} 
condition for \eqref{axiom:continuity} to hold.
In the remainder of this section we thus begin by presenting desirable properties 
$c(P_n) \to c(P)$ and resulting \emph{necessary} conditions on $P_n \to P$.

% Moreover, to make the implication in \eqref{axiom:continuity} as weak as possible, we 
% need to consider a strong notion for $ P_n \to P$ and a weak notion for $c(P_n) \to c(P)$.
% Finally, for both convergences there are some natural requirements that avoid contradictions.
% In the following we will try to describe these constraints in an informal way.

Let us begin by %considering $c(P_n) \to c(P)$. To this end, let us
assuming that 
 all $P_n$ are contained in $\clS(\clA)$ and let us further
  denote the corresponding forests in the unique representation of $P_n$ by  $F_n$.
Then we already know that $c(P_n) = s(F_n)$, so that the convergence on the right hand side of 
\eqref{axiom:continuity} becomes 
\begin{equation}\label{axiom:cont-h0}
   s(F_n) \to c(P)\, .
\end{equation}
Now, every  $s(F_n)$, as well as $c(P)$, is a finite forest, and so a minimal requirement for \eqref{axiom:cont-h0}
is that 
$s(F_n)$ and $c(P)$ are graph isomorphic, at least  for all sufficiently large $n$.
Moreover,  we certainly
 also need to demand that 
  every node  in $s(F_n)$    converges to the corresponding 
node in $c(P)$. To describe the latter postulation more formally, 
we fix graph isomorphisms $\zeta_n: s(F_1)\to s(F_n)$ and $\zeta:s(F_1)\to c(P)$.
Then our postulate reads as 
\begin{equation}\label{axiom-cont-h1}
   \zeta_n(A) \to \zeta(A),
\end{equation}
for all $A\in s(F_1)$.
Of course, there do exist various notions for describing convergence of sets, e.g.~in terms of the symmetric difference
or the Hausdorff metric, so at this stage we need to make a decision.
To motivate our choice, we first note that \eqref{axiom-cont-h1} actually 
contains two statements, namely, that $\zeta_n(A)$ converges for $n\to \infty$, and that its limit equals 
$\zeta(A)$. Now recall from various branches of mathematics that 
definitions of continuous extensions typically separate these two statements by considering approximating 
sequences that automatically converge.
Based on this observation, we decided to consider monotone sequences in  \eqref{axiom-cont-h1}, 
i.e.~we assume that $A\subset \zeta_1(A)\subset \zeta_2(A) \subset \dots$ for all 
$A\in s(F_1)$. Let us denote the resulting limit forest by $F_\infty$, i.e.
\[
   F_\infty := \biggl\{\, \bigcup_n \zeta_n(A) \mid A\in s(F_1) \,\biggr\}\, ,
\]
which is indeed a forest under some mild  assumptions on $\clA$ and $\orthoG$.
Moreover,  $\zeta_\infty:s(F_1)\to F_\infty$ defined by $\zeta_\infty(A) := \bigcup_n \zeta_n(A)$
becomes a graph isomorphism, and hence \eqref{axiom-cont-h1} reduces to 
\begin{equation}\label{axiom-cont-h2}
   \zeta_\infty(A) = \zeta(A) \qquad \qquad P\mbox{-almost surely for all  $A\in s(F_1)$.}
\end{equation}

Summing up our considerations so far, we have seen that our demands on $c(P_n) \to c(P)$ 
imply some conditions on the forests associated to the sequence $(P_n)$, namely
 $\zeta_n(A) \nearrow$ for all $A\in s(F_1)$. Without a formalization of 
$P_n \to P$, however, there is clearly no hope that this monotone convergence 
alone can guarantee \eqref{axiom:continuity}.
Like for \eqref{axiom-cont-h1}, there are again various ways for 
formalizing a convergence of $P_n \to P$. 
To motivate our decision, we first note that a weak continuity axiom is certainly more desirable 
since this would potentially lead  to more  instances of  clusterings. 
Furthermore, observe that 
 \eqref{axiom:continuity}
becomes weaker the stronger the notion of $P_n \to P$ is chosen. 
Now, if 
 $P_n$ and $P$ had densities $f_n$ and 
$f$, then one of the strongest notions of convergence would be $f_n\nearrow f$.
 In the absence of densities 
such a convergence can be expressed by $P_n\nearrow P$, i.e.~by 
\begin{equation*}%\label{axiom-cont-h3}
   P_n(B) \nearrow P(B) \qquad \qquad \mbox{for all measurable $B$.}
\end{equation*}
Combining these ideas we write $(P_n,F_n)\nearrow P$ iff  $P_n\nearrow P$ and there are graph isomorphisms
$\zeta_n: s(F_1)\to s(F_n)$ with $\zeta_n(A) \nearrow$ for all $A\in s(F_1)$.
Our formalization of \eqref{axiom:continuity} then  becomes
\begin{equation}\label{axiom:almost-cont}
   (P_n,F_n)\nearrow P \then 
   F_\infty = c(P)  \mbox{ in the sense of \eqref{axiom-cont-h2},}
\end{equation}
which should hold for all $P_n\in \clS(\clA)$ and their representing forests $F_n$.

While it seems tempting to stipulate such a continuity axiom it is unfortunately 
\emph{inconsistent}. To illustrate this inconsistency, consider, 
for example, the uniform distribution $P$ on $[0,1]$. Then $P$
can be approximated by 
the following two sequences
\begin{align*}
 &\begin{split}P_n^{(1)} & :=  \mathbf{1}_{[1/n,1-1/n]}P\\
 P_n^{(2)} & := \mathbf{1}_{[0,1/2-1/n]}P+\mathbf{1}_{[1/2,1]}P\end{split}
 &\qquad\qquad
 \begin{tikzpicture}[baseline=.5cm,xscale=1.5]
   \draw[dashed] (-.2,0) -| (0,1) -| (1,0) -- (1.2,0) ;
   \draw[blue] (.2,0) |- (.8,1) -- +(0,-1) ;
   \draw[dotted] (0,.5) -- (.2,.5) (.8,.5) -- (1,.5);
   \node[anchor=east] at (0,.5) {$P$} ;
   \node[] at (.5,.5) {$P_n^{(1)}$} ;
   \begin{scope}[xshift=2cm]
   \draw[dashed] (-.2,0) -| (0,1) -| (1,0) -- (1.2,0) ;
   \draw[blue] (0,0) |- (.3,1) |- (.5,0) |- (1,1) --  +(0,-1) ;
   \draw[dotted] (.3,.5) -- (.5,.5) ;
   \node[anchor=east] at (0,.5) {$P$} ;
   \node[] at (.8,.5) {$P_n^{(2)}$} ;
   \end{scope}
 \end{tikzpicture}
\end{align*}
By \eqref{axiom:almost-cont}
the first approximation would then lead to the clustering $c(P) = \{[0,1]\}$ while the second 
approximation would give $c(P) = \{[0,1/2), [1/2,1]\}$.

Interestingly, this example not only shows that \eqref{axiom:almost-cont} is
inconsistent but it also gives a hint how to resolve the inconsistency. Indeed
the first sequence seems to be ``adapted'' to the limiting distribution, whereas
the second sequence $(P_n^{(2)})$ is intuitively too complicated since its members have two clusters 
rather than the anticipated one cluster.
Therefore, the  idea to find a consistent alternative to \eqref{axiom:almost-cont} is to
restrict the left-hand side of \eqref{axiom:almost-cont} to ``adapted sequences'', so that 
our \begriff{continuity axiom} becomes 
\[
  (P_n,F_n)\nearrow P \text{ and $P_n$ is $P$-adapted for all $n$}
  \then 
   F_\infty = c(P)  \mbox{ in the sense of \eqref{axiom-cont-h2}.}
\]
In simple words, our main result 
then states that there exists exactly one such continuous clustering 
on the closure of $\clS(\clA)$. The main message of this paper thus is:

\noindent
\emph{Starting with very simple building blocks $\clQ  = (Q_A)_{A\in \clA}$
for which we (need to)  agree that they only have one trivial cluster $\{A\}$, we can 
construct a unique additive and continuous 
clustering on a  rich set of distributions.
Or, in other words, as soon as we have fixed $(\clA, \clQ)$ and a separation relation $\orthoG$,
there is no ambiguity left what a clustering is.}

What is left is to explore how the choice of the \begriff{clustering base}
$(\clA, \clQ, \orthoG)$ influences the resulting clustering. To this end, we first present
various clustering bases, which, e.g.~describe minimal thickness of clusters, their shape,
and how far clusters need to be apart from each other. For distributions 
having a Lebesgue density we then illustrate how different clustering bases lead 
to different clusterings. Finally, we show that our approach goes beyond density-based 
clusterings by considering %clustering bases and resulting clusterings for 
distributions consisting of several lower dimensional, overlapping parts.

\makeatletter{}%!TEX root = article.tex

%%%%%%%%%%%%%%%%%%%%%%%%%%%%%%%%%%%%%%%%%%%%%%%%%%%%%%%%%%%%%%%%
%%%%%%%%%%%%%%%%%%%%%%%%%%%%%%%%%%%%%%%%%%%%%%%%%%%%%%%%%%%%%%%%
%%%%%%%%%%%%%%%%        Additive Clustering       %%%%%%%%%%%%%%
%%%%%%%%%%%%%%%%%%%%%%%%%%%%%%%%%%%%%%%%%%%%%%q%%%%%%%%%%%%%%%%%%
%%%%%%%%%%%%%%%%%%%%%%%%%%%%%%%%%%%%%%%%%%%%%%%%%%%%%%%%%%%%%%%%

\section{Additive Clustering}
\label{section.additive.clustering}
% \proofssection{section.additive.clustering}

In this section we introduce base sets, separation relations, and simple measures,
as well as the corresponding axioms for clustering.
Finally, we show that there exists a unique additive clustering on the set of simple measures.

Throughout this work let $\Omega=(\Omega,\clT)$ be a Hausdorff space and
let $\clB\supset \sigma(\clT)$ be a $\sigma$-algebra that contains the Borel sets.
Furthermore we assume that $\clM=\clM_\Omega$ is the set of finite, non-zero, inner regular measures $P$
on $\Omega$.
Similarly  $\clM_\Omega^\infty$ denotes the set of non-zero measures on $\Omega$
if $\Omega$ is a Radon space and else of non-zero, inner regular measures on $\Omega$.
In this respect, recall that any Polish space---i.e.\ a completely metrizable separable space---is Radon.
In particular all open and closed subsets of $\dsR^d$ are Polish spaces and thus Radon.
For inner regular measures the support is well-defined 
and satisfies the usual properties, see Appendix~\ref{app.support} for details.
The set $\clM_\Omega$ forms a cone: for all $P,P'\in\clM_\Omega$ and all $\alpha>0$ we have
$P+P'\in\clM_\Omega$ and $\alpha P\in\clM_\Omega$.

%We will start by introducing separation relations and base measures.
%Next we investigate forests and simple measures since they
%will turn out to be the natural objects for additive clustering
%which is defined in the end of the current section.
%We end by showing existence and uniqueness of additive clustering.

\subsection{Base Sets, Base Measures, and Separation Relations}

Intuitively, any notion of a clustering should combine aspects of concentration and 
contiguousness.
What is a possible  core of this?
On one hand clustering should be \emph{local}
in the sense of disjoint additivity, which was presented in the introduction:
If a measure $P$ is understood on two parts of its support
and these parts are \emph{nicely separated}
then the clustering should be just a union of the two local ones.
Observe that in this case $\supp P$ is not connected!
On the other hand---in view of base clustering---%
base sets need to be impossible to partition into nicely separated components.
Therefore they ought to be \emph{nicely connected}.
Of course,  the meaning of \emph{nicely connected} and \emph{nicely separated} are interdependent,
and highly disputable.
For this reason, our notion of clustering assumes that both meanings are specified in front,
e.g.~by the user. Provided that both meanings satisfy certain technical criteria, we then
show, that there exists exactly one clustering.
To motivate how these technical criteria may look like, let us recall that 
for  all connected sets $A$ and all closed sets $\rep Bk$ we have
\begin{equation}\label{uniq-elem}
  A\subset \rep[\dotcup]Bk \then \exists!i\le k\colon A\subset B_i.
\end{equation}
The left hand side here contains the condition that the $\rep Bk$ are
pairwise disjoint, for which we already introduced the following notation:
\[
  B\orthoD B' \deff B\cap B' = \emptyset\, .
\]
In order to transfer the notion of connectedness to other relations
it is handy to generalize the notation $\rep[\dotcup]Bk$.
To this end, let $\orthoG$ be a relation on subsets of $\Omega$. 
Then
we denote the union $\rep[\cup]Bk$ of some $\rep Bk\subset \Omega$ by
\[
  \rep[\orthocupG]Bk \, ,%\deff \rep[\cup]Bk \tand B_i\orthoG B_j, \,\forall i\ne j.
\]
iff we have $B_i\orthoG B_j$ for all $i\neq j$.
% \begin{notation}
% Given a relation $\orthoG$ on subsets of $\Omega$, we denote the union $\rep[\cup]Bk$ of some $\rep Bk\subset \Omega$ by
% \[
%   \rep[\orthocupG]Bk \, ,%\deff \rep[\cup]Bk \tand B_i\orthoG B_j, \,\forall i\ne j.
% \]
% iff we have $B_i\orthoG B_j$ for all $i\neq j$.
% %We understand $A= B\orthocupG B'$, $A\supset B\orthocupG B'$, and $x\in B\orthocupG B'$
% %in an analogue fashion.
% %`$A\subset$' can be replaced by other set-theoretic relations as e.g.\ in:
% %\begin{align*}
% %  A= B\orthocupG B' &{}\deff A= B\cup B' \tand B\orthoG B', \\
% %  A\supset B\orthocupG B' &{}\deff A\supset B\cup B' \tand B\orthoG B', \\
% %  x\in B\orthocupG B' &{}\deff x\in B\cup B' \tand B\orthoG B'.
% %\end{align*}
% \end{notation}
% Of course, we can then rewrite \eqref{uniq-elem} using the notation $\rep[\orthocupD]Bk$. 
Now the key idea of the next definition is to generalize the notion of
connectivity and separation by replacing $\orthoD$ in \eqref{uniq-elem} by another suitable 
relation.

\begin{defi}\label{defi.ortho}
Let $\clA \subset \clB$ be a collection of closed, non-empty sets.
A symmetric relation $\orthoG$ defined on $\clB$
is called a $\clA$-\begriff{separation relation} iff the following holds:
\begin{description.defi}
\item[\begriff{Reflexivity}] For all $B\in\clB$: $B\orthoG B\then B=\emptyset$.

\item[\begriff{Monotonicity}] For all $A,A',B\in\clB$:
\[
  A\subset A' \tand A'\orthoG B
  ~ \then ~ A\orthoG B.
\]
\item[$\clA$-\begriff{Connectedness}] For all $A\in\clA$ and all closed $\rep Bk\in\clB$:
\[
  A \subset \rep[\orthocupG] Bk% \tand B\orthoG B'
  ~\then~ \exists i\le k\colon A\subset B_i.
\]
\end{description.defi}
Moreover, an $\clA$-separation relation $\orthoG$ is called \begriff{stable}, iff
for all $\iseq\subset A$ with $A_n\in\clA$, all $n \geq 1$, %\fix{Note that I changed the definition. I tried to catch all follow ups, but I do not give a guarantee.}
and all $B\in\clB$:
\begin{equation}\label{stable-implic}
  A_n\orthoG B \quad \text{for all } n\geq 1 ~ \then ~ \bigcup_{n\geq 1} A_n~\orthoG ~B.
\end{equation}
Finally, given a separation relation $\orthoG$ then we say that $B,B'$ are $\orthoG$-\begriff{separated}, if $B\orthoG B'$.
We write $B\northoG B'$ iff not $B\orthoG B'$,
and say in this case that $B,B'$ are $\orthoG$-\begriff{connected}.
\end{defi}

% \todo{I split the definition into two parts to be conssitent with forests in Lemma \ref{lemma.limit.is.forest}.
% It needs to be checked that I used the stable version at all places where this is necessary!
% Philipp: \emph{grep -v stable *.tex|grep -i --color clustering\ base}}

It is not hard to check that 
the disjointness relation $\orthoD$ is a stable $\clA$-separation relation,
whenever all $A\in\clA$ are topologically connected.
To present another example of a separation relation, 
we fix a metric $d$ on $\Omega$ and some $\tau>0$.  Moreover, 
for $B,B'\subset \Omega$ we write
\[
  B\orthoT B' \deff d(B, B') \geq \tau\, .
\]
In addition, recall that a $B\subset \Omega$ is $\tau$-connected, if, for all $x,x'\in B$, there exists 
$x_0,\dots,x_n\in B$ with $x_0 = x$, $x_n = x'$, and $d(x_{i-1},x_i)< \tau$ for all $i=1,\dots,n$.
Then it is easy to show that 
$\orthoT$ is an stable $\clA$-separation relation if all $A\in\clA$ are $\tau$-connected.
For more examples of separation relations we refer to 
Section~\ref{sub.examples.separations}.

It can be shown that $\orthoD$ is the weakest separation relation, i.e.~for every $\clA$-separation relation $\orthoG$ we have 
  $A\orthoG A' \then A\orthoD A'$  for all $A,A' \in \clA$. We refer to Lemma \ref{lemma.separation.base.2}, also showing 
  that $\orthoG$-unions are unique, i.e., for all $\rep Ak$ and all $\rep{A'}{k'}$ in $\clA$ we have
\[
  \rep[\orthocupG] Ak = \rep[\orthocupG]{A'}{k'} \then
  \{ \rep Ak \} = \{\rep{A'}{k'}\}.
\]  
% for all $\rep Ak\in\clA$ and all $\rep{A'}{k'}\in\clA$.
% The following two important consequences of the definition are proved in Lemma~\ref{lemma.separation.base.2}.
% 
% \begin{lemma}\label{lemma.separation.base}
% Let $\orthoG$ be an $\clA$-separation relation. Then:
% \begin{enumerate}
% \item $\orthoD$ is the weakest separation relation, i.e.:
%   $A\orthoG A' \then A\orthoD A'$  for all $A,A' \in \clA$.
% \item \begriff{Unique Decomposition}: For all $\rep Ak\in\clA$ and all $\rep{A'}{k'}\in\clA$ we have:
% \[
%   \rep[\orthocupG] Ak = \rep[\orthocupG]{A'}{k'} \then
%   \{ \rep Ak \} = \{\rep{A'}{k'}\}.
% \]
% \end{enumerate}
% \end{lemma}
Finally, the stability implication \eqref{stable-implic} is 
trivially satisfied
for \emph{finite} sequences $A_1\subset \dots\subset A_m$ in $\clA$, since in this case 
we have $A_1\cup \dots\cup A_m = A_m$.
For this reason stability will only become important when we will consider limits in Section~\ref{section.continuous.clustering}.

We can now describe the properties a clustering base should satisfy.

\begin{defi}\label{defi.base}
A (stable) \begriff{clustering base} is a triple $(\clA,\clQ,\orthoG)$ where
$\clA\subset \clB\setminus\{\emptyset\}$ is a class of non-empty sets,
$\orthoG$ is a (stable) $\clA$-separation relation, and
$\clQ=\{Q_A\}_{A\in\clA}\subset\clM$ is a family of probability measures on $\Omega$
with the following properties:
\begin{description.defi}
\item[\begriff{Flatness}] For all $A,A'\in\clA$ with $A\subset A'$ we either have
  $Q_{A'}(A) = 0$ or 
%   
%   $Q_A(\cdot) = \frac{Q_{A'}(\cdot \cap A)}{Q_{A'}(A)}$.
\[
 Q_A(\,\cdot\,) = \frac{Q_{A'}(\, \cdot\, \cap A)}{Q_{A'}(A)}\, .
\] 
\item[\begriff{Fittedness}] For all $A\in\clA$ we have $A=\supp Q_A$.
\end{description.defi}
We call a set $A$ a \begriff{base set} iff $A\in\clA$ and
a measure $\a\in\clM$ a \begriff{base measure on $A$} iff $A\in\clA$ and there is an $\ca>0$
with $\a=\ca Q_A$.
\end{defi}

\begin{wrapfigure}{R}{.3\textwidth}
\begin{tikzpicture}[yscale=1.5,xscale=.8,samples=50,
	/pgf/declare function={
    	S(\x,\m,\r) = (abs(\x-\m)<\r) * ((abs(\x-\m)-\r/2)^3*8-\r^2*6*abs(\x-\m)+\r^3*5)));
		S1(\x) = S(\x,0,1) / 10;
		S2(\x) = S(\x,.6,.4);
    }]
\draw[domain=-1:1] plot (\x, {S1(\x)});
\draw[domain=-1:1] plot (\x, {S(\x,.6,.4)});
\drawset{-.05}{-1}1 \drawset{-.1}{.2}1
\node at (-.5,.2) {$Q_{A'}$} ;
\node at (.8,.4) {$\alpha Q_{A}$} ;
\node at (1.5,.2) {$\then$} ;
\begin{scope}[xshift=3cm]
\draw[dotted,domain=-1:1] plot (\x, {S1(\x)});
\draw plot[domain=-1:1] (\x, {S1(\x) + S2(\x)});
\node at (0.3,.1) {$\alpha Q_A+Q_{A'}$} ;
\drawset{-.05}{-1}1 \drawset{-.1}{.2}1
\draw[dotted] (.2,-.1) -- (.2,{S1(.2) + S2(.2)}) ;
\end{scope}
%\draw[xshift=2cm,scale=2] (-1,0) -- (.3,0) -- plot[domain=0.3:.7,samples=200] (\x, {S(\x,0,1)/8 + S(\x,.5,.2)*4}) -- (.7,0) -- (1,0);
%\draw[xshift=5cm,scale=4] (0.5,0) -- (.6,0) -- plot[domain=.6:.95,samples=200] (\x, {S(\x,0,1)/4 + 7*S(\x,.8,.2)}) -- (1,0);
\end{tikzpicture}
\end{wrapfigure}
Let us motivate the two conditions of clustering bases.
Flatness concerns nesting of base sets:
Let $A\subset A'$ be base sets and consider the sum of their base measures $Q_A+Q_{A'}$.
If the clustering base is not flat, weird things can happen---see the  right.
The way we defined flatness excludes such cases
without taking densities into account. As a result we will be able to handle
% 
% 
% The sum shows two modes and part of the left mode still belongs to $A$!
% This should motivate the need for some notion of flatness.
% The above definition of flatness is easy enough to handle and powerful enough to handle
aggregations of measures of different Hausdorff-dimension  in Section~\ref{sub.examples.hausdorff}.
% Still we believe that there are more general versions of flatness,
% for which our results work.
% Then also mixtures of Hausdorff-measures are possible for basic measures,
% and the uniqueness $A\mapsto Q_A$ is abandoned.
Fittedness, on the other hand, establishes a link between the sets $A\in\clA$ and their associated base measures.

Probably, the easiest example of a clustering base 
has measures of the form 
\begin{equation}\label{ex-clust-base}
  Q_A(\,\cdot\,) =   \frac{\mu(\,\cdot\, \cap A)}{\mu(A)}
  = \frac{1_A \, d\mu}{\mu(A)}
  \, ,
\end{equation}
where $\mu$ is some reference measure independent of $Q_A$.
The next proposition shows that under mild technical assumptions such  
distributions do indeed provide a clustering base.

\begin{proposition}\label{prop.indicators}
Let $\mu\in\clM^\infty_\Omega$ and $\orthoG$ be a (stable) $\clA$-separation relation for some $\clA\subset \clK(\mu)$, where 
\[
  \clK(\mu) := \set{ C\in\clB\mid 0<\mu(C)<\infty \tand C=\supp~\mu(\cdot\cap C) }
\]
denotes the set of 
 \begriff{$\mu$-support sets}. 
% 
% Let $\orthoG$ be a (stable) $\clA$-separation relation for some $\clA\subset \clK(\mu)$ and 
We write $\clQ^{\mu,\clA} := \set{ Q_A \mid A\in\clA }$, where $Q_A$ is defined by 
\eqref{ex-clust-base}.
Then $(\clA,\clQ^{\mu,\clA},\orthoG)$ is a (stable) clustering base. % where $\clQ^{\mu,\clA} := \set{ Q_A \mid A\in\clA }$.
\end{proposition}

% Note that $\clK(\mu)$ consists of closed non-empty sets by its construction, and therefore
% all $\clA$ considered in Proposition \label{prop.indicators} also consists of such elements.
% In other words, 

Interestingly, distributions of the form \eqref{ex-clust-base} are not the only examples for 
clustering bases. For further details we refer to Section~\ref{sub.examples.hausdorff}, where
we discuss distributions % $Q_A$ 
supported by sets of different Hausdorff dimension.

% Fittedness on the other hand is not essential to our results.
% We investigated more general (flat) clustering bases, and the most important results can be
% achieved as well --at the expense of adding a vast machinery of notation and trivial
% assumptions-- without any additional insights into clustering.

\subsection{Forests, Structure, and Clustering}

As outlined in the introduction we are interested in hierarchical clusterings, i.e.~in clustering 
that map a finite measure to a forest of sets.
In this section we therefore recall some fundamental definitions and notations for such forests.

\begin{defi}
Let $\clA$ be a class of closed, non-empty sets, $\orthoG$ be an $\clA$-separation relation,
and $\clC$ be a class with
$\clA\subset\clC\subset \clB\setminus\{\emptyset\}$.
%s.t.\ $\orthoG$ fulfills $\clC$-connectedness
We say that a  non-empty 
$F\subset \clC$ is a 
$(\clC$-valued) $\orthoG$-\begriff{forest} iff
\[  A,A'\in F \then A\orthoG A' \tor A\subset A' \tor A'\subset A. \]
We denote the set of all such finite forests
 by $\clF_\clC$ and write $\clF := \clF_{\clB\setminus\{\emptyset\}}$.
\end{defi}

A finite $\orthoG$-forest $F\in\clF$ is partially ordered by the inclusion relation. 
The maximal elements  $\max F := \{ A \in F : \nexists A'\in F \text{ s.t. } A\subsetneqq A'\}$
are called \begriff{roots} and the minimal elements 
 $\min F := \{ A \in F : \nexists A'\in F \text{ s.t. } A'\subsetneqq A\}$
are called \begriff{leaves}. It is not hard to see that $A\orthoG A'$, whenever $A,A'\in F$ is a pair of roots or leaves.
Moreover, the \begriff{ground} of $F$ is 
$$
\Gr(F):=\bigcup\limits_{A \in F} A\, ,
$$ 
that is, $\Gr(F)$ 
equals the union over the roots of $F$.
Finally, $F$ is a \begriff{tree}, iff it has only a single root, or equivalently, $\Gr(F)\in F$,
and $F$ is a \begriff{chain} iff it has a single leaf, or equivalently, iff 
it is totally ordered. 

\begin{figure}[t]
\centering\begin{tikzpicture}[draw=blue,yscale=.4,xscale=.65]

%\draw[help lines] (-6,-3) grid (10,3);

\node at (-4.5,2) {$F$:};

\filldraw[fill=red!5] (0,0) ellipse (5 and 2);	%\node at (0,1.7) {$A_1$} ;
\filldraw[fill=red!10] (0,0) ellipse (4.5 and 1.5); %\node at (0,1.1) {$A_2$} ;
\filldraw[fill=red!15] (-3,1) to[out=0, in=180] (0,.75) to[out=0, in=180] (3,1) 
		to[out=0,in=90] (4,0) to[out=-90,in=0] (3,-1) to[out=180,in=0]
		(0,-.75) to[out=180,in=0] (-3,-1) to[out=180,in=-90] (-4,0) to[out=90,in=180] (-3,1);
		%%\node at (0,0) {$A_3$} ;

\filldraw[fill=blue!20] (-2.5,0) ellipse (1.2 and .9); %\node at (-2.5,-.75) {$A_{11}$} ;
\filldraw[fill=blue!25] (-2.5,0) ellipse (1 and .55); %\node at (-1.8,0) {$A_{12}$} ;
\filldraw[fill=blue!30] (-2.5,0) ellipse (.35); %\node at (-2.5,0) {$A_{13}$} ;

\filldraw[fill=blue!20] (-.7,.25) circle (.4) ; %\node at (-.7,.25) {$A_{21}$} ;

\filldraw[fill=blue!20] (2.2,0) ellipse (1.7 and .9); %\node at (1.8,-.5) {$A_{31}$} ;
\filldraw[fill=green!25] (1.3,0) circle (.4) ; %\node at (1.3,0) {$A_{311}$} ;
\filldraw[fill=green!25] (2.75,0) ellipse (.9 and .6) ; %\node at (3.2,0) {$A_{321}$} ;
\filldraw[fill=green!25] (2.3,0) circle (.4) ; %\node at (2.3,0) {$A_{322}$} ;

\filldraw[fill=black!5] (4.5,2.5) circle (1) ; %\node at (4,2) {$B_{1}$} ;
\filldraw[fill=black!10] (4.5,2.5) circle (.5) ; %\node at (4.5,2.5) {$B_{2}$} ;

%\begin{scope}[grow=up,level distance=.75cm,sibling distance=1cm,
%	edge from parent/.style={red,thick,draw}
%]
%\node[fill=red] at (7,-2) {$\mathbf{A_1}$}
%   child { node[fill=red] {$A_2$} child { node[fill=red] {$A_3$}
%       child { node[fill=blue!40] {$\mathbf{A_{31}}$}
%       		child { node[fill=green] {$\mathbf{A_{321}}$} child { node[fill=green] {$A_{322}$} } }
%       		child { node[fill=green] {$\mathbf{A_{311}}$} }
%       }
%       child { node[fill=blue!40] {$\mathbf{A_{21}}$} }
%       child { node[fill=blue!40] {$\mathbf{A_{11}}$} child { node[fill=blue!40] {$A_{12}$} child { node[fill=blue!40] {$A_{13}$} } }}
%   } }
%;
%
%\node at (8,-2) {$\mathbf{B_1}$}
%   child { node {$B_2$} } ;
%\end{scope}

\begin{scope}[xshift=12cm]
\node at (-4.5,2) {$s(F)$:};
\filldraw[fill=red!5] (0,0) ellipse (5 and 2);	%\node at (0,1.7) {$A_1$} ;
%\filldraw[fill=blue!10] (0,0) ellipse (4.5 and 1.5); \node at (0,1.1) {$A_2$} ;
%\filldraw[fill=blue!15] (-3,1) to[out=0, in=180] (0,.75) to[out=0, in=180] (3,1) 
%		to[out=0,in=90] (4,0) to[out=-90,in=0] (3,-1) to[out=180,in=0]
%		(0,-.75) to[out=180,in=0] (-3,-1) to[out=180,in=-90] (-4,0) to[out=90,in=180] (-3,1);
%		\node at (0,0) {$A_3$} ;

\filldraw[fill=blue!20] (-2.5,0) ellipse (1.2 and .9); %\node at (-2.5,0) {$A_{11}$} ;
%\filldraw[fill=blue!25] (-2.5,0) ellipse (1 and .55); \node at (-1.8,0) {$A_{12}$} ;
%\filldraw[fill=blue!30] (-2.5,0) ellipse (.35); \node at (-2.5,0) {$A_{13}$} ;

\filldraw[fill=blue!20] (-.7,.25) circle (.4) ; %\node at (-.7,.25) {$A_{21}$} ;

\filldraw[fill=blue!20] (2.2,0) ellipse (1.7 and .9); %\node at (1.8,-.5) {$A_{31}$} ;
\filldraw[fill=green!25] (1.3,0) circle (.4) ; %\node at (1.3,0) {$A_{311}$} ;
\filldraw[fill=green!25] (2.75,0) ellipse (.9 and .6) ; %\node at (2.75,0) {$A_{321}$} ;
%\filldraw[fill=blue!25] (2.3,0) circle (.4) ; \node at (2.3,0) {$A_{322}$} ;

\filldraw[fill=black!5] (4.5,2.5) circle (1) ; %\node at (4.5,2.5) {$B_{1}$} ;
%\filldraw[fill=blue!10] (4.5,2.5) circle (.5) ; \node at (4.5,2.5) {$B_{2}$} ;

%\begin{scope}[grow=up,level distance=.75cm,sibling distance=.8cm,
%	edge from parent/.style={red,thick,draw}
%]
%\node at (7,0) {${A_1}$}
%       child { node {${A_{31}}$}
%       		child { node {${A_{321}}$}  }
%       		child { node {${A_{311}}$} }
%       }
%       child { node {${A_{21}}$} }
%       child { node {${A_{11}}$} }
%;
%
%\node at (8,0) {${B_1}$}  ;
%\end{scope}
\end{scope}
\end{tikzpicture}

% \bigskip

%\begin{example}
%Consider the forest presented in Figure~\ref{fig.forest}.
%Then we have the following identities:
%Observe:

%The forests on the upper line are chains, on the lower line we first have a forest with two roots
%and then a tree.
%\end{example}
\caption[Illustration of a forest and its structure.]{\label{fig.forest}%
\small{Illustrations of a forest $F$ %=\{ A_1, A_2, A_3, B_1, B_2, A_{11}, \ldots \}$
and of its structure $s(F)$. 
%The elements of $F$ that are kept in $s(F)$ are printed in boldface in the top-right figure.
%Note that the roots and leaves are given  by $\max F{}= \{ A_1,B_1 \}$ and 
%$\min F {}= \{ A_{13},A_{21},A_{311},A_{322},B_2 \}$, respectively.
%Moreover, $A_1$ and $B_1$ are roots and therefore direct siblings without parent, while
%$A_{11},A_{21},A_{31}$ are direct siblings with direct parent $A_3$, 
%and $A_{311},A_{321}$ are direct siblings with direct parent $A_{31}$.
%Finally,   $F\Rg{A_{11}} {}= F\Rg{A_{21}} = F\Rg{A_{11}} = \{A_1,A_2,A_3\}$
%and $F\Rge{A_{11}} {}= \{A_1,A_2,A_3,A_{11}\}$
%are chains, $F\Rle{A_{31}} {}= \{A_{31},A_{311},A_{321},A_{322}\}$ is a forest, and 
%$F\Rl {A_{31}} {}= \{A_{311},A_{321},A_{322}\}$ is a tree.
}
%
%
% \begin{align*}
%   \max F&{}= \{ A_1,B_1 \}, &
%   \min F &{}= \{ A_{13},A_{21},A_{311},A_{322},B_2 \}, \\
%   s(F) &{}= \{A_1,B_1,A_{11},A_{21},A_{31},A_{311},A_{321} \}
% \end{align*}
% $A_1$ and $B_1$ are roots and therefore direct siblings without parent.
% $A_{11},A_{21},A_{31}$ are direct siblings with direct parent $A_3$
% and $A_{311},A_{321}$ are direct siblings with direct parent $A_{31}$.
% For the navigation we have:
% \begin{align*}
%   F\Rg{A_{11}} &{}= F\Rg{A_{21}} = F\Rg{A_{11}} = \{A_1,A_2,A_3\} &
%   F\Rge{A_{11}} &{}= \{A_1,A_2,A_3,A_{11}\} &&\text{are chains,}\\
%   F\Rle{A_{31}} &{}= \{A_{31},A_{311},A_{321},A_{322}\} \text{ is a forest,}&
%   F\Rl {A_{31}} &{}= \{A_{311},A_{321},A_{322}\} &&\text{is a tree}
% \end{align*}
%
%

}

\end{figure}
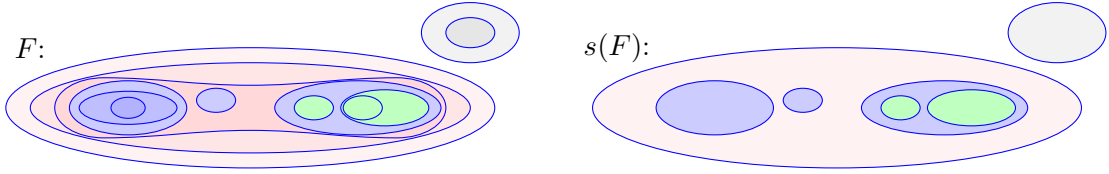

In addition to these standard notions, we often need a notation for 
describing certain sub-forests. Namely, for a finite forest $F\in \clF$ with $A\in F$ we write 
\begin{align*}
%  F\Rge{A} &{}:= \{ A'\in F\mid A'\supset A\},&
  F\Rg{A}  &{}:= \{ A'\in F\mid A'\supsetneqq A\}
%  \\
%  F\Rle{A} &{}:= \{ A'\in F\mid A'\subset A\},&
%  F\Rl{A}  &{}:= \{ A'\in F\mid A'\subsetneqq A\}.
\end{align*}
for the chain of strict ancestors  of $A$.
Analogously, we will use the notations
$F\Rge{A}$, $F\Rle{A}$, and $F\Rl{A}$  for the chain of ancestors of $A$ (including $A$),
the tree of descendants of $A$ (including $A$), and
the finite forest of strict descendants of $A$, respectively.
We refer to Figure~\ref{fig.forest} for an example of these notations.

% The following notation facilitates the navigating in forests:
% \begin{notation}\label{notation.restriction}
% Let $F\in \clF$ be a forest and $A\in F$ a node. Then we write:
% \begin{align*}
% %  F\Rge{A} &{}:= \{ A'\in F\mid A'\supset A\},&
%   F\Rg{A}  &{}:= \{ A'\in F\mid A'\supsetneqq A\}
% %  \\
% %  F\Rle{A} &{}:= \{ A'\in F\mid A'\subset A\},&
% %  F\Rl{A}  &{}:= \{ A'\in F\mid A'\subsetneqq A\}.
% \end{align*}
% for the chain of (strict) ancestors of $A$.
% $F\Rge{A}$, $F\Rle{A}$, and $F\Rl{A}$ are defined in a similar way
% and are the chain of parents of $A$ (including $A$),
% the tree of descendants of $A$ (including $A$), and
% the forest of descendants of $A$ (without $A$), resp.
% See Figure~\ref{fig.forest} for an example of these notions.
% \end{notation}
%Here $F\Rge A$ and $F\Rle{A}$ are always trees.
%$F\Rg{A}$ is empty iff $A\in\max F$ and a tree else,
%$F\Rl{A}$ is empty iff $A\in \min F$ and a forest or tree else.

\begin{defi}
Let $F$ be a finite forest. Then we call $A_1,A_2\in F$ \begriff{direct siblings} 
iff $A_1\ne A_2$ and they have the same strict 
ancestors, i.e. $F\Rg{A_1}=F\Rg{A_2}$.
In this case, any element 
 $$
 A' \in \min F\Rg{A_1} = \min F\Rg{A_2}
 $$ is called 
a \begriff{direct parent} of  $A_1$ and $A_2$. On the other hand for $A,A' \in F$ we denote $A'$ as a \begriff{direct child} of $A$, iff   
\[
 A' \in \max F\Rl{A}.
\] 
Moreover, 
the \begriff{structure} of   $F$ is defined by
\[
  s(F) := \Big\{ A\in F \,\big|\, \text{$A$ is a root or it has a direct sibling $A'\in F$}  \,\Big\}\,
\]
and $F$ is a \begriff{structured forest} iff $F=s(F)$. 
\end{defi}

For later use we note that  direct siblings $A_1,A_2$ in a $\orthoG$-forest  $F$ always satisfy  $A_1 \orthoG A_2$.
Moreover, 
the structure of a forest is obtained 
by pruning all sub-chains in $F$, see 
Figure~\ref{fig.forest}.
We further note  that $s(s(F))=s(F)$ for all forests, and if 
$F,F'$ are structured $\orthoG$-forests with $\Gr(F)\orthoG \Gr(F')$ then we have 
$s(F\cup F') = F\cup F'$.

%\addtoproofs{
%\begin{lemma}
%\begin{enumerate}\label{lemma.structure}
%\item For all $F\in\clF_\clA$: $s\big(s(F)\big)=s(F)$.
%\item Let $\rep Fk$, $k\ge1$, be trees with roots $\rep Ak$. If the roots are pairwise $\orthoG$-disjoint then
%  $F:=\bigcup_i F_i$ is a $\orthoG$-forest and
% \[
%   s(F) = s\left(\bigcup_i F_i\right) = \bigcup_i s(F_i)
% \]
%\item Let $A\in\clA$ and $F\in\clF_\clA$. If $\Gr F\subset A$ then $\{A\}\cup F\in\clF_A$ and
%  $F:=\bigcup_i F_i$ is a $\orthoG$-forest.
%\end{enumerate}
%\end{lemma}
%\begin{myproof}{lemma.structure}
%\ldots
%\end{myproof}
%}

Let us  now present our first set of axioms for 
(hierarchical) clustering.

\begin{axiom}[Clustering]\label{axiom.base.clustering}
Let $(\clA,\clQ,\orthoG)$ be a clustering base and $\clP\subset\clM_\Omega$ be a set of measures with $\clQ\subset\clP$.
A map $\cl \colon\clP \to \clF$ is called an 
\begriff{$\clA$-clustering} if it satisfies:
\begin{description.defi}
\item[\begriff{Structured}] For all $P\in\clP$ the forest $\cl(P)$ is   structured, i.e. $\cl(P) = s(\cl(P))$.
\item[\begriff{ScaleInvariance}]
For all $P\in\clP$ and $\ca>0$ we have $\ca P\in\clP$ and $\cl(\ca P) = \cl(P)$.
% \[  \, . \]
\item[\begriff{BaseMeasureClustering}]
For all $A\in\clA$ we have $\cl(Q_A) = \{ A\}$.
% \[ \cl(Q_A) = \{ A\}\, . \]
\end{description.defi}
\end{axiom}

Note that the scale invariance is solely for notational convenience.
Indeed, we could have defined clusterings for distributions only, in which case 
the scale invariance would have been obsolete. Moreover, assuming that 
a clustering produces structured forests essentially means that 
the clustering is only interested in the skeleton %\margincomment{should sceleton be further described?} 
of the cluster forest.
Finally, the axiom of  base measure clustering means that we have a set of elementary measures, 
namely the base measures, for which we already agreed upon that they can only be clustered in a trivial way.
In Section~\ref{section.examples} we will present a couple of examples of 
 $(\clA,\clQ,\orthoG)$ for which such an agreement is possible.
Finally note that these axioms guarantee that if $\cl \colon\clP \to \clF$
is a clustering and $\a$ is a base measure on $A$ then
$\a\in\clP$ and $\cl(\a)=\{A\}$.

% \begin{description.defi}
% \item[\begriff{Flatness}] For all $A,A'\in\clA$ with $A\subset A'$ we either have
%   $Q_{A'}(A) = 0$ or 
% %   
% %   $Q_A(\cdot) = \frac{Q_{A'}(\cdot \cap A)}{Q_{A'}(A)}$.
% \[
%  Q_A(\,\cdot\,) = \frac{Q_{A'}(\, \cdot\, \cap A)}{Q_{A'}(A)}\, .
% \] 
% \item[\begriff{Fittedness}] For all $A\in\clA$ we have $A=\supp Q_A$.
% \end{description.defi}

\subsection{Additive Clustering}

So far our axioms only determine the clusterings for base measures.
Therefore, the goal of this subsection is to
describe the behaviour of clusterings on certain combinations of measures.
Furthermore, we will show that the axioms describing this behaviour
are consistent and uniquely
determine a hierarchical clustering on a certain set of measures induced by $\clQ$.

Let us begin by introducing the axioms of additivity which we have already described and motivated 
in the introduction.

\begin{axiom}[Additive Clustering]\label{axiom.clustering}
Let $(\clA,\clQ,\orthoG)$ be a clustering base and $\clP\subset\clM_\Omega$ be a set of measures with 
$\clQ\subset\clP$.
A clustering  $\cl \colon\clP \to \clF$ is called  \begriff{additive} iff the following conditions are satisfied:
\begin{description.defi}
\item [\begriff{DisjointAdditivity}]
For all $\rep Pk\in\clP$ with mutually $\orthoG$-separated supports, i.e. $\supp P_i~\orthoG~ \supp P_j$
for all $i\ne j$, we have 
$\rep[+]Pk \in\clP$ and
\begin{align*}\Use{Ortho:DisjointSupport}
%  \supp P_i\orthoG \supp P_j \,\forall i\ne j
%  \then& \rep[+]Pn \in\clP \tand\\&{}
  c(\rep[+]Pn) = c(P_1)\cup\dots\cup c(P_n) \, .
\end{align*}

\item[\begriff{BaseAdditivity}]
For all $P\in\clP$ and all base measures $\a$ with $\supp P \subset \supp \a$  we have  $\a+P\in\clP$ and
\[
%  \supp P \subset A \then \a+P\in\clP \tand
  \cl(\a+P) = s\big( \{\supp\a\} \cup \cl(P) \big).
\]
\end{description.defi}
\end{axiom}

%\addtoproofs{
%\begin{remark}
%If $0\in\clP$ then $\supp 0=\emptyset$ and disjoint additivity imply that
%\[
%  c(0) = \bigcap_{P\in\clP} c(P)
%\]
%is the minimal element in the image of $c(\clP)$.
%Hence it is natural to stipulate $c(0)=\emptyset$ if at all.
%In this case base additivity already implies clustering of base measures.
%\end{remark}
%}

Our next goal is to show that there exist additive clusterings and that these are uniquely
on a set $\clS$ of measures that, in some sense, is spanned by $\clQ$.
The following definition introduces this set.

% 
% The definition of simple measure suits the correspondence to Lebesgue integration theory
% and will turn out to be the natural habitat of additive clusterings.

\begin{defi}
Let $(\clA,\clQ,\orthoG)$ be a clustering base and $F\in\clF_\clA$ be an $\clA$-valued finite $\orthoG$-forest.
A measure $Q$ is \begriff{simple on $F$} iff there exist % an $F\in\clF_\clA$ and  
base measures $\a_A$ on $A\in F$ such that
\begin{equation}
	Q = \sum_{A\in F} \a_A.
\label{eq: representation of simple measures}
\end{equation}
We denote the set of all simple measures with respect to $(\clA,\clQ,\orthoG)$ by $\clS:=\clS(\clA)$ .
\end{defi}

\begin{wrapfigure}{r}{0.3\linewidth}
\centering\begin{tikzpicture}[thick]
%\draw[help lines] (-4,-3) grid (4,0);

\begin{scope}[xscale=.25,yscale=.5]
\draw[] (-2,0) -- +(0,1) -| (8,0);
\draw[yshift=1cm] (-1,0) -- +(0,.5) -| (4,0) (5,0) -- +(0,1) -| (7,0);
\draw[yshift=1.5cm] (0,0) -- +(0,.2) -| (1,0) (2,0) -- +(0,.5) -| (3,0);

\draw[thin,->] (-2.5,0) -- +(11,0) node[anchor=west]{$\Omega$} ;
\draw[thin,->] (-2.2,0) -- +(0,2) ; %node[anchor=east]{$dQ/dx$} ;

\begin{scope}[cluster set,yshift=.5cm]
\drawset{-.8}01 \drawset{-.8}23
\drawset{-1}{-1}4 \drawset{-1}57
\drawset{-1.2}{-2}8
\end{scope}
\end{scope}

\end{tikzpicture}
\caption{\label{fig.simple.measure}%
Simple measure.
%A simple measure is the sum of base measures
%organized in a forest.
}
\end{wrapfigure}
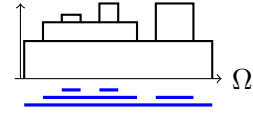

Figure~\ref{fig.simple.measure} provides an example of a simple measure.
The next lemma shows that the representation \ref{eq: representation of simple measures} of simple measures is actually unique.

\begin{lemma}\label{lemma.simple.measures.unique}
Let $(\clA,\clQ,\orthoG)$ be a clustering base and $Q\in\clS(\clA)$. Then there exists exactly one
$F_Q\in \clF_\clA$ such that $Q$ is simple on $F_Q$. Moreover, the representing 
base measures $\a_A$ in \eqref{eq: representation of simple measures} are also unique and 
we have $\supp Q = \Gr F_Q$.
% representation 
% \eqref{eq: representation of simple measures} is unique, i.e.~there exists exactly one 
% 
% Let $Q\in\clS$ on a forest $F$.
% Then 
% Furthermore the representation $Q = \sum_{A\in F} \alpha_AQ_A$ is unique.
\end{lemma}

Using  Lemma \ref{lemma.simple.measures.unique} we can now define 
certain restrictions of simple measures $Q\in \clS(\clA)$ 
 with representation  \eqref{eq: representation of simple measures}.
 Namely, 
any  subset $F'\subset F$ gives  a measure 
\[ 
Q\R{F'} :=\sum_{A\in F'} \a_A\, . 
\]
We write 
$Q\Rge{A} := Q\R{F\Rge{A}}$ and similarly  $Q\Rg{A},Q\Rle{A},Q\Rl{A}$.

% \begin{notation}
% Let $Q=\sum_{A\in F} \a_A$ be a simple measure on a forest $F$.
% For all subsets $F'\subset F$ we write:
% \[ Q\R{F'} :=\sum_{A\in F'} \a_A \]
% and for short $Q\Rge{A} := Q\R{F\Rge{A}}$ and similarily $Q\Rg{A},Q\Rle{A},Q\Rl{A}$.
% %\begin{align*}
% %  Q\Rge{A} &{}:= Q\R{F\Rge{A}},&
% %  Q\Rg{A} &{}:= Q\R{F\Rg{A}}, &
% %  Q\Rle{A} &{}:= Q\R{F\Rle{A}},&
% %  Q\Rl{A} &{}:= Q\R{F\Rl{A}}
% %\end{align*}
% \end{notation}

With the help of Lemma \ref{lemma.simple.measures.unique} it is now easy to explain how a possible additive 
clustering could look like on $\clS(\clA)$. Indeed, for a $Q\in \clS(\clA)$, Lemma 
\ref{lemma.simple.measures.unique} provides a unique finite forest $F_Q\in \clF_\clA$ that represents $Q$
and therefore the structure $s(F_Q)$ is a natural candidate for a clustering of $Q$.
The next theorem shows that this idea indeed leads to an additive clustering and that 
every additive clustering on $\clS(\clA)$ retrieves the structure of the underlying forest of a simple measure.

\begin{thm}\label{thm.clustering.of.simple.measures}
Let $(\clA,\clQ,\orthoG)$ be a clustering base and $\clS(\clA)$ the set of simple measures. 
Then we can define an additive $\clA$-clustering
$\cl : \clS(\clA) \rightarrow \clF_\clA$   by
\begin{equation}\label{def-uniq-add-clust}
\cl(Q) := s(F_Q)\, , \qquad \qquad Q\in \clS(\clA) \, .
\end{equation}
% is an additive $\clA$-clustering. 
Moreover,  every additive $\clA$-clustering $\cl: \clP \rightarrow \clF$
satisfies both $\clS(\clA) \subset \clP$ and \eqref{def-uniq-add-clust}.
% is an additive $\clA$-clustering, 
% then $\clS \subset \clP$ and $\cl(Q) = s(F_Q)$ f.a. $Q \in \clQ$.
%Let $(\clA,\clQ,\orthoG)$ be a clustering baseand $\cl: \clP \rightarrow \clF_\clA$ an additive clustering.
%If $Q$ is a simple measure on a forest $F\in\clF_\clA$ then $Q\in\clP$ and:
%\[
  %\cl(Q) = s(F).
%\]
%On the other hand the mapping $\clS\to\clF$ with $Q\mapsto s(F)$ is an additive clustering.
\end{thm}
%So if all simple measures are simple measures then
% So we always have $\clS\subset\clP$ and there exists a clustering function
% $\cl(\cdot)$ at least on $\clS$ and there it is even unique.

% To give a more formal motivation we first introduce the following

\makeatletter{}
%%%%%%%%%%%%%%%%%%%%%%%%%%%%%%%%%%%%%%%%%%%%%%%%%%%%%%%%%%%%%%%%
%%%%%%%%%%%%%%%%%%%%%%%%%%%%%%%%%%%%%%%%%%%%%%%%%%%%%%%%%%%%%%%%
%%%%%%%%%%%%%%%        Continuous Clustering       %%%%%%%%%%%%%
%%%%%%%%%%%%%%%%%%%%%%%%%%%%%%%%%%%%%%%%%%%%%%%%%%%%%%%%%%%%%%%%
%%%%%%%%%%%%%%%%%%%%%%%%%%%%%%%%%%%%%%%%%%%%%%%%%%%%%%%%%%%%%%%%

%\clearpage
\section{Continuous Clustering}
\label{section.continuous.clustering}
% \proofssection{section.continuous.clustering}

As described in the introduction, 
we typically need, besides additivity, also some notion of continuity for clusterings.
The goal of this section is to introduce such a notion and to show that, similarly to  
Theorem \ref{thm.clustering.of.simple.measures}, this continuity uniquely defines a clustering 
on a suitably defined extension of $\clS(\clA)$.

To this end, we first introduce a notion of monotone convergence for sequences of simple 
measures that does not change the graph structure of the corresponding clusterings given by Theorem 
\ref{thm.clustering.of.simple.measures}.
We then discuss a richness property of the clustering base, which essentially ensures that 
we can approximate the non-disjoint union of two base sets by another base set. 
In the next step we  describe monotone sequences of simple measures that are in some sense 
adapted to the limiting distribution. In the final part of this section we then 
axiomatically describe continuous clusterings and show both their existence and their uniqueness.

\makeatletter{}
\subsection{Isomonotone Limits}
\label{subsec.isomonotone.limits}

The goal of this section is to introduce a notion of monotone convergence for simple measures 
that preserves the graph structure of the corresponding clusterings.

Our first step in this direction is done in the following definition that introduces a sort 
of monotonicity for set-valued isomorphic forests.

\begin{defi}\label{defi.forests.compare}
Let $F,F' \in \clF$ be two finite forests. Then 
$F$ and $F'$ are \begriff{isomorphic}, denoted by $F\cong F'$,
iff there is a bijection $\zeta\colon F\to F'$ such that for all $A,A'\in F$ we have:
\begin{equation}
	A\subset A' \iff \zeta(A) \subset \zeta(A').
\label{eq: forest isomorph}
\end{equation}
Moreover, we write   $F\le F'$ iff $F\cong F'$ and there is a map 
$\zeta : F \rightarrow F'$ satisfying \ref{eq: forest isomorph} and
\begin{equation}\label{def-frm-eq}
A\subset\zeta(A) \, , \qquad \qquad  A\in F.
\end{equation}
In this case, the map $\zeta$, which is uniquely determined by \eqref{eq: forest isomorph}, \eqref{def-frm-eq} and the fact that $F$ and $F'$ are finite, is called the
\begriff{forest relating map} (FRM) between $F$ and $F'$.
\end{defi}
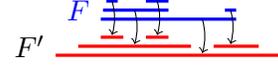
\begin{wrapfigure}{r}{.32\textwidth}
\centering
\begin{tikzpicture}[very thick,xscale=.3,yscale=.6]
\begin{scope}[blue]
\drawset0{.25}{.75} \drawset023
\drawset{-.2}{.1}{3} \drawset{-.2}{5.5}{6}
\drawset{-.4}06
\node[anchor=east] at (0,-.2) {$F$};
\end{scope}

%\draw [decorate,decoration={brace,amplitude=3pt},yshift=-.4cm]
%	 (-3,-.1) -- +(0,.6) node[midway,anchor=east] {$F\ $};

\begin{scope}[cluster set/.append style=red]
\drawset{-.8}01 \drawset{-.8}23
\drawset{-1}{-1}4 \drawset{-1}57
\drawset{-1.2}{-2}8
\node[anchor=east] at (-2,-1) {$F'$};
\end{scope}

%\draw [decorate,decoration={brace,amplitude=3pt},yshift=-1.25cm]
%	 (-3,-.1) -- +(0,.6) node[midway,anchor=east] {$F'\ $};

\begin{scope}[thin]
\foreach \p in { (.5,0), (2.5,0), (1.5,-.2), (5.75,-.2), (4.5,-.4) }
\draw[->,bend left] \p to +(0,-.75) ;
\end{scope}

%\draw[thin,->] (-2.5,-1.5) -- +(11,0) ;% node[anchor=west]{$\dsR$} ;

\end{tikzpicture}
\caption{\label{fig.forest.compare}%
%Two interval-valued forests  with
$F\le F'$ and the arrows indicate $\zeta$. }
\end{wrapfigure}
Forests can be viewed as directed acyclic graphs:
There is an edge between $A$ and $A'$ in $F$ iff $A\subset A'$ and no other node is in between.
Then $F\cong F'$ holds iff  $F$ and $F'$ are isomorphic as directed graphs.
From this it becomes clear that $\cong$ is an equivalence relation.
Moreover, the relation $F\le F'$  means that each node $A$ of $F$ can be graph isomorphically mapped
to a node of $F'$ that contains $A$, see 
Figure~\ref{fig.forest.compare} for an illustration.
Note that   $\le$ is a partial order on $\clF$ and in particular it is transitive.
Consequently, if we have finite forests  $F_1\le\dots\le F_k$ then $F_1\le F_k$ and there is 
an FRM $\zeta_k\colon F_1\to F_k$. This observation is used in the following definition,
which introduces monotone sequences of forests and their limit.

% With this we are able to define what monotone convergence means for forests.
\begin{defi}\label{defi.forests.limit}
A \begriff{isomonotone sequence of forests} is defined as a sequence of finite forests $(F_n)_n\subset \clF$ such that
%We say that $(F_n)\subset \clF$ is an \begriff{isomonotone sequence of forests} iff 
$s(F_n)\le s(F_{n+1})$ for all $n\geq 1$.
If this is the case, we define the limit by
\begin{align*}
  F_\infty := \lim\limits_{n\rightarrow \infty} s(F_n) &{}:= \bigg\{\, \bigcup\limits_{n\geq1} \zeta_n(A) \mid A \in s(F_1) \,\bigg\},
%  \in\clF_{\bar\clA}
\end{align*}
where $\zeta_n\colon s(F_1)\to s(F_n)$ is the FRM obtained from $s(F_1)\le  s(F_n)$.
%Let $F_1,F_2,\ldots\in \clF_\clA$ be forests.
\end{defi}

It is easy to see that 
in general, the limit forest $F_\infty$ of an isomonotone sequence of $\clA$-valued forests
is not $\clA$-valued. To describe the values of $F_\infty$ we define the 
 \begriff{monotone closure} of an $\clA \subset \clB$  by\label{eq.closure.of.a}
\[
  \bar\clA := \biggl\{\, \bigcup_{n\geq 1} A_n \mid A_n\in\clA \tand \iseq\subset A \,\biggr\}.
\]
The next lemma states some useful properties of $\bar\clA$ and $F_\infty$.
\begin{lemma}\label{lemma.limit.is.forest}
Let $\orthoG$ be an  $\clA$-separation relation. Then 
 $\orthoG$ is  actually an $\bar\clA$-separation relation.
Moreover, if $\orthoG$ is stable and 
$(F_n)\subset \clF_\clA$ is an isomonotone sequence then  $F_\infty:=\lim_n s(F_n)$
is an $\bar\clA$-valued $\orthoG$-forest and we have 
 $s(F_n)\le F_\infty$ for all $n\geq 1$.
\end{lemma}

Unlike forests, it is straightforward to compare two measures $Q_1$ and $Q_2$ on $\clB$. Indeed, 
we say that $Q_2$ \begriff{majorizes} $Q_1$, in symbols $Q_1\le Q_2$, 
iff
\[
    Q_1(B) \le Q_2(B), \qquad \qquad \text{ for all  } B\in\clB.
\]
For $(Q_n)\subset \clM$ and $P\in \clM$, we similarly speak of  \begriff{monotone convergence} $Q_n\uparrow P$
iff $Q_1\le Q_2\le\cdots\le P$ and 
\[
\lim\limits_{n\rightarrow \infty} Q_n(B)=P(B), \qquad \qquad \text{ for all  } B\in\clB.
\]
Clearly,   $Q\le Q'$ implies $\supp Q\subset \supp Q'$ and it is easy to show, %cf.~Lemma~\ref{lemma.monotone.convergence.1},
that $Q_n \uparrow P$ implies
\begin{align*}
P\big(\supp P\setminus \bigcup\limits_{n\geq 1} \supp Q_n\big) = 0.
\end{align*}
We will use such arguments  throughout this section. For example,  if $\a,\a'$ are base measures on $A,A'$
with $\a\le \a'$ then $A\subset A'$.
% Observe that for any simple measure $Q$ on a forest $F$ we have $\supp Q=\Gr F$.
%
%
With the help of these preparations we can now define isomonotone convergence of simple measures.

\begin{defi}
\label{defi.monotone.isostructural.convergence}
Let $(\clA,\clQ,\orthoG)$ be a clustering base and $(Q_n)\subset \clS(\clA)$ be a sequence 
of simple measures on finite forests $(F_n)\subset \clF_\clA$. Then 
\begriff{isomonotone convergence},
denoted by $(Q_n,F_n)\uparrow P$, means that both
% $Q_n$ is a simple measure on $F_n\in\clF_\clA$ for all $n$ and
\[
  Q_n\uparrow P \quad \tand \quad  s(F_1)\le s(F_2)\le \ldots \, .
\]
In addition, $\bar\clS:=\bar\clS(\clA)$ denotes 
 the set of all isomonotone limits, i.e.
\[
  \bar\clS(\clA) = \set{ P\in\clM\mid (Q_n,F_n)\uparrow P \text{ for some } (Q_n)\subset \clS(\clA) \text{ on } (F_n)\subset\clF_{\clA}}.
\]
\end{defi}

For a measure $P\in \bar\clS(\clA)$ it is probably tempting to define its clustering
by $c(P) := \lim_n s(F_n)$, where $(Q_n,F_n)\uparrow P$ is some isomonotone sequence.
Unfortunately, such an approach does not yield a well-defined clustering as we have discussed in the introduction.
% For example, the uniform distribution $P$ on $[0,1]$ can be isomonotonically approximated by 
% both, the sequence $(P_n)$ and the sequence $(Q_n)$ given by $Q_n := \mathbf{1}_{[0,1/2-1/n]}+\mathbf{1}_{[1/2,1]}P$.
% However, the first approximation would lead to the clustering $c(P) = \{[0,1]\}$ while the second 
% approximation would give $c(P) = \{[0,1/2), [1/2,1]\}$.
For this reason, we need to develop some tools that help us to distinguish between ``good'' and ``bad'' 
isomonotone approximations. This is the goal of the following two subsections.

\makeatletter{}\subsection{Kinship and Subadditivity}

In this subsection we present and discuss a technical assumption on
a clustering base that will make it possible to obtain unique continuous clusterings.

Let us begin by introducing a notation that will be frequently used in the following. 
To this end, we fix a clustering base $(\clA,\clQ,\orthoG)$ and a $P\in \clM$. For 
$B\in\clB$ we then define % consider the following sets of base measures:
\begin{align*}
  \clQ_P(B) &{}:= \{ \ca Q_A\mid \ca >0, A\in\clA, B\subset A, \ca Q_A\le P \}\,  ,
\end{align*}
i.e. $\clQ_P(B)$ denotes the set of all basic measures below $P$ whose support contains $B$.
% In addition, we write $\clQ_P{}:=\clQ_P(\emptyset)$. 
Now, our first definition describes events that can be combined in a base set:

\begin{wrapfigure}{l}{0.4\textwidth}
\centering
\begin{tikzpicture}[scale=.5]
\begin{scope}%[xshift=§7cm]
\theP
\draw[smooth] (6.5,0) -- ++(.5,0) to ++(1,2) to ++(1,0) to ++(1,-2) ;

\drawset{.1}1{2.4} \node[anchor=south] at (1.7,.1) {$A_1$};
\drawset{.1}3{3.8} \node[anchor=south] at (3.4,.1) {$A_2$};
\drawlevel[dashed,red,align=center]{.8}{4}{1.2}{$\b$}

\drawset{.1}{8.2}9 \node[anchor=south] at (8.6,.1) {$A_3$} ;
\node[anchor=west,align=left] at (5.5,3) {$A_1\sim_P A_2$,\\ $A_1\not\sim_P A_3$} ;
\end{scope}
\end{tikzpicture}
\captionadd{Kinship.}\label{fig.kinship}
\end{wrapfigure}

\begin{defi}\label{defi.kinship}
Let $\clAll$ be a clustering base and $P\in\clM$.
Two non-empty  $B,B'\in\clB$ are called \begriff{kin below $P$}, denoted as \begriff{$B\sim_P B'$}, iff
$\clQ_P(B\cup B')\ne\emptyset$, i.e., iff there is a base measure $\a\in\clQ$ such that the following holds:
\begin{align*}
\text{(a) } B\cup B'&{}\subset \supp \a &
\text{(b) } \a&{}\le P.
\end{align*}
Moreover, we say that every such $\a\in \clQ_P(B\cup B')$ \begriff{supports} $B$ and $B'$ below $P$.
\end{defi}

Kinship of two events can be used to test whether they belong
to the same root in the cluster forest.
To illustrate this  we consider two events $B$ and $B'$ with 
$B\not\sim_P B'$. Moreover, assume that there is an $A\in \clA$ with 
$B\cup B'\subset A$. Then 
 $B\not\sim_P B'$ implies 
that for all such $A$ there is no $\ca>0$ with $\ca Q_A\le P$. This situation 
is displayed on the right-hand side of Figure~\ref{fig.kinship}.
% In terms of density levels it can be said there is no `path' connecting $C$ and $C'$
% with density bounded away from zero.
Now assume that we have two base measures $\a,\a'\leq P$ on 
$A,A'\in \clA$ that satisfy $A\sim_P A'$ and 
$P(A\cap A')>0$. 
If $\clA$ is rich in the sense of $A\cup A'\in \clA$, then we can find a base measure $\b$ on 
$B:=A\cup A'$ with $\a\leq \b\leq P$  or  $\a'\leq \b\leq P$. 
% Indeed, there is a maximal $\cb>0$ with $\cb Q_B\le \a\vee\a'$
% and it is straightforward to see that then $\a\le\cb Q_B$ or $\a'\le\cb Q_B$.
The next definition relaxes the requirement $A\cup A'\in \clA$, see also Figure~\ref{fig.intersection.subadditivity}
for an illustration.

\begin{wrapfigure}{r}{0.4\textwidth}
\centering
\begin{tikzpicture}[scale=.5]
\begin{scope}%[xshift=§7cm]
\theP

\drawset[red]{.1}{1}{1.8} \node[anchor=south] at (1.2,.1) {$A$};
\drawset[red]{0}{1.5}{4} \node[anchor=south] at (3.5,0) {$A'$};
%\drawset[red]{-.1}{1.5}{1.8} \node[anchor=north] at (1.65,-.1) {$A\cap A'$};

\drawlevel{1}{1.8}{3}{$\a$}
\drawlevel{1.5}{4}{1}{$\a'$}
\drawlevel[dashed,red,align=center,anchor=south]{.8}{4.1}{1.1}{$\b$}

%\node[anchor=west,align=left] at (1.5,4) {$A\PorthoP A'\then\exists\b\le P$ \\ $A\cup A'\subset B$
%and $\a'\le \b$} ;
%\node[anchor=west,align=left] at (6,3) {$A\PorthoP A'$ so there is $\b\le P$ with $A\cup A'\subset B$
%and here $\a'\le \b$} ;
\end{scope}
\end{tikzpicture}
\vspace*{-2ex}
\captionadd{$P$-subadditivity.
%$A\PorthoP A'$ so there is $\b\le P$ with $A\cup A'\subset B$
%and $\a'\le \b$.% There would be no $\b$ with $\a\le \b$ in this case!
}\label{fig.intersection.subadditivity}
\end{wrapfigure}

\begin{defi}
Let $P\in\clM_\Omega^\infty$ be a  measure. For  $B,B'\in\clB$ we write 
\begin{align*}
  B\PorthoP B' &\deff P(B\cap B')=0 \text{ and} \\
  B\nPorthoP B' &\deff P(B\cap B')>0 \, .
\end{align*}
Moreover, a   clustering base
$(\clA,\clQ,\orthoG)$ is called \begriff{$P$-subadditive} iff for all base measures $\a,\a'\le P$ on $A,A'\in\clA$ we have
\begin{equation}\label{p-sub-add}
     A\nPorthoP A' \then
  \exists \b\in\clQ_P(A\cup A')\colon \b\ge \a \tor \b\ge \a'.
\end{equation}
\end{defi}

Note that the implication \eqref{p-sub-add} in particular ensures $\clQ_P(A\cup A') \neq \emptyset$, 
i.e.~$A\sim_P A'$. Moreover, the relation  $\PorthoP$ is weaker than any separation relation $\orthoG$ since we obviously have 
$A\nPorthoP A'\then A\northoD A' \then A\northoG A'$, where 
the second implication is shown in Lemma \ref{lemma.separation.base.2}.
The following definition introduces a stronger notion of additivity.

\begin{defi}
Let $\northoG$ be a relation on $\clB$. An 
$\clA\subset \clB$ is $\northoG$-\begriff{additive} iff for all $A,A'\in\clA$ %we have
\[
  A\northoG A' \then A\cup A'\in \clA.
\]
\end{defi}

The next proposition compares the several notions of (sub)-additivity. In particular it implies that if 
$\clA$ is $\northoD$-additive then $(\clA,\clQ^{\mu,\clA},\orthoG)$ is $P$-subadditive for all $P\in\clM$.

\begin{proposition}\label{prop.additive.implies.subadditive}
Let  
$(\clA,\clQ^{\mu,\clA},\orthoG)$ be a clustering base as in % of the form considered in
Proposition
\ref{prop.indicators}.
If $\clA$ is $\nPorthoP$-additive for some $P\in \clM$, then $(\clA,\clQ^{\mu,\clA},\orthoG)$ is $P$-subadditive.
Conversely, if $(\clA,\clQ^{\mu,\clA},\orthoG)$ is $P$-subadditive for all $P\ll\mu$ then $\clA$ is $\nPortho\mu$-additive
and thus also  $\nPorthoP$-additive.% for all $P\ll\mu$.
\end{proposition}

%  and consequently, 
% if $\clA$ is $\northoD$-additive it is $P$-subadditive for all $P\in\clM$.

\makeatletter{}%!TEX root = article.tex

\subsection{Adapted Simple Measures}

We have already seen that isomonotone
approximations by simple measures are not structurally unique.
In this subsection we will therefore 
identify the most economical structure needed
to approximate a distribution by simple measures.
Such most parsimonious structures will then be used to define 
continuous clusterings.

Let us begin by introducing a different view on simple measures.
\begin{defi}\label{defi.level.of.Q}
Let $(\clA,\clQ,\orthoG)$ be a clustering base and 
 $Q$ be a simple measure on  $F\in\clF_\clA$ with the 
  unique representation
$Q=\sum_{A\in F} \alpha_AQ_A$.
We define 
the map  $\lambda_Q\colon F \to \clQ$  by 
\[
\lambda_Q(A) {}:=  \left(\sum_{A'\in F\colon A'\supset A} \alpha_{A'} Q_{A'}(A)\right)\cdot Q_A\, , \qquad \qquad A\in F.
\]
Moreover, we call
the base measure $\lambda_Q(A)\in \clQ$  the \begriff{level of $A$ in $Q$}.
\end{defi}

\begin{wrapfigure}[5]{r}{0.25\textwidth}
\centering\begin{tikzpicture}[thick,xscale=.2,yscale=.6]
\draw[] (-2,0) -- +(0,.8) -| (8,0);
\draw[yshift=.8cm] (-1,0) -- +(0,.5) -| (4,0) (5,0) -- +(0,1) -| (7,0);
\draw[yshift=1.3cm] (0,0) -- +(0,.2) -| (1,0) (2,0) -- +(0,.5) -| (3,0);

\drawset{.15}{-1}4
\draw[very thick,red] (-1,0) -- (-1,1.3) -| (4,0);
\node[anchor=west] at (4,.4) {$\lambda_Q(A)$} ;
\node[anchor=south] at (1.5,.15) {$A$} ;

\draw[thin,->] (-2.5,0) -- +(11,0) ;% node[anchor=west]{$\Omega$} ;
\draw[thin,->] (-2.2,0) -- +(0,2) ;% node[anchor=south]{$dQ/dx$} ;
\end{tikzpicture}
\vspace*{-2ex}
\caption{\label{fig.simple.measure.level}%
%The left picture shows the level of $A$ in $Q$ and 
%the right picture shows all levels of $Q$.
Level.}
\end{wrapfigure}

In some sense, the level of an $A$ in $Q$ combines all ancestor measures including $Q_A$ and then restricts this combination 
to $A$, see 
 Figure~\ref{fig.simple.measure.level} for an illustration of the level of a node.
With the help of levels we can now describe structurally economical approximations of measures
by simple measures.

\begin{defi}\label{defi.adapted}
Let $(\clA,\clQ,\orthoG)$ be a clustering base and $P\in\clM_\Omega$ a finite measure.
Then a simple measure $Q$ on a forest $F\in\clF_\clA$ is $P$-\begriff{adapted} iff
all direct siblings $A_1,A_2$ in $F$ are:
\begin{description.defi}
\item[$P$-\begriff{grounded}] if they are kin below $P$, i.e.~$\clQ_P(A_1\cup A_2)\neq\emptyset$,  then
there is a parent around them in $F$.

\item[$P$-\begriff{fine}] every $\b\in\clQ_P(A_1\cup A_2)$
can be majorized by a base measure $\tilde\b$ 
that supports all direct siblings $\rep Ak$ of $A_1$ and $A_2$, i.e.
\[
  \b\in\clQ_P(A_1\cup A_2) \then \exists \tilde \b\in\clQ_P(\rep[\cup]Ak) \mbox{ with }
  \tilde \b\ge\b.
\]
\item[\begriff{strictly motivated}] for their levels $\a_1:=\lambda_Q(A_1)$ and $\a_2:=\lambda_Q(A_2)$ in $Q$
there is an $\ca \in (0,1)$ such that  every base measure $\b$ that
supports them below $P$ is not larger than $\ca\a_1$ or $\ca\a_2$, i.e.
\begin{equation}\label{motiv-implicat}
  \forall \b\in \clQ: \b\ge \ca\a_1 \tor \b \ge \ca\a_2 \then \b\not\in\clQ_P(A_1\cup A_2).
%   \nexists \b\in\clQ_P(A\cup A') \colon\qquad
%   \b\ge \ca\a \tor \b \ge \ca\a'
\end{equation}
\end{description.defi}
Finally, an isomonotone sequence 
$(Q_n,F_n)\uparrow P$ is adapted, if $Q_n$ is $P$-adapted for all $n\geq 1$.
\end{defi}

Since siblings are $\orthoG$-separated, they are $\PorthoP$-separated, so strict motivation is
no contradiction to $P$-subadditivity.
Levels are called \begriff{motivated} iff they satisfy condition \eqref{motiv-implicat}
for   $\ca=1$. 
Figure~\ref{fig.adapted} illustrates the three conditions describing adapted measures.
It can be shown that if $\clA$ is $\northoG$-additive, then any isomonotone sequence
can be made adapted.

\begin{figure}[bth]
\centering\begin{tikzpicture}[scale=.6]
%%% Not Motivated
\begin{scope}
\theP
\drawlevel1{1.8}3{$\a$} \drawlevel3{3.8}1{$\a'$}
\drawlevel[dashed,anchor=south,red]{.8}{4}{1.2}{$\b$}
\node[anchor=north west,align=left] at (-1,0) {Not motivated:\\
$\exists \b\in\clQ_P(A\cup A')$ with $\a'\le \b$} ;
\end{scope}

%%% Motivated but not Grounded
\begin{scope}[xshift=8cm]
\theP
\drawlevel[anchor=south]1{1.8}{2.8}{$\a$} \drawlevel[anchor=south]3{3.8}{2.2}{$\a'$}
\drawlevel[dashed,red,align=center]{.8}{4}{1.2}{$\a,\a'\not\le \b$}
\drawlevel[dashed,red,align=center,anchor=north]{.9}{3.9}{2.3}{$\b'\not\le P$}
\node[anchor=north,align=left] at (4,0) {Motivated} ;
\end{scope}

%%% Motivated and Grounded but not fine
%%% adapted
%\end{tikzpicture}
%\caption{\label{fig.motivated} Illustration for strictly motivated.
%On the left  the  two
%base measures $\a,\a'\le P$  are not motivated since $\b$ supports $A,A'$,
%and  majorizes $\a'$. On the right we see that 
%if $\a'$ is made larger then $\a,\a'$ are motivated, even strictly. Indeed any $\b$ that supports $A,A'$
%cannot be larger than  $\a$ or $\a'$ unless it fails to be below $P$.
%}
%\end{figure}
%
%
%
%
%
%\begin{figure}[t]
%\centering\begin{tikzpicture}[scale=.7]
%%% Motivated but not Grounded
\begin{scope}[xshift=16cm]
\theP
\drawlevel1{1.8}3{}%{$\a_1$}
\drawlevel3{3.8}{2.2}{}%{$\a_2$}
%\drawlevel5{5.8}{1.4}{}%{$\a_3$}
\drawset01{1.8} \node[anchor=south] at (1.4,-.25) {$A_1$} ;
\drawset03{3.8} \node[anchor=south] at (3.4,-.25) {$A_2$} ;
\drawlevel[dashed,red,align=center,anchor=south]{.8}{4}{.7}{}%{$A_1\sim_P A_2$}
\node[anchor=north west,align=left] at (0,0) {Not grounded:
$A_1\sim_P A_2$ \\ but without parent} ;
\end{scope}

\begin{scope}[yshift=-6cm,xshift=0cm]
%%% Motivated and Grounded but not fine
\begin{scope}[xshift=2cm]
\theP
\drawlevel1{1.8}3{}%{$\a_1$}
\drawlevel3{3.8}{2.2}{}%{$\a_2$}
\drawlevel5{5.8}{1.4}{}%{$\a_3$}
\drawlevel[align=center,fill=white]{.8}{6}{.7}{}%{$\a_4$}
\drawlevel[dashed,red,align=center,anchor=north]{.8}{4}{1.5}{$\b$}
\drawset01{1.8} \node[anchor=south] at (1.4,-.25) {$A_1$} ;
\drawset03{3.8} \node[anchor=south] at (3.4,-.25) {$A_2$} ;
\drawset05{5.8} \node[anchor=south] at (5.4,-.25) {$A_3$} ;

\node[anchor=north west,align=left] at (-3,0) {Grounded but not fine:
$\b\in\clQ_P(A_1\cup A_2)$ \\cannot be majorized to support $A_3$} ;
\end{scope}

%%% adapted
%\end{tikzpicture}
%\caption{\label{fig.grounded.fine} Illustration for grounded and fine.
%On the left, the levels of the measure  $Q=\a_1+\a_2+\a_3$ are strictly motivated but not grounded.
%Indeed  the direct siblings $A_1,A_2,A_3$ are pairwise kin below $P$
%but they don't have a parent.
%On the right, the direct siblings  $A_1,A_2,A_3$ are grounded
%but they are not fine, since we cannot find a $\b'\leq P$ with $A_1\cup A_2\cup A_3 \subset \supp \b'$. 
%}
%\end{figure}
%
%
%
%
%\begin{wrapfigure}{R}{0.3\textwidth}
%\centering
%\begin{tikzpicture}[scale=.7]
%%% adapted
\begin{scope}[xshift=14cm]
\theP
\begin{scope}%[yshift=.7cm]
\drawlevel1{1.8}3{}%{$\a_1$}
\drawlevel3{3.8}{2.2}{}%{$\a_2$}
\drawlevel5{5.8}{1.4}{}%{$\a_3$}
\end{scope}
\drawlevel[fill=white]{.8}{4}{1.2}{}%{$\a_5$}
\drawlevel[fill=white]{.8}{6}{.7}{}%{$\a_4$}
\node[anchor=north west,align=left] at (-2,0)
{Adapted: grounded, fine and motivated} ;
\end{scope}

%%% adapted
\end{scope}
\end{tikzpicture}
\caption{\label{fig.adapted} 
Illustrations for motivated, grounded, fine, and therefore adapted.
%Compare to Figure 
%\ref{fig.grounded.fine}: $\a_5$ circumvents that  $\a_1$, $\a_2$, $\a_3$ 
% are direct siblings.
}
%\end{wrapfigure}
\end{figure}

The following self-consistency result shows that every simple measure is adapted to itself.
This result
will guarantee that the
extension of the clustering from $\clS$ to $\bar\clS$ is indeed an extension.

\begin{prop}\label{prop.simple.measures.adapted.to.themselves}
Let $(\clA,\clQ,\orthoG)$ be a clustering base. Then every $Q\in\clS(\clA)$ is 
$Q$-adapted.
\end{prop}

\makeatletter{}
\subsection{Continuous Clustering}\label{sec:cont-cluster}

In this subsection we finally introduce continuous clusterings with the help of adapted, isomonotone
sequences. Furthermore, we will 
  show the existence and uniqueness of such clusterings.

Let us begin by introducing a notation that will be used to 
identify two clusterings as identical. To this end let $F_1,F_2\in \clF$ be two forests and $P\in \clM_\Omega$ be 
finite measure. Then we write 
\[
 F_1 =_P F_2\, ,
\]
if there exists a graph isomorphism $\zeta:F_1\to F_2$ such that $P(A \triangle \zeta(A) ) = 0$
for all $A\in F_1$.
Now our first result shows that 
adapted isomonotone limits of
two different sequences coincide in this sense.

\begin{thm}\label{thm.uniqueness}
Let $(\clA,\clQ,\orthoG)$ be a stable clustering base
and $P\in\clM_\Omega$ be a finite measure such  that $\clA$ is $P$-subadditive.
% 
% Assume $P\in\clM_\Omega$ is a finite measure and $(\clA,\clQ,\orthoG)$ is a $P$-subadditive clustering base.
If $(Q_n,F_n),(Q_n',F_n')\uparrow P$ are adapted  isomonotone sequences
then we have 
\[
\lim_n s(F_\infty) =_P \lim_n s(F_\infty')\, .
\]
% 
% then their isomonotone limits $F_\infty^i:=\lim_n s(F_\infty^i)$, $i=1,2$ are
% equal up to $P$-null sets, i.e.\ there is a graph isomorphism $\zeta\colon F_\infty^1\to F^2_\infty$
% such that for all $A\in F_\infty^1$ we have 
% \[
%   P\big(A\triangle \zeta(A)\big)=0 \, .
% \]
\end{thm}

 Theorem \ref{thm.uniqueness} shows that different adapted sequences approximating a measure $P$ 
 necessarily have isomorphic forests and that the corresponding limit  nodes of the forests 
 coincide up to $P$-null sets. This result makes the following axiom possible.

\begin{axiom}[Continuous Clustering]\label{axiom.continuous.clustering}
Let $(\clA,\clQ,\orthoG)$ be a clustering base, $\clP\subset\clM_\Omega$ be a set of measures.
% and $\cl \colon\clP \to \clF$ be an additive clustering. 
We say that 
$\cl \colon\clP \to \clF$ is a \begriff{continuous clustering}, if it is an additive clustering and 
for all $P\in \clP$ and all
%for which 
% 
% A \begriff{continuous clustering} is an additive clustering $\cl \colon\clP \to \clF$
% 
% 
% if for all   $P\in\clM_\Omega$ for which 
 adapted isomonotone sequences
$(Q_n,F_n)\uparrow P$ we have 
\[
 c(P) =_P \lim_n s(F_\infty)\, .
\]
% 
% 
% 
% 
% there exists a graph isomorphism $\zeta\colon c(P) \to \lim_n s(F_\infty)$ such that for all $A\in \cl(P)$ we have 
% \[
%   P\big(A\triangle \zeta(A)\big)=0 \, .
% \]
% % % 
% % % we have %$P\in\clP$ and
% % % %\[
% % % %  \tilde\clP := \set{ P\in\bar\clS\mid \text{$\clA$ is $P$-subadditive} }.
% % % %\]
% % % %For all $P\in\tilde\clP$ and all simple adapted isomonotone sequences $(Q_n,F_n)\uparrow P$:
% % % \[
% % %   \cl(P) = \lim_n s(F_n) \qqtext{(up to $P$-null sets)}.
% % % \]
% % % 
% % % such that \ $\cl$ and $\clP$ satisfy the following axiom:
% % % \begin{description}
% % % \item[\Continuity:] for all finite $P\in\clM_\Omega$ for which there is a simple adapted isomonotone sequence
% % % $(Q_n,F_n)\uparrow P$ we have $P\in\clP$ and
% % % %\[
% % % %  \tilde\clP := \set{ P\in\bar\clS\mid \text{$\clA$ is $P$-subadditive} }.
% % % %\]
% % % %For all $P\in\tilde\clP$ and all simple adapted isomonotone sequences $(Q_n,F_n)\uparrow P$:
% % % \[
% % %   \cl(P) = \lim_n s(F_n) \qqtext{(up to $P$-null sets)}.
% % % \]
% % % \end{description}
\end{axiom}

The following, main result of this section shows that there exist continuous clusterings and that 
they are uniquely determined on a large subset of $\bar\clS(\clA)$.

% We now show that by the uniqueness theorem
% any continuous clustering is uniquely determined (up to $P$-null sets)
% for all $P$ which are limits of adapted isomonotone sequences.

\begin{thm}\label{thm.extension.axioms}
Let $(\clA,\clQ,\orthoG)$ be a stable clustering base and set
\[
  \clP_{\clA} := \set{ P\in\bar\clS(\clA)\mid \text{$\clA$ is $P$-subadditive and
  there is $(Q_n,F_n)\nearrow P$ adapted} }.
\]
Then there exists a continuous clustering $\cl_{\clA}\colon \clP_{\clA}\to\clF_{\bar\clA}$.
% we have 
% 
Moreover, $\cl_{\clA}$ is unique on $\clP_{\clA}$, that is, for all continuous clusterings $\cl\colon\tilde\clP\to\clF$ 
we have 
% 
% is an extension of $\cl_{\clA}$,
% i.e.\ $\clP_{\clA}\subset\tilde\clP$ and for all $P\in\clP_{\clA}$ we have
\[
  \cl_{\clA}(P) =_P  \cl(P)  \, , \qquad \qquad P\in \clP_{\clA}\, .
\]
\end{thm}

Recall from Proposition~\ref{prop.additive.implies.subadditive} that 
$\clA$ is $P$-subadditive for all $P\in \clM_\Omega$ if $\clA$ is $\northoD$-additive.
It can be shown that if $\clA$ is $\northoA$-additive, then any
isomonotone sequence can be made adapted.
In this case we thus have $\clP_{\clA} = \bar\clS(\clA)$ and Theorem 
\ref{thm.extension.axioms} shows that there exists a unique continuous clustering on $\bar\clS(\clA)$.

\makeatletter{}%!TEX root = article.tex

\subsection{Density Based Clustering}\label{sub.density}

Let us recall from Proposition \ref{prop.indicators} 
that a simple way to define a set of base measures $\clQ$ was with the help of 
a reference measure $\mu$. Given a stable separation relation $\orthoG$, we denoted
the resulting stable clustering base by 
 $(\clA,\clQ^{\mu,\clA},\orthoG)$. Now observe that 
 for this clustering base
 every $Q\in \clS(\clA)$ is $\mu$-absolutely continuous and its unique representation
%  $Q=\sum_{A\in F}\alpha_A Q_A$ 
 yields the $\mu$-density  $f = \sum_{A\in F}  {\alpha_A} 1_A$ for suitable coefficients $\alpha_A>0$.
Consequently,  each level set $\{f>\lambda\}$ consists of 
some  elements $A\in F$,
% namely, the roots of the sub-forest 
% $F_\lambda := \{A\in F: \lambda_Q(A)(A) >\lambda\}$.
and if all elements in $\clA$ are connected, the additive clustering $c(Q)$ of $Q$
thus coincides with the ``classical'' cluster tree obtained from the  level sets.
It is therefore natural to ask, whether such a relation still holds for continuous clusterings
on distributions $P\in \clP_\clA$.

Clearly, the first  answer to this question needs to be negative, since in general the cluster tree 
is an infinite forest whereas our clusterings are always finite. To illustrate this, let us consider
the \begriff{Factory} density on $[0,1]$, which is defined by 
% $f(x) = \tfrac13 - |x-\tfrac13|$ for $x\le \tfrac12$
% and $f(x):=f(1-x)$ for $x>\tfrac12$:
\begin{align*}
  f(x) &{}:= \begin{cases}
    1-x, & \text{if }x\in [0,\tfrac12) \\ 1, & \text{if }x\in [\tfrac12,1]
  \end{cases}
  &
 \begin{tikzpicture}[baseline=.5cm,scale=1,domain=0:1]
 \newcommand{\jump}[3]{\draw (#1,#2) circle (2pt) ; \fill (#1,#3) circle (2pt); \draw[dotted](#1,#2) to (#1,#3); }
  \draw (-.2,0) -- (0,0)  (0,1) -- (.5,.5)   (.5,1) -- (1,1)  (1,0) -- (1.2,0);
  \jump001 \jump{.5}{.5}1 \jump101
%  \draw (0,0) circle (2pt) ; \fill (0,1) circle (2pt);
%  \draw (.5,.5) circle (2pt) ; \fill (.5,1) circle (2pt);
%  \draw (1,0) circle (2pt) ; \fill (1,1) circle (2pt);
  \end{tikzpicture}
\end{align*}
Clearly, this gives the following $\orthoD$-decomposition of the level sets: 
\begin{align*}
  \{f >\lambda\} &{}= \begin{cases}
    [0,1], & \text{if }\lambda < \tfrac12, \\
    [0,1-\lambda) \orthocupD [\tfrac12,1],
    & \text{if }\tfrac12 \le \lambda < 1,
  \end{cases}
\end{align*}
which leads to the  clustering forest 
$F_f = \set{ [0,1],[\tfrac12,1]}\cup\set{[0,1-\lambda)\mid \tfrac12\le\lambda<1}$.
Now observe that 
even though $F_f$ is infinite, it is as graph somehow simple: there is a root $[0,1]$, 
a node $[\tfrac12,1]$, and an infinite chain $[0,1-\lambda)$, $\tfrac12\leq \lambda<1$.
Replacing this chain by its supremum $[0,\tfrac 12)$ we obtain the structured forest 
\[
  \bigl\{[0,1],\,[0,\tfrac12),[\tfrac12,1] \bigr\},
\]
for which we can then ask whether it coincides with the continuous clustering obtained from 
$(\clA,\clQ^{\mu,\clA},\orthoD)$ if $\clA$ consists of all closed intervals in $[0,1]$
and $\mu$ is the Lebesgue measure.

To answer this question we first need to formalize the operation that assigns a structured
to an infinite forest.
To this end, let $F$ be an arbitrary $\orthoG$-forest. We say that $\clC\subset F$ is a pure chain, iff
for all $C,C'\in \clC$ and $A\in F\setminus \clC$ the following two implications hold:
\begin{align*}
   A\subset C&\then A\subset C' ,\\
  C\subset A &\then C'\subset A .
\end{align*}
Roughly speaking, the first implication ensures that no node above a bifurcation is contained in the chain, while the second
implication ensures that no node below a  bifurcation is contained in the chain.
With this interpretation in mind it is not surprising that we can define the structure of the forest $F$ with the help of
the maximal pure chains
by setting
\begin{align*}
      s( F) := \bigg\{ \,&{}\bigcup \clC \mid \text{$\clC \subset  F$ is a maximal pure chain} \bigg\}\, .
\end{align*}
Note that for infinite forests the structure  $s( F)$ may or may not be finite. For example, for the factory density
it is finite as we have already seen above. 

We have seen in Lemma \ref{lemma.limit.is.forest} that the
 nodes of a continuous clustering are $\orthoG$-separated elements of $\bar \clA$. Consequently, it only makes sense to 
 compare continuous clustering with the structure of a level set forest, if this forest shares this property.
This is ensured in the following definition.

\begin{defi}
Let $f\colon\Omega\to[0,\infty]$ be a measurable function and $\clAll$ be a stable clustering base.
Then $f$ is of $\clAll$-\begriff{type} iff there is a dense subset $\Lambda\subset [0,\sup f)$
such that  for all $\lambda\in\Lambda$ the level set $\{f >\lambda\}$
is a finite union of pairwise $\orthoG$-separated events $B_1(\lambda),\ldots,B_{k(\lambda)}(\lambda)\in\bar\clA$.
If this is the case the   \begriff{level set $\orthoG$-forest} is given by
\[
    F_{f,\Lambda} := \{ B_i(\lambda) \mid i\le k(\lambda)\tand \lambda \in \Lambda \}.
\]
%For any countable forest $\tilde F$ we define the \begriff{generalized structure} as:
%\begin{align*}
%    \tilde s(\tilde F) := \bigg\{ \,&{}\bigcup \clC \mid \text{$\clC \subset \tilde F$ is a maximal chain s.t.\ for all
%  $C,C'\in\clC$ and $A\in F\setminus\clC$: } \\&{} (A\subset C\then A\subset C') \tand (A\supset C \then A\supset C') \bigg\}
%\end{align*}
\end{defi}

Note that for given $f$ and $\Lambda$ the forest $F_{f,\Lambda}$ is indeed well-defined 
since $\orthoG$ is an $\bar\clA$-separation relation by Lemma \ref{lemma.limit.is.forest} 
and therefore the decomposition of $\{f >\lambda\}$ into the sets 
 $B_1(\lambda),\ldots,B_{k(\lambda)}(\lambda)\in\bar\clA$ is unique by  Lemma~\ref{lemma.separation.base.2}.
 
With the help of these preparations we can now formulate the main result of this subsection,
which compares continuous clusterings with the structure of level set $\orthoG$-forests:

\begin{thm}\label{thm.clustering.density.level.sets}
Let $\mu\in\clM_\Omega$,
$(\clA,\clQ^{\mu,\clA},\orthoG)$ the stable clustering based described in Proposition  \ref{prop.indicators},
and $P\in\clM_\Omega$ such that $\clA$ is $P$-subadditive.
Assume that  $P$ has a $\mu$-density $f$
that is of $\clAll$-type 
with a dense subset $\Lambda$ such that $s(F_{f,\Lambda})$ is finite
and for all 
% 
% Let $\mu\in\clM_\Omega$ be a measure, $P\in\clM_\Omega$ with $P\ll \mu$,
% $\clA\subset\clK(\mu)$ be $P$-subadditive,
% and consider the clustering base $(\clA,\clQ^{\mu,\clA},\orthoG)$.
% 
% $P\in\bar\clS$ if there is a version of the density $f=\frac{dP}{d\mu}$
% that is of $\clAll$-type with a dense subset $\Lambda$
% s.t.\ 
% 
% 
% $\tilde s(\tilde F_{f,\Lambda})$ is finite and for all 
$\lambda\in\Lambda$
and all $i<j\le k(\lambda)$ we have  $\overline B_i(\lambda)\orthoG \overline B_j(\lambda)$.
Then we have $P\in \bar \clS(\clA)$ and 
\[
   \cl(P)=_\mu  s( F_{f, \Lambda})\, .
\]
\end{thm}
On the other hand, it is not difficult to show that if $P\in\bar\clS(\clA)$ then $P$ has a density of $\clAll$-type.
We do not know though whether there has to a density of $\clAll$-type for that even the closure of siblings are separated.

If $\supp\mu\ne\Omega$ one might think that this is not true
since on the complement of the support anything goes.
To be more precise---if $\mu$ is not inner regular and hence no support is defined---assume
there is an open set $O\subset\Omega$ with $\mu(O)=0$.
This then means that there is no base set $A\subset O$, because base sets are support sets.
Hence anything that would happen on $O$ is determined by what happens in $\supp P$!
%\fixedtodo{I think we  need $\supp \mu = \Omega$ since otherwise there seems to be 
% too much room to manipulate the density outside the support! Does this condition (implicitely) 
%appear in the proof? If not, is there a gap/bug?}
%
%
%\fixedtodo{Mention that for $P\in \bar \clS(\clA)$ everything but the closure condition is satisfied.}

In the literature density based clustering is only considered for continuous densities since
they may serve as a canonical version of the density. The following result investigates such densities.

% Now what about continuous densities?
% The best case is when all open, connected sets are in $\bar\orthoA$, cf.~Proposition~\ref{prop.neighborhood.base}.

\begin{prop}\label{prop.clustering.density.level.sets.continuous}
For a compact  $\Omega\subset\dsR^d$
and a measure $\mu\in\clM_\Omega$  we consider the 
stable clustering base $(\clA,\clQ^{\mu,\clA},\orthoD)$. We assume that 
all open, connected sets are contained in $\bar\clA$ and that 
$P\in\clM_\Omega$ is a finite measure such that  
$\clA$ is $P$-subadditive. If $P$ has a continuous density $f$ that has only finitely many 
% 
% with $P\ll \mu$,
% 
% and consider the stable clustering base $(\clA,\clQ^{\mu,\clA},\orthoD)$.
% Assume that all open, connected sets are in $\bar\clA$ and $f\colon\Omega\to\dsR$ is a continuous density.
% 
% If $f$ has only finitely 
local maxima $\rep{x^*}k$ then $P\in\clP_\clA$ and there a bijection
$\psi\colon\{\rep{x^*}k\}\to\min\cl(P)$
such that 
\[
   x^*_i\in \psi(x^*_i)\, .
\]
In this case $\cl(P)=_\mu \set{ B_i{\lambda} \mid i\le k(\lambda) \tand \lambda\in \Lambda_0 }$
where $\Lambda_0=\set{0=\lambda_0<\ldots<\lambda_m<\sup f}$ is the finite set of levels at which the splits occur.
\end{prop}

\makeatletter{}
%%%%%%%%%%%%%%%%%%%%%%%%%%%%%%%%%%%%%%%%%%%%%%%%%%%%%%%%%%%%%%%%
%%%%%%%%%%%%%%%%%%%%%%%%%%%%%%%%%%%%%%%%%%%%%%%%%%%%%%%%%%%%%%%%
%%%%%%%%%%%%%%%%%%%           Examples         %%%%%%%%%%%%%%%%%
%%%%%%%%%%%%%%%%%%%%%%%%%%%%%%%%%%%%%%%%%%%%%%%%%%%%%%%%%%%%%%%%
%%%%%%%%%%%%%%%%%%%%%%%%%%%%%%%%%%%%%%%%%%%%%%%%%%%%%%%%%%%%%%%%

%\clearpage
\section{Examples}
\label{section.examples}

After having given the skeleton of this theory we now give more examples of how to use it.
This should as well motivate some of the design decisions.
It will also become clear in what way the choice of a clustering base $(\clA,\clQ,\orthoG)$
influences the clustering.

% We begin with examples for separation relations and base sets and give
% tools to construct many more interesting examples.
% Afterwards we proceed by examining the clustering of indicator functions
% -- i.e.\ we give a clustering of sets, or show how to generalize connected components.
% In the last semi-theoretical section we look at the way base sets of different Hausdorff
% dimension can be intertwined.

\makeatletter{}%!TEX root = article.tex

\subsection{Base Sets and Separation Relations}
\label{sub.examples.separations}\label{sub.examples.base.sets}

In this subsection we present several examples of clustering bases. 
Our first three examples consider different separation relations.

\begin{example}[Separation relations]\label{example.separation.relations}
The following define stable $\clA$-separation relations:
\begin{description.defi}
\item[Disjointness]
If $\clA\subset \clB$ is a collection of non-empty, closed, and topologically connected sets
then 
\[
  B \orthoD B' \iff B\cap B'=\emptyset.
\]
%is a stable $\clA$-separation relation.
% if all $A\in\clA$ are closed and topologically connected
\item[$\tau$-separation]
Let $(\Omega,d)$ be a metric space, $\tau>0$, and $\clA\subset \clB$ be 
 a collection of non-empty, closed, and $\tau$-connected sets
then 
\begin{align*}
  B \orthoT B' &{}\deff d(B,B')\ge\tau.
\end{align*}
%is a stable $\clA$-separation relation.
% $A$ is \begriff{$\tau$-connected} iff for all $x,x'\in A$ there are
% $\rep xk\in A$ with $x=x_1$, $x'=x_k$, and $d(x_i,x_{i+1})<\tau$ for all $i<k$.
% Then $\orthoT$ is an $\clA$-separation relation if all $A\in\clA$ are $\tau$-connected.
%Remark that $\orthoT$ is $\clB$-stable and hence fulfills~\eqref{eq.stability.alternative}.
\item[Linear separation]
Let $H$ be a %finite-dimensional
Hilbert space with inner product $\<\cdot\mid\cdot>$ and $\Omega\subset H$.
Then non-empty events $A,B\subset\Omega$ are \begriff{linearly separated},
$A\orthoL B$, iff $A\orthoD B$ and
\begin{align*}
   \exists v\in H\setminus\{0\}, \alpha\in\dsR \,
  \forall a\in A, b\in B\colon
  \<a\mid v> \le \alpha \tand \<b\mid v> \ge \alpha.
\end{align*}
The latter means there is an affine hyperplane $U\subset \Omega$
such that  $A$ and $B$ are on different sides.
Then $\orthoL$ is a $\clA$ separation relation
if no base set $A\in\clA$ can be written as a finite union
of pairwise $\orthoL$-disjoint closed sets.
It is stable if $H$ is finite-dimensional.
%
% 
% 
% 
% 
% The previous separation relations had the property
% that if $\rep Ak$ were pairwise separated then $A_k\orthoD (\rep[\cup] A{k-1})$.
% For linear separation this does not hold in general:
% \begin{center}
% \begin{tikzpictureC}
% %\draw (0,0) -- (150:1) (0,0) -- (30:1) ;
% \fill[blue](150:.05) -- +(0,.5) -- +(210:.5) -- cycle ;
% \fill[blue] (30:.05) -- +(0,.5) -- +(330:.5) -- cycle ;
% \fill[blue] (0,-.05) -- +(330:.5) -- +(210:.5) -- cycle ;
% \node[anchor=east] at (150:.2) {$A_1$};
% \node[anchor=west] at (30:.2) {$A_2$};
% \node[anchor=north] at (0,-.2) {$A_3$};
% \end{tikzpictureC}
% \end{center}
% Here $A_1,A_2,A_3$ are pairwise linear separated but $A_3\northoL (A_1\cup A_2)$.
\end{description.defi}
\end{example}

%\streichbar{
%The following relations are not separation relations, yet they are still useful as we will see afterwards:
%
%
%\begin{example}\label{example.Portho}
%For $B,B'\in \clB$, $\tau>0$ and $P\in \clM_\Omega$ consider the following relations:
%\begin{enumerate}
%\item $B \PorthoO B' \deff \inner B\cap \inner B' = \emptyset$
%\item $B \PorthoT B' \deff \nexists x_0\in\Omega\colon  \{x\in \Omega: d(x,x_0)\leq \tau  \}\subset B\cap B'$, if $(\Omega,d)$ is a
%metric space. \margincomment{This is easy to write, 
%but maybe we actually want that the $\tau$-inner is non-empty and connected?}
%\item $B \PorthoP B' \deff P(B\cap B')=0$.
%\end{enumerate}
%These are not separation relations, since  connectedness is general violated as the following inclusion shows 
%\[
  %[-1,1] \subset [-1,0] \orthocup[\PorthoG] [0,1]\, .
%\]
%Still we write $\nPorthoO,\nPorthoT,\nPorthoP$ for the the negations of these relations.
%Moreover, observe the following implications which hold provided that the relations are defined.
%\[
  %B \orthoG B' \then B \orthoD B' \then B \PorthoO B' \tand B\PorthoT B' \tand B\PorthoP B'
%\]
%\end{example}
%}

Our next goal is to present some examples of base set collections $\clA$. Since 
these describe the sets we need to agree upon that their can only be trivially clustered, 
smaller collections $\clA$ are generally preferred.
Let $\mu$ be the Lebesgue measure on $\dsR^d$.
To define possible collections $\clA$ we will consider the following building blocks in $\dsR^d$:
% 
% Design of base sets $\clA$ gives a powerful tool which can be
% tailored using not only separation relations but also more general relations as
% in Example~\ref{example.Portho}.
% There are usually nice building blocks one might want to use, e.g.:
\begin{align*}
\clCdyad &{}:= \big\{ \text{axis-parallel boxes  with dyadic coordinates}\, \big\}, \\
\clC_p &{}:= \big\{ \text{closed $\ell_p^d$-balls }\, \big\}, \quad p \in \left[1,\infty \right],\\ % = \set{\dsB_p(x,\rho) \mid x\in\Omega,\rho>0}, \\
\clCconv &{}:= \big\{ \text{convex and compact $\mu$-support sets}\, \big\}.
% \qquad\text{if $\Omega$ is contained in a vector space}.
\end{align*}
$\clCdyad$ corresponds to the cells of a histogram whereas $\clC_p$ has connections to moving-window density estimation.
When combined with $\orthoD$ or $\orthoT$ and base measures of the form \eqref{ex-clust-base}
these collections may already serve as clustering bases. 
 However, 
$\bar\clC_\bullet$ and $\bar\clS_\clC$ are not very rich since monotone increasing 
sequences in $\clC_\bullet$ converge to sets of the same shape, and hence the sets in 
$\bar\clC_\bullet$ have the same shape constraint as those in $\clC_\bullet$.
As a result the sets of measures $\bar\clS_{\clC_\bullet}$ for which we can determine the unique
continuous clustering are rather small.
However, more interesting collections can be obtained by considering 
    finite, connected unions built of such sets.
    To describe such unions in general we need the following definition.
    
 \begin{defi}\label{defi.intersection.graph}
Let $\PorthoG$ be a relation on $\clB$,  $\nPorthoG$ be its negation, and
$\clC\subset \clB$ be a class of non-empty events.
The $\PorthoG$-\begriff{intersection graph} on $\clC$, $\clG_{\PorthoG}(\clC)$,
has $\clC$ as nodes and there is an edge between $A,B\in \clC$ iff $A\nPorthoG B$.
We define:
\begin{align*}
    \dsC_{\PorthoG}(\clC) &{}
    := \{\, \rep[\cup]Ck \mid  \rep Ck\in \clC  \text{ and the graph
      $\clG_{\PorthoG}(\{\rep Ck\})$ is connected} \,\}.
\end{align*}
\end{defi}   
Obviously any separation relation can be used. But one can also consider
weaker relations like $\PorthoP$, or e.g.\ $A\PorthoG A'$ if $A\cap A'$ has empty interior,
or if it contains no ball of size $\tau$.
Such examples yield smaller $\clA$ and indeed in these cases $\bar\clS$ is much smaller.

The following example provides stable clustering bases.
    
% Recall that for continuous clustering we need $P$-subadditivity.
% Proposition~\ref{prop.separation.main} gives sufficient conditions
% for $\orthoG$ being a $\dsC_{\PorthoG}(\clC)$-separation relation and the latter
% being $\cancel{\PorthoG}$-additive.

\begin{example}[Clustering bases]\label{ex-Axx}
%Consider the following collections of path-connected, compact sets:
%\begin{align*}
%\clCdyad &{}:= \big\{ \text{axis-parallel boxes with dyadic coordinates}\, \big\}, \\
%\clC_p &{}:= \big\{ \text{closed $L^p$-balls}\, \big\} = \set{\dsB_p(x,\rho) \mid x\in\Omega,\rho>0}, \\
%\clCconv &{}:= \big\{ \text{convex and compact sets}\, \big\}.
%\end{align*}
The following examples are $\northoD$-additive:
\begin{align*}
\clAdyad &{}:= \dsC_{\orthoD}(\clCdyad) &&= \set{\text{finite connected unions of boxes with dyadic coordinates}}, \\
\clA_p &{}:= \dsC_{\orthoD}(\clC_p) &&= \set{\text{finite connected unions of closed $L^p$-balls}}, \\
\clAconv &{}:= \dsC_{\orthoD}(\clCconv) &&= \set{\text{finite connected unions of convex $\mu$-support sets}}.
\end{align*}
Then $\clAdyad,\clA_p,\clAconv\subset\clK(\mu)$.
Furthermore the following examples are $\northoT$-additive:
\begin{align*}
\clAdyadT &{}:= \dsC_{\orthoT}(\clCdyad), &
\clA_p^\tau &{}:= \dsC_{\orthoT}(\clC_p), &
\clAconvT &{}:= \dsC_{\orthoT}(\clCconv).
\end{align*}
This leads to the following examples of stable clustering bases:
\begin{align*}
(\clAdyad,\clQ^{\mu,\clAdyad},\orthoD) &{}, &
(\clA_p,\clQ^{\mu,\clA_p},\orthoD) &{}, &
(\clAconv,\clQ^{\mu,\clAconv},\orthoD) &{}, \\
(\clAdyadT,\clQ^{\mu,\clAdyadT},\orthoT) &{}, &
(\clA_p^\tau,\clQ^{\mu,\clA_p^\tau},\orthoT) &{}, &
(\clAconvT,\clQ^{\mu,\clAconvT},\orthoT) &{}, \\
(\clAdyad,\clQ^{\mu,\clAdyad},\orthoT) &{}, &
(\clA_p,\clQ^{\mu,\clA_p},\orthoT) &{}, &
(\clAconv,\clQ^{\mu,\clAconv},\orthoT) &{}.
\end{align*}
The first row is the most common case, using connected sets and their natural separation relation.
The second row is the $\tau$-connected case.
The third row shows how the fine tuning can be handled:
We consider connected base sets, but siblings need to be $\tau$-separated,
hence e.g.\ saddle points cannot be approximated.
%\fixedtodo{Is this last example okay? And the informal decription?}
\end{example}

%\streichbar{
%\begin{example}
%We define $\clA_p^\circ :=\dsC_{\PorthoO}(\clC_p)$. Then 
  %$(\clA_p^\circ,\clQ^{\mu,\clA_p^\circ},\orthoD)$ is a stable clustering base
%and fulfills $\PorthoO$-intersection additivity.
%Hence it is $P$-subadditive for all $P\ll \mu$.
%\end{example}
%
%
%\begin{example}
%We define 
%\begin{align*}
  %\clA_p^{-\tau} :=\dsC_{\PorthoT}(\set{C\in\clC_p\mid \diam(C)>\tau}).
%\end{align*}
%Then
%$(\clA_p^{-\tau},\clQ^{\mu,\clA_p^{-\tau}},\orthoD)$ is  a stable clustering base, which is 
 %$P$-subadditive for all $P\in\bar\clS(\clA_p^{-\tau})$. %  see Proposition~\ref{prop.densities.subadditivity}.
%\end{example}
%}

 The larger the extended class $\bar\clA$ is, the more measures we can cluster. The following proposition
provides a sufficient condition for $\bar\clA$  being rich.

\begin{prop}\label{prop.neighborhood.base}
Assume all $A\in\clA$ are path-connected. Then all $B\in\bar\clA$ are path-connected.
Furthermore assume that $\clA$ is intersection-additive and that it
contains a countable neighbourhood base. % $\clA'\subset \clA$, i.e.:
% \[
%   \forall \omega \in\Omega, O\in\clT\colon
%   \omega\in O\then \exists A\in\clA'\colon x\in \inner A\subset A\subset O
% \]
Then $\bar\clA$ contains all open, path-connected sets.
\end{prop}

One can show  that the   first statement also holds for topological connectedness.
% cf.\ Lemma~\ref{lemma.topological.connectedness.monotone.limit}.
Furthermore note that 
$\clCdyad$ is  a countable neighbourhood base, and therefore
 $\clAdyad$, $\clA_p$, and $\clAconv$ fulfill the conditions of Proposition~\ref{prop.neighborhood.base}.

\begin{figure}\label{fig.example.base.sets}
\centering
\begin{tikzpicture}[scale=.4]
%\fill[gray!20] plot [smooth cycle] coordinates {(-1,-1)  (-1,2) (2,1)  (3,2)  (5,0) };
%\fill[white] (0,0) circle (.4cm) ;
\fill[blue] (.5,-.5) rectangle (3,.5) (1,1) rectangle (-1,.5) rectangle(-.5,-1) rectangle(0,-.5) rectangle(1,-.75)
(4,0) rectangle ++(-1,1) rectangle +(.5,.5) +(.5,0) rectangle +(.75,.25)  +(0,.5) rectangle +(.25,.75)
(2.5,.5) rectangle +(.5,.75) rectangle (2.75,1.875)
(-1,1) rectangle (0,1.5);
\node at (5.5,1) {$\in\clAdyad$} ;
\draw[dotted,thick] plot [smooth cycle] coordinates {(-1,-1)  (-1,2) (2,1)  (3,2)  (5,0) };
\draw[dotted,thick] (0,0) circle (.4cm) ;

\begin{scope}[xshift=10cm]
%\fill[gray!20] plot [smooth cycle] coordinates {(-1,-1)  (-1,2) (2,1)  (3,2)  (5,0) };
%\fill[white] (0,0) circle (.4cm) ;
\fill[blue] (1.5,.1) circle(.9cm) (3.3,.6) circle (1)
\foreach \x in {(0,1.1),(.9,.6),(-.8,.65),(-.5,1.3),(1,-.3),(4.2,.4)} {\x circle (.6) }
\foreach \x in {(-.9,0),(3,1.5),(-.7,-.6)} {\x circle (.4) }
\foreach \x in {(2.525,-.35),(-.1,-.7),(.55,-.65)} {\x circle (.3) }
(4.8,.2) circle (.2)
;
\node at (5.5,1) {$\in\clA_2$} ;
\draw[dotted,thick] plot [smooth cycle] coordinates {(-1,-1)  (-1,2) (2,1)  (3,2)  (5,0) };
\draw[dotted,thick] (0,0) circle (.4cm) ;
\end{scope}

\begin{scope}[xshift=20cm]
\fill[blue] plot [smooth cycle] coordinates {(-1,-1)  (-1,2) (2,1)  (3,2)  (5,0) };
\fill[white] (0,0) circle (.4cm) ;
\node at (7,1.5) {$\in\overline{\clAdyad},\bar\clA_2$} ;
\end{scope}
\end{tikzpicture}
\caption[Some examples of base sets]{Some examples of sets in $\clAbox$, $\clAconv$
and their closure. }
\end{figure}

\subsection{Clustering of Densities}
\label{subsec.examples.clustering.densities}

Following the manual to cluster densities given in Theorem~\ref{thm.clustering.density.level.sets} by  
decomposing the density level sets into $\orthoG$-disjoint components, one first needs to understand
the $\orthoG$-disjoint components of general events. In this subsection we investigate such decompositions
and the resulting clusterings.
%Theorem~\ref{thm.clustering.density.level.sets} gives a manual on
%how to cluster densities:
%Decompose the density level sets into $\orthoG$-disjoint components.
%To apply this approach successfully, one first needs to understand
%the $\orthoG$-disjoint components of general events.
%The   goal of this subsection is to investigate such decompositions
%and the resulting clusterings. 
%  
%  
% Then it is trivial deduct the clustering from that.
% 
% We begin by considering some examples on the real line $\Omega \subset \dsR$.
% For higher dimensions we have to first understand what the components are.
% Then we can give some examples of clustering.
% 
We assume 
$\mu$ to be the Lebesgue measure on some suitable $\Omega\subset\dsR^d$ and let the base 
measures be the ones considered in Proposition \ref{prop.indicators}.
For visualization purposes we further restrict our considerations 
to the one- and two-dimensional case, only.

\makeatletter{}\subsubsection{\texorpdfstring{Dimension $d=1$}{Dimension d=1}}

In the one-dimensional case, in which $\Omega$ is an interval, 
 the examples  $\clA_p = \clAconv$ simply consist of 
 compact intervals, and their monotone closures  
 consist  of all intervals.
 To understand the resulting clusters let us first consider 
 the   \begriff{twin peaks} density:
% $f(x) = \tfrac13 - |x-\tfrac13|$ for $x\le \tfrac12$
% and $f(x):=f(1-x)$ for $x>\tfrac12$:
\begin{align*}
  f(x):= \tfrac{1}{3} - \min\set{|x-\tfrac{1}{3}|,|x-\tfrac{2}{3}|}. \qquad
\begin{tikzpicture}[baseline=.25cm,yscale=.7,xscale=.7]
\begin{scope}
\draw[->] (0,0)--(0,1) node[anchor=west] at (2.5,.6) {$f(x)$};
\draw[->] (0,0)--(3,0) node[anchor=west] {$x$};
\draw (.5,0) -- +(0,-.1) ;% node[anchor=north] {$1/6$};
\draw (1,0) -- +(0,-.1) ;% node[anchor=north] {\small $1/3$};
\draw (1.5,0) -- +(0,-.1) ;% node[anchor=north] {$1/2$};
\draw (2,0) -- +(0,-.1) ;% node[anchor=north] {\small $2/3$};
\draw (2.5,0) -- +(0,-.1) ;% node[anchor=north] {$5/6$};
\draw (0,.5) -- +(-.1,0) ;% node[anchor=east] {$1/6$};
%\draw (0,1) -- +(-.1,0) ;% node[anchor=east] {$1/3$};
\draw[thick] (0,0) -- (1,1) -- (1.5,.5) -- (2,1) -- (3,0) ;
\draw[dotted] (.5,.5) -- (2.5,.5) ;
\end{scope}
\end{tikzpicture}
\end{align*}
Clearly, this gives the following $\orthoD$-decomposition of the level sets: 
\begin{align*}
  H_f(\lambda) &{}= %\begin{cases}
    (\lambda,1-\lambda) \text{ for }  \lambda < \tfrac16, &
  H_f(\lambda) &{}= 
    (\lambda,\tfrac12-\lambda) \orthocupD (\tfrac12+\lambda,\lambda)
    \text{ for } \tfrac16 \le \lambda < \tfrac13
  %\end{cases}
\end{align*}
and hence the $\orthoD$-clustering forest is $\set{ (0,1), (\tfrac16,\tfrac12),(\tfrac12,\tfrac56) }$.
Since, none of the boundary points can be reached,
any isomonotone, adapted sequence yields this result.
 However, the clustering changes, if the separation relation $\orthoT$ is considered.
We obtain
 \begin{align*}
  H_f(\lambda) &{}=% \begin{cases}
    (\lambda,1-\lambda), \text{ for }\lambda < \tfrac16 + \tfrac \tau 2, &
  H_f(\lambda) &{}=
    (\lambda,\tfrac12-\lambda) \orthocupT (\tfrac12+\lambda,\lambda),
    \text{ for } \tfrac16 + \tfrac \tau 2 \le \lambda < \tfrac13
  %\end{cases}
\end{align*}
if $\tau \in (0,\tfrac13)$ and  the resulting $\orthoT$-clustering is 
 $\set{ (0,1), (\tfrac16+\tfrac\tau2,\tfrac12-\tfrac\tau2),(\tfrac12+\tfrac\tau2,\tfrac56-\tfrac\tau2) }$.
 Finally, if $\tau\ge\tfrac13$ then all level sets are $\tau$-connected and the forest
is simply $\{(0,1)\}$.
In Table~\ref{table.examples.dim.one} more examples of clustering of densities can be found.

\begin{table}
\tikzstyle{background rectangle}=[color=white]%
\centering
\begin{tabular}{|c|c|c|c|c|} \hline 
\textbf{Name} & %\textbf{Density} & $(\clA_p, \orthoD)$ & $(\clA_p, \orthoT)$ \\ \hline\hline
Merlon & Camel & M & Factory \\\hline\hline
  \begin{tikzpicture}[baseline=.5cm,framed,scale=1]
  \node at (0,.2) {Density} ;
  \node at (0,-.5) {$(\clA_p, \orthoD)$} ;
  \node at (0,-1.25) {$(\clA_p, \orthoT)$ with $\tau$ small} ;
  \node at (0,-2) {$(\clA_p, \orthoT)$ with $\tau$ large} ;
\end{tikzpicture}
&
  \begin{tikzpicture}[baseline=.5cm,framed,scale=1]
  \draw[yscale=.75] (-.2,0) -- (0,0) -- (0,1) -- (1/3,1) -- (1/3,.5)
  -- (2/3,.5) -- (2/3,1) -- (1,1) -- (1,0) -- (1.2,0);
  \begin{scope}[yshift=-.5cm]
  \drawsetCC{-.2}01
  \drawsetCC{0}0{1/3}
  \drawsetCC{0}{2/3}1
  \end{scope}
  \begin{scope}[yshift=-1.25cm]
  \drawsetCC{-.2}01
  \drawsetCC{0}0{1/3}
  \drawsetCC{0}{2/3}1
  \end{scope}
  \begin{scope}[yshift=-2cm]
  \drawsetCC001
  \end{scope}
  \end{tikzpicture}
  &
  \begin{tikzpicture}[baseline=.5cm,framed,scale=1,domain=0:1]
  \draw[yscale=.75] (-.2,0) -- (0,0) .. controls (.2,0) and
     (.05,1) .. (.25,1) .. controls (.45,1) and
     (.3,.5) .. (.5,.5) .. controls (.7,.5) and
     (.55,1) .. (.75,1) .. controls (.95,1) and
     (.8,0) .. (1,0) -- (1.2,0)
     ;
  \begin{scope}[yshift=-.5cm]
  \drawsetOO001
  \drawsetOO{.1}{.1}{.5}
  \drawsetOO{.2}{.5}{.9}
  \end{scope}
  \begin{scope}[yshift=-1.25cm]
  \drawsetOO001
  \drawsetOO{.1}{.1}{.4}
  \drawsetOO{.2}{.6}{.9}
  \end{scope}
  \begin{scope}[yshift=-2cm]
  \drawsetOO001
  \end{scope}
\end{tikzpicture}
&
 \begin{tikzpicture}[baseline=.5cm,framed,scale=1,domain=0:1]
  \draw[yscale=.75] (-.2,0) -- (0,0) -- (0,1) -- (.5,.5) -- (1,1) -- (1,0) -- (1.2,0);
  \begin{scope}[yshift=-.5cm]
  \drawsetCC001
  \drawsetCO{.1}0{.5}
  \drawsetOC{.2}{.5}1
  \end{scope}
  \begin{scope}[yshift=-1.25cm]
%  \draw[very thick] (0,-.2) -- (1,-.2) (0,-.05) -- (.5,-.05) (.5,-0.1)--(1,-0.1);
  \drawsetCC001
  \drawsetCO{.1}0{.4}
  \drawsetOC{.2}{.6}1
  \end{scope}
  \begin{scope}[yshift=-2cm]
  \drawsetCC001
  \end{scope}
\end{tikzpicture} %&
 &
 \begin{tikzpicture}[baseline=.5cm,framed,scale=1,domain=0:1]
  \draw[yscale=.75] (-.2,0) -- (0,0) -- (0,1) -- (.5,.5) -- (.5,1) -- (1,1) -- (1,0) -- (1.2,0);
  \begin{scope}[yshift=-.5cm]
  \drawsetCC001
  \drawsetCO{.1}0{.5}
  \drawsetCC{.2}{.5}1
  \end{scope}
  \begin{scope}[yshift=-1.25cm]
  \drawsetCC001
  \drawsetCO{.1}0{.3}
  \drawsetCC{.2}{.5}1
  \end{scope}
  \begin{scope}[yshift=-2cm]
  \drawsetCC001
  \end{scope}
\end{tikzpicture} %&
  \\\hline
\end{tabular} 
\captionadd{Examples of clustering in dimension $d=1$ using $\clA_p$ and three separation relations.}
% {
% Here we always state the maximal achievable intervals.
% Remark that for large enough $\tau$ only the roots remain.
% }
\label{table.examples.dim.one}
\end{table}
 
\makeatletter{}\subsubsection{Dimension \texorpdfstring{$d=2$}{d=2}}
\label{subsec.designing.base.sets}

Our goal in this subsection is to 
understand the $\orthoG$-separated decomposition of closed events.
We further present the resulting clusterings for some densities that are indicator
functions and illustrate clusterings for continuous densities having a saddle point.

Let us begin by assuming that $P$ has a Lebesgue density of the form $1_B$, where 
$B$ is some   $\mu$-support set.
Then one can show, see Lemma~\ref{lemma.clustering.indicator.functions} for  details,
that adapted, isomonotone sequences $(F_n)$ of forests $F_n\uparrow B$ are of the form 
% 
% 
% suffices to consider approximating 
% forests $F=\{\rep{A}{k}\}$
% of pairwise disjoint base sets.
% Note that a 
%  sequence $(F_n)$ of such forests  is isomonotone iff all have the same number of elements and
% each forest can be enumerated 
$F_n=\{\rep{A^n}{k}\}$, where the elements of each forest $F_n$ are mutually disjoint
and can be ordered 
in such a way that
$A_i^1\subset A_i^2\subset\ldots$.
The limit forest $F_\infty$ then consists of the $k$ pairwise $\orthoG$-separated sets:
\[
  B_i := \bigcup_{n\geq 1} A_i^n\, ,
\]
and there is a $\mu$-null set $N\in\clB$ with
\begin{equation}\label{B-decomp}
  B = \rep[\orthocupG]Bk\orthocupG N.
\end{equation}

% Consequently,  a $\mu$-support set  $B$ can be approximated isomonotonically
% iff it is, up to a $\mu$-null set,  a finite $\orthoG$-disjoint union of sets in $\bar\clA$.
% See Remark~\ref{remark.clustering.indicator.functions} for more details.
Let us now consider the base sets $\clA_p$ in Example \ref{ex-Axx}.
By Proposition \ref{prop.neighborhood.base} we know that 
$\bar \clA_p$ contains all open, path-connected sets and therefore all
open $L^q$-balls.
Moreover,  all closed $L^q$-balls $B$ are $\mu$-support sets with $\mu(\partial B) = 0$.
Our initial consideration shows that $1_B$ can be approximated by an adapted, 
isomonotone sequence $(F_n)$ of forests of the form $F_n = \{A^n\}$ with $A^n\in \clA_p$.
However, depending on $p$ and $q$ the $\mu$-null set $N$ in \eqref{B-decomp} may differ.

Now that we have an understanding of $\bar\clA_p$ 
and adapted, isomonotone approximations
we can investigate some more
interesting cases and appreciate the influence of the choice of $\clA$ on the
outcome of the clustering in the following example.

\begin{example}[Clustering of indicators]
We consider 6 examples of $\mu$-support sets $B\in \dsR^2$.
The first 4 have two parts that only intersect at one point,
the second to last has two topological components, and the last one is connected in a fat way.
% and consider $\orthoD$.
%we consider the following compact events:
%\begin{align*}
%  \tikz[scale=.3,baseline=-.5ex] \filldraw[blue] (-1,-1) rectangle (0,0) rectangle (1,1) ;
%  &{}:= [-1,0]^2\cup[0,1]^2 %= \{ (x,y)\in [-1,1]^2 \mid x\cdot y \ge 0\}
%  \\
%  \tikz[scale=.3,baseline=-.5ex] \filldraw[blue,scale=.7071] (-2,0) -- (-1,1) -- (0,0) -- (1,1) -- (2,0) --
%  (1,-1) -- (0,0) -- (-1,-1) -- (-2,0) -- cycle ;
%  &{}:= \ball_1( \,(-r,0),r)\cup\ball_1( \,(r,0),r)
%  %=\{ (x,y)\in\Omega \mid |y| \in [ -|x|,|x|] \tand |x|+|y|\le \sqrt2 \}
%  \qquad r := \tfrac1{\sqrt2}
%\end{align*}
%Obviously, the second one is just a rotation of the first one by angle $\frac\pi4$.
By natural approximations we get
the clusterings of Table~\ref{table.indicators}.
\begin{table}[t]
\tikzstyle{every picture}+=[scale=.3,baseline=-.5ex,draw=blue,fill=blue]
\pgfmathsetmacro{\r}{sqrt(2)}\pgfmathsetmacro{\ir}{1/sqrt(2)*.5}
\begin{align*}
\newcommand{\filldot}[1]{\fill[red]  (#1) circle (.2);}
\begin{array}{c|c|c|c|c|c}
 & \clA_1 = \dsC(\tikz[scale=.5] \fill (-1,0)--(0,1)--(1,0)--(0,-1)--(-1,0) ;)
 & \clA_2 = \dsC(\tikz[scale=.5] \fill (0,0) circle (1) ;)
 & \clA_\infty = \dsC(\tikz[scale=.5] \fill (-1,-1) rectangle (1,1) ;)
 & \clAconv & \clA_2^\tau \\%& \clA_2^{-\tau},\clA_2^\circ \\
\hline
\tikz \filldraw (-1,-1) rectangle (0,0) rectangle (1,1) ;
  & \set{ \tikz{\fill (-1,-1) rectangle (0,0) ; \draw[dotted] (0,0) rectangle (1,1) ;
 			 \foreach \i in {-1,0} \foreach \j in {-1,0} \filldot{\i,\j} ; },
       \tikz{\fill (1,1) rectangle (0,0) ; \draw[dotted] (0,0) rectangle (-1,-1) ;
 			 \foreach \i in {1,0} \foreach \j in {1,0} \filldot{\i,\j} ; } }
  & \set{ \tikz{\fill (-1,-1) rectangle (0,0) ; \draw[dotted] (0,0) rectangle (1,1) ;
 			 \foreach \i in {-1,0} \foreach \j in {-1,0} \filldot{\i,\j} ; },
       \tikz{\fill (1,1) rectangle (0,0) ; \draw[dotted] (0,0) rectangle (-1,-1) ;
 			 \foreach \i in {1,0} \foreach \j in {1,0} \filldot{\i,\j} ; } }
  & \set{ \tikz \filldraw (-1,-1) rectangle (0,0) rectangle (1,1) ;}
  & \set{ \tikz \filldraw (-1,-1) rectangle (0,0) rectangle (1,1) ;}
  & \set{ \tikz{ \filldraw (-1,-1) rectangle (0,0) rectangle (1,1) ;
 			 \foreach \i in {-1,0} \foreach \j in {-1,0} \filldot{\i,\j} ; 
 			 \foreach \i in {1,0} \foreach \j in {1,0} \filldot{\i,\j} ; } }
  %& \set{ \tikz{\fill (-1,-1) rectangle (0,0) ; \draw[dotted] (0,0) rectangle (1,1) ;
 			 %\foreach \i in {-1,0} \foreach \j in {-1,0} \filldot{\i,\j} ; },
       %\tikz{\fill (1,1) rectangle (0,0) ; \draw[dotted] (0,0) rectangle (-1,-1) ;
 			 %\foreach \i in {1,0} \foreach \j in {1,0} \filldot{\i,\j} ; } }
\\
\tikz\filldraw[blue,scale=.5000] (-2,0) -- (-1,1) -- (0,0) -- (1,1) -- (2,0) --
  (1,-1) -- (0,0) -- (-1,-1) -- (-2,0) -- cycle ;
  & \set{ \tikz\filldraw[blue,scale=.5000] (-2,0) -- (-1,1) -- (0,0) -- (1,1) -- (2,0) --
  (1,-1) -- (0,0) -- (-1,-1) -- (-2,0) -- cycle ; }
  & \set{ \tikz[scale=.5000]{\fill (-1,-1) (-2,0) -- (-1,1) -- (0,0)  -- (-1,-1) -- cycle ;
							\draw[dotted] (0,0) -- (1,1) -- (2,0) -- (1,-1) -- cycle ;
 			 				\filldot{-2,0}\filldot{0,0}\filldot{-1,-1}\filldot{-1,1} },
       \tikz[scale=.5000]{\draw[dotted] (-1,-1) (-2,0) -- (-1,1) -- (0,0)  -- (-1,-1) -- cycle ;
                          \fill (0,0) -- (1,1) -- (2,0) -- (1,-1) -- cycle ;
 			 				\filldot{2,0}\filldot{0,0}\filldot{1,-1}\filldot{1,1} } }
  & \set{ \tikz[scale=.5000]{\fill (-1,-1) (-2,0) -- (-1,1) -- (0,0)  -- (-1,-1) -- cycle ;
                          \draw[dotted] (0,0) -- (1,1) -- (2,0) -- (1,-1) -- cycle ;
 			 				\filldot{-2,0}\filldot{0,0}\filldot{-1,-1}\filldot{-1,1} },
       \tikz[scale=.5000]{\draw[dotted] (-1,-1) (-2,0) -- (-1,1) -- (0,0)  -- (-1,-1) -- cycle ;
                          \fill (0,0) -- (1,1) -- (2,0) -- (1,-1) -- cycle ;
 			 				\filldot{2,0}\filldot{0,0}\filldot{1,-1}\filldot{1,1} } }
  & \set{ \tikz\filldraw[blue,scale=.5000] (-2,0) -- (-1,1) -- (0,0) -- (1,1) -- (2,0) --
  (1,-1) -- (0,0) -- (-1,-1) -- (-2,0) -- cycle ; }
  & \set{ \tikz[scale=.5000]{\filldraw[blue] (-2,0) -- (-1,1) -- (0,0) -- (1,1) -- (2,0) --
  (1,-1) -- (0,0) -- (-1,-1) -- (-2,0) -- cycle ;
  					\filldot{-2,0}\filldot{0,0}\filldot{-1,-1}\filldot{-1,1} 
  					\filldot{2,0}\filldot{0,0}\filldot{1,-1}\filldot{1,1}} }
  %& \set{ \tikz[scale=.5000]{\fill (-1,-1) (-2,0) -- (-1,1) -- (0,0)  -- (-1,-1) -- cycle ;
							%\draw[dotted] (0,0) -- (1,1) -- (2,0) -- (1,-1) -- cycle ;
 			 				%\filldot{-2,0}\filldot{0,0}\filldot{-1,-1}\filldot{-1,1} },
       %\tikz[scale=.5000]{\draw[dotted] (-1,-1) (-2,0) -- (-1,1) -- (0,0)  -- (-1,-1) -- cycle ;
                          %\fill (0,0) -- (1,1) -- (2,0) -- (1,-1) -- cycle ;
 			 				%\filldot{2,0}\filldot{0,0}\filldot{1,-1}\filldot{1,1} } }
\\
\tikz[scale=\ir]{\fill (-1,-1) circle(\r) (1,1) circle(\r) ;}
 & \set{ \tikz[scale=\ir]{\fill (-1,-1) circle(\r) ; \draw[dotted] (1,1) circle(\r) ;
 				\filldot{-1.9,-1.9}\filldot{-.1,-1.9}\filldot{-1.9,-.1}\filldot{-.1,-.1} } ,
         \tikz[scale=\ir]{\fill (1,1) circle(\r) ; \draw[dotted] (-1,-1) circle(\r) ;
 				\filldot{1.9,1.9}\filldot{.1,1.9}\filldot{1.9,.1}\filldot{.1,.1}} }
 & \set{ \tikz[scale=\ir]{\fill (-1,-1) circle(\r) (1,1) circle(\r) ;} }
 & \set{ \tikz[scale=\ir]{\fill (-1,-1) circle(\r) (1,1) circle(\r) ;
 			\filldot{-1,.4}\filldot{-1,-2.4}\filldot{-2.4,-1}\filldot{.4,-1}
 			\filldot{1,-.4}\filldot{1,2.4}\filldot{2.4,1}\filldot{-.4,1} } }
 & \set{ \tikz[scale=\ir]{\fill (-1,-1) circle(\r) (1,1) circle(\r) ;} }
 & \set{ \tikz[scale=\ir]{\fill (-1,-1) circle(\r) (1,1) circle(\r) ;} }
 %& \set{ \tikz[scale=\ir]{\fill (-1,-1) circle(\r) ; \draw[dotted] (1,1) circle(\r) ; } ,
         %\tikz[scale=\ir]{\fill (1,1) circle(\r) ; \draw[dotted] (-1,-1) circle(\r) ; } }
 \\
\tikz[scale=.5]{\fill (-1,0) circle(1) (1,0) circle(1) ;}
 & \set{ \tikz[scale=.5]{\fill (-1,0) circle(1) (1,0) circle(1) ;
 			\foreach \i in {-1.707,-.293,.293,1.707} \foreach \j in {.707,-.707}
 			\filldot{\i,\j} ; } }
 & \set{ \tikz[scale=.5]{\fill (-1,0) circle(1) (1,0) circle(1) ;} }
 & \set{ \tikz[scale=.5]{\fill (-1,0) circle(1) ; \draw[dotted] (1,0) circle(1) ;
 			\filldot{-2,0}\filldot{-1,1}\filldot{-1,-1}\filldot{0,0}},
         \tikz[scale=.5]{\fill (1,0) circle(1) ; \draw[dotted] (-1,0) circle(1) ;
 			\filldot{2,0}\filldot{1,1}\filldot{1,-1}\filldot{0,0} } }
 & \set{ \tikz[scale=.5]{\fill (-1,0) circle(1) (1,0) circle(1) ;} }
 & \set{ \tikz[scale=.5]{\fill (-1,0) circle(1) (1,0) circle(1) ;} }
 %& \set{ \tikz[scale=.5]{\fill (-1,0) circle(1) ; \draw[dotted] (1,0) circle(1) ; },
         %\tikz[scale=.5]{\fill (1,0) circle(1) ; \draw[dotted] (-1,0) circle(1) ; } }
 \\
\tikz \filldraw (-1,-1) rectangle (-.2,-.2) (.2,.2) rectangle (1,1) ;
  & \set{ \tikz{\fill (-1,-1) rectangle (-.2,-.2) ; \draw[dotted] (.2,.2) rectangle (1,1) ;
 			 \foreach \i in {-1,-.2} \foreach \j in {-1,-.2} \filldot{\i,\j} ; },
       \tikz{\fill (1,1) rectangle (.2,.2) ; \draw[dotted] (-.2,-.2) rectangle (-1,-1) ;
 			 \foreach \i in {1,.2} \foreach \j in {1,.2} \filldot{\i,\j} ; } }
  & \set{ \tikz{\fill (-1,-1) rectangle (-.2,-.2) ; \draw[dotted] (.2,.2) rectangle (1,1) ;
 			 \foreach \i in {-1,-.2} \foreach \j in {-1,-.2} \filldot{\i,\j} ; },
       \tikz{\fill (1,1) rectangle (.2,.2) ; \draw[dotted] (-.2,-.2) rectangle (-1,-1) ;
 			 \foreach \i in {1,.2} \foreach \j in {1,.2} \filldot{\i,\j} ; } }
  & \set{ \tikz{\fill (-1,-1) rectangle (-.2,-.2) ; \draw[dotted] (.2,.2) rectangle (1,1) ; },
       \tikz{\fill (1,1) rectangle (.2,.2) ; \draw[dotted] (-.2,-.2) rectangle (-1,-1) ; } }
  & \set{ \tikz{\fill (-1,-1) rectangle (-.2,-.2) ; \draw[dotted] (.2,.2) rectangle (1,1) ; },
       \tikz{\fill (1,1) rectangle (.2,.2) ; \draw[dotted] (-.2,-.2) rectangle (-1,-1) ; } }
  & \set{ \tikz{\fill (-1,-1) rectangle (-.2,-.2) (.2,.2) rectangle (1,1) ;
				\foreach \i in {-1,-.2} \foreach \j in {-1,-.2} \filldot{\i,\j} ;
				\foreach \i in {1,.2} \foreach \j in {1,.2} \filldot{\i,\j} ;
				\node[scale=.5,anchor=north west] at (0,0) {$<\tau$} ; } }
  %& \set{ \tikz{\fill (-1,-1) rectangle (-.2,-.2) ; \draw[dotted] (.2,.2) rectangle (1,1) ;
 			 %\foreach \i in {-1,-.2} \foreach \j in {-1,-.2} \filldot{\i,\j} ; },
       %\tikz{\fill (1,1) rectangle (.2,.2) ; \draw[dotted] (-.2,-.2) rectangle (-1,-1) ;
 			 %\foreach \i in {1,.2} \foreach \j in {1,.2} \filldot{\i,\j} ; } }
\\
\tikz[scale=.5]{\fill (-1,0) circle(1) (.7,0) circle(1) ;}
 & \set{ \tikz[scale=.5]{\fill (-1,0) circle(1) (.7,0) circle(1) ;
 			\foreach \i in {-1.707,-.293,-.007,1.407} \foreach \j in {.707,-.707}
 			\filldot{\i,\j} ; } }
 & \set{ \tikz[scale=.5]{\fill (-1,0) circle(1) (.7,0) circle(1) ;} }
 & \set{ \tikz[scale=.5]{\fill (-1,0) circle(1) (.7,0) circle(1) ;
 			\filldot{-2,0}\filldot{-1,1}\filldot{-1,-1}
 			\filldot{1.7,0}\filldot{.7,1}\filldot{.7,-1} } }
 & \set{ \tikz[scale=.5]{\fill (-1,0) circle(1) (.7,0) circle(1) ;} }
 & \set{ \tikz[scale=.5]{\fill (-1,0) circle(1) (.7,0) circle(1) ;} }
  %& ?/\set{ \tikz[scale=.5]{\fill (-1,0) circle(1) (.7,0) circle(1) ;} }
%  & \set{ \tikz{\fill (-1,-1) rectangle (0,0) ; \draw[dotted] (0,0) rectangle (1,1) ;
% 			 \foreach \i in {-1,0} \foreach \j in {-1,0} \filldot{\i,\j} ; },
%       \tikz{\fill (1,1) rectangle (0,0) ; \draw[dotted] (0,0) rectangle (-1,-1) ;
% 			 \foreach \i in {1,0} \foreach \j in {1,0} \filldot{\i,\j} ; } }
\\
\end{array}
\end{align*}
\caption{Clusterings of indicators.}\label{table.indicators}
\end{table}
\tikzstyle{every picture}+=[scale=.3,baseline=-.5ex,draw=blue,fill=blue]
The red dots indicate points which never are achieved by any approximation.
Observe how the geometry encoded in $\clA$ shapes the clustering.
Since $\clAconv$ and $\clA_2$ are invariant under rotation, they yield the same structure of clustering
for rotated sets. The classes 
$\clA_1$ and $\clA_\infty$ on the other hand are not rotation-invariant and therefore the clustering
depends on the orientation of $B$.

%We need to discuss the case marked with a question mark in the bottom right cell:
%\tikz[scale=.5]{\fill (-1,0) circle(1) (.7,0) circle(1) ;}
%and $\clA_2^{-\tau}$. The intersection of the two balls has some diameter $\tau_0$.
%So for $\tau\in(0,\tfrac{\tau_0}2]$ the clustering is as in the other cases.
%For $\tau>\tfrac{\tau_0}2$ on the other hand we have $B\notin\bar\clA_2^{-\tau}$ and 
%therefore our theory does not provide a clustering.
%\fixedtodo{Double check the last sentence.}
\end{example}

\makeatletter{}% \subsubsection{Densities in \texorpdfstring{$d\ge2$}{d>=2}}

After having familiarized ourselves with the clustering of indicator functions
we finally consider a continuous density that has a saddle point.

\begin{example}\label{example.2d.saddle.point}
On   $\Omega := [-1,1]^2$ consider the density $f:\Omega\to [0,2]$ given by 
% 
% 
% with the usual topology, denote by $\mu$ the Lebesgue measure on $\Omega$ and set:
$%\begin{align*}
  f(x,y) %&{}
  := x\cdot y + 1  
%  \qquad\qquad \qquad 
%\begin{tikzpicture}[scale=.4,baseline=1cm]
%\begin{axis}[
%	domain=-1:1,
%	xmax=1,
%	ymax=1,
%]
%\addplot3[surf,samples=4] {x*y+1};
%\end{axis}
%\end{tikzpicture}
$.%\end{align*}
Then we have the following $\orthoD$-decomposition of the level sets $H_f(\lambda)$ of $f$:
\begin{displaymath}
 H_f(\lambda)
 = 
 \begin{cases}
  \{ (x,y): xy > \lambda -1  \}& \mbox{ if } \lambda\!\in\! [0,1), \\
  {[-1,0)}^2 \dotcup {(0,1]}^2 & \mbox{ if } \lambda \!=\! 1,\\
  \{ (x,y): x< 0 \mbox{ and } xy > \lambda -1  \} \dotcup \{ (x,y): x> 0 \mbox{ and } xy > \lambda -1  \} 
  & \mbox{ if } \lambda\!\in\! (1,2).
 \end{cases}
\end{displaymath}
% 
% 
% 
% Then $H_f(\lambda)$ is the interior of a hyperbola $x\cdot y\ge\lambda-1$ for $\lambda<1$
% and a parabola $x\cdot y\ge\lambda-1$ for $\lambda>1$.
% For $\lambda=1$ it is the following disjoint union of open quadrants:
% \[
%   H_f(1) = \set{ x,y<0 } \dotcup \set{x,y>0} = {[-1,0)}^2 \dotcup {(0,1]}^2.
% \]
For $(\clA_p,\clQ^{\mu,\clA_p},\orthoD)$ %and $(\clAo_p,\clQ^{\mu,\clAo_p},\orthoD)$
the clustering forest  is therefore given by:
\tikzstyle{every picture}+=[scale=.3,baseline=-.5ex,draw=blue,fill=blue]
\[
  \set{ [-1,1]^2, [-1,0)^2 , (0,1]^2 }
  = \set{ \tikz \fill (-1,-1) rectangle (1,1) ; ,
		\tikz{\fill (-1,-1) rectangle (0,0) ; \draw[dotted] (0,0) rectangle (1,1) rectangle (-1,-1) ; },
		\tikz{\fill (1,1) rectangle (0,0) ; \draw[dotted] (0,0) rectangle (-1,-1) rectangle (1,1) ; }
  }.
\]
Moreover, for  $(\clA_2^\tau,\clQ^{\mu,\clA_2^\tau},\orthoT)$ the clustering forest looks like
$
%  \set{ [-1,1]^2, [-1,0)^2 , (0,1]^2 }
%  =
  \set{ \tikz \fill (-1,-1) rectangle (1,1) ; ,
		\tikz{\fill[out=0,in=90] (-1,-1) -- (-1,-.1) to (-.1,-1) -- (-1,-1) ; \draw[dotted,out=180,in=270] (1,1) -- (1,.1) to (.1,1) -- (1,1) rectangle (-1,-1); },
		\tikz{\fill[out=180,in=270] (1,1) -- (1,.1) to (.1,1) -- (1,1) ; \draw[dotted,out=0,in=90] (-1,-1) -- (-1,-.1) to (-.1,-1) -- (-1,-1) rectangle (1,1) ; }
  }.
$
\end{example}

% \todo{An example of two clustering bases with same $\clA$ and $\orthoA$
% but different $\clQ$ and the influence on the clustering\ldots}

\makeatletter{}\subsection{Hausdorff Measures}
\label{sub.examples.hausdorff}

So far we have only considered clusterings of Lebesgue absolutely continuous distributions.
In this subsection we provide some examples indicating that the developed theory
goes far beyond this standard example.
At first, lower dimensional base sets and their resulting clusterings are investigated. Afterwards we  discuss collections of base sets of different 
dimensions and provide clusterings for some measures that are not absolutely continuous to 
any Hausdorff measure.
For the sake of 
simplicity we will restrict our considerations to $\orthoD$-clusterings, but 
generalizations along the lines of the previous subsections are straightforward.

\subsubsection{Lower Dimensional Base Sets}

Let us begin by recalling that the $s$-dimensional Hausdorff-measure on $\clB$ is defined by 
% We denote by $\clH^s$ the Hausdorff-measure to parameter $s\in[0,d]$:
\begin{align*}
  \clH^s(B) &{}:= \lim_{\epsilon\to0} \inf \set{\sum_{i=1}^\infty (\diam(B_i))^s
  \mid B\subset \bigcup_i B_i \tand \forall i\in\dsN\colon \diam(B_i)\le \epsilon }\, .
\end{align*}
Moreover, the Hausdorff-dimension of a  $B\in\clB$ is the value $s\in [0,d]$ at which
$s\mapsto\clH^s(B)$ jumps from $\infty$ to $0$.
If $B$ has Hausdorff-dimension $s$, then $\clH^s(B)$ can be either  zero, finite, or infinite.
Hausdorff-measures are inner regular \citep[Cor.~2.10.23]{Federer} and $\clH^d$ equals the Lebesgue-measure up to a normalization factor.
For a reference on Hausdorff-dimensions and -measures we refer to \cite{Falconer-1993} and \cite{Federer}.
Recall that given a Borel set $C\subset\dsR^s$ a map
$\phi\colon C\to\Omega$ is \begriff{bi-Lipschitz}
iff there are constants $0<c_1,c_2<\infty$ s.t.
\[
  c_1 d(x,y) \le d(\phi(x),\phi(y)) \le c_2 d(x,y).
\]
\begin{lemma}\label{lemma.bilipschitz}
If $C$ is a Lebesgue-support set in $\dsR^s$ and $\phi\colon C\to\Omega$ is bi-Lipschitz
then $C':=\phi(C)$ has Hausdorff-dimension $s$ and it is an $\clH^s$-support set in $\Omega$.
\end{lemma}
Motivated by Lemma \ref{lemma.bilipschitz}, consider the following collection of  $s$-dimensional base sets in $\Omega$:
% Consider for any natural number $s$ with $0\le s\le d$ the collection of sets:
\[
  \clC_{p,s} := \set{ \phi(C)\subset\Omega \mid \text{$C$ is the closed unit $p$-ball in $\dsR^s$ and
  $\phi\colon C\to\Omega$ is bi-Lipschitz }}.
\]
Using the notation of Definition \ref{defi.intersection.graph} and Proposition \ref{prop.indicators}
we further write 
\[
  \clA_{p,s} := \dsC_{\orthoD}(\clC_{p,s}) \qqtext{and}
  \clQ^{p,s} := \clQ^{\clH^s,\clA_{p,s}}.
\]
By $\clA_0:=\set{\{x\}\mid x\in\Omega}$ we denote the singletons
and $\clQ_0$ the collection of Dirac measures.
Since continuous mappings of connected sets are connected,
$(\clA_{p,s},\clQ^{p,s},\orthoD)$ is a stable $\orthoD$-additive clustering base.
Remark that we take the union after embedding into $\dsR^d$
and therefore also crossings do happen, e.g.\ the cross $[-1,1]\times \{0\} \cup \{0\}\times [-1,1]\in\clA_{p,1}$.
Another possibility would be to embed $\clA_p$ via a set of transformations into $\dsR^d$.
Finally we confine the examples here only to integer Hausdorff-dimensions---it would be interesting
though to consider e.g.\ the Cantor set or the Sierpinski triangle.
The following example presents a resulting clustering of an $\clH^1$-absolutely continuous 
measure on $\dsR^2$.

\begin{example}[Measures on curves in the plane]\label{example.dimension.1.in.2}
On  $\Omega := [-1,1]^2$ consider the measure $P_1{}:=f\,d\clH^1$ whose density is given by 
\begin{align*}
%  f_2(x,y) &{}:= x\cdot y + 1 \qquad \forall x,y\in\Omega, &
%   P_1&{}:=f_1\,d\clH^1, &
  f(x,y) &{}:= \begin{cases}
    f_{\text{Merlon}}(x) & \text{if }x\ge0\tand y=0, \\
    f_{\text{Camel}}(t) & \text{if }x=-3^{2t-2}\tand y = 3^{-2t}, \\
    f_{\text{M}}(t) & \text{if }x=2^{2t-2}\tand y = -2^{-2t}\, .
  \end{cases}
\end{align*}
Here the densities and clusterings for the Merlon, the Camel and the M can be
seen in Table~\ref{table.examples.dim.one}.
So for $(\clA_{p,1},\clQ^{p,1},\orthoD)$ with any fixed $p\ge1$
the clustering forest of $P_1$ is given by:
\begin{align*}
  \cl(P_1) &{}= 
  \Bigg\{\begin{matrix}
  [0,1]\times\{0\}, [0,\tfrac13]\times\{0\}, [\tfrac23,1]\times\{0\}, \\
  g_1\big((0,1)\big),g_1\big((0.2,0.5)\big),g_1\big((0.5,0.8)\big), \\
  g_2\big([0,1]\big),g_2\big([0,0.5)\big),g_2\big((0.5,1]\big)\end{matrix} \Bigg\} 
  \qquad\qquad
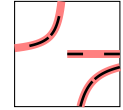
  \begin{tikzpicture}[baseline=0cm,scale=.7]
\tikzset{root1/.style={red!50,line width=3pt}}
\tikzset{child1/.style={black,line width=1pt}}
\draw[thin] (-1,-1) rectangle (1,1) ;  %% The box
\draw[root1] (0,0) -- (1,0) ;
\draw[child1] (0,-.0) -- +(.3,0) ;
\draw[child1] (1,.0) -- +(-.3,0) ;
\draw[root1] plot[domain=0:1,smooth,variable=\t] ({exp(ln(2)*(2*\t-2))},{-exp(-ln(2)*2*\t)});
\draw[domain=.001:.47,smooth,variable=\t,child1] plot ({exp(ln(2)*(2*\t-2))},{-exp(-ln(2)*2*\t)});
\draw[domain=.53:1,smooth,variable=\t,child1] plot ({exp(ln(2)*(2*\t-2))},{-exp(-ln(2)*2*\t)});
\draw[domain=0:1,smooth,variable=\t,root1] plot ({-exp(ln(3)*(2*\t-2))},{exp(-ln(3)*2*\t)});
\draw[domain=.15:.47,smooth,variable=\t,child1] plot ({-exp(ln(3)*(2*\t-2))},{exp(-ln(3)*2*\t)});
\draw[domain=.53:.85,smooth,variable=\t,child1] plot ({-exp(ln(3)*(2*\t-2))},{exp(-ln(3)*2*\t)});
\end{tikzpicture}
\end{align*}
where $g_i\colon[0,1]\to\Omega$ are given by $g_1(t) = (-3^{2t-2},3^{-2t})$ and $g_2(t) = (2^{2t-2},-2^{-2t})$.
\end{example}

\subsubsection{Heterogeneous Hausdorff-Dimensions}

In this subsection we consider measures that can be decomposed into
measures that are absolutely continuous with respect to Hausdorff measures
of different dimensions.
To this end, 
we write $\mu\prec\mu'$ 
for two measures $\mu$ and $\mu'$ on $\clB$,
iff for all $B\in\clB$ with $B\subset \supp \mu\cap \supp\mu'$ we have 
\begin{align*}
  \mu(B) < \infty \then \mu'(B) = 0.
\end{align*}
For $\clQ,\clQ'\subset\clM_\Omega$
we further write $\clQ\prec\clQ'$ if $\mu\prec\mu'$ for all $\mu\in\clQ$ and $\mu'\in\clQ'$.
Clearly, the relation $\prec$ is  transitive.
Moreover, we have $\clH^s\prec\clH^t$ whenever  $s<t$. 
The next proposition shows that clustering bases whose base measures dominate each other in the sense of 
$\prec$ can be merged.

\begin{prop}\label{prop.base.sets.aggregated}
%Let $\rep[\prec]\mu m$ be measures,
%assume $\clA^i\subset \clK(\mu_i)$ for all $i\le m$
Let $(\clA^1,\clQ^1,\orthoG),\ldots,(\clA^m,\clQ^m,\orthoG)$ be 
stable clustering bases sharing the same separation relation $\orthoG$
and assume $\clQ^1\prec \dots\prec \clQ^m$.
% \margincomment{$\ortho{\clA^i}$?}
We define
\[
  \clA:=\bigcup_i \clA^i \qquad \tand \qquad \clQ := \bigcup_i\clQ^i.
\]
Then $(\clA,\clQ,\orthoG)$ is a stable clustering base.
% \begin{enumerate}
% \item For all $A\in\clA$ there is a unique index $i(A)$ with $A\in \clA_{i(A)}$.
% In particular $\clQ$ is a family indexed by $\clA$ and we can still write $Q_A\in\clQ$ for all $A\in\clA$.
% \item If $A,A'\in\clA$ and $A\subset A'$ then $i(A)\le i(A')$.
% \item $\orthoG$ is a stable $\clA$-separation relation.
% \item $(\clA,\clQ,\orthoG)$ is a stable clustering base.
% \end{enumerate}
\end{prop}
% \begin{myproof*}[Proposition]{prop.base.sets.aggregated}
% \begin{enumerate}
% \item Let $A\in\clA$. Then there is $i\le m$ with $A\in\clA_i$.
% Let $\mu\in\clQ_i$ be the corresponding base measure with $\supp \mu=A$.
% Let $j\le m$ and $\mu'\in\clQ_j$ be another measure with $\supp \mu'=A$.
% Then $\mu(A)=1$ and $\mu'(A)=1$.
% If $j>i$ then by assumption $\mu\prec \mu'$ and this would give $\mu'(A)=0$.
% If $j<i$ we have $\mu'\prec \mu$ and this would give $\mu(A)=0$.
% So $i=j$.
% 
% \item Let $A\in\clA_i$ and $A'\in\clA_j$ with $A\subset A'$.
% Now $A=A\cap A'=\supp Q_{A}\cap\supp Q_{A'}$.
% If we had $i>j$ then $Q_{A'}\in\clQ_j\prec \clQ_i\ni Q_{A}$ and
% since $Q_{A'}(A)\le Q_{A'}(A')=1<\infty$ we would have $Q_A(A)=0$.
% Therefore $i\le j$.
% 
% \item We have to show $\clA$-stability and $\clA$-connectedness.
% The former follows since $i(A_n)$ is monotone if $\iseq\subset A$ by (b)
% and hence eventually is constant.
% For the latter let $A\in\clA_i$ and $\rep Bk\in\clB$ closed with
% $A\subset \rep[\orthocupG]Bk$. Then since $\orthoG$ is an $\clA_i$-separation relation there
% is $j\le k$ with $A\subset B_j$.
% 
% \item Fittedness is inherited from the individual clustering bases.
% Let $A\in\clA_i$ and $A'\in\clA_j$ with $A\subset A'$. Then $i\le j$ by (b).
% If $i=j$ then flatness follows from flatness of $\clA_i$.
% If $i<j$ then by assumption $Q_A\prec Q_{A'}$ and because $Q_A(A)=1<\infty$
% we have $Q_{A'}(A) = 0$.
% \qedhere
% \end{enumerate}
% \end{myproof*}%
%

Proposition \ref{prop.base.sets.aggregated}  shows that 
the $\orthoD$-additive, stable    bases $(\clA_{p,s},\clQ^{p,s},\orthoD)$ on $\dsR^d$
can be merged. Unfortunately, however, its union is no longer $\orthoD$-additive,
and therefore we need to investigate $P$-subadditivity in order to describe 
distributions for which our theory provides a clustering.
This is done in the next proposition.%\fixedtodo{double check its reformulation}

\begin{prop}\label{prop.unions.are.good}
Let $(\clA^1,\clQ^1,\orthoG)$ and $(\clA^2,\clQ^2,\orthoG)$ be clustering bases with 
$\clQ^1 \prec \clQ^2$ and $P_1$ and $P_2$ be finite measures with 
$P_1\prec\clA^2$ and $\clA^1\prec P_2$. Furthermore, assume that 
$\clA^i$ is $P_i$-subadditive for both $i=1,2$
and let $P:=P_1 + P_2$. Then we have
% 
% 
% Assume $\clA^1\prec\clA^2$ and let $\clA:=\clA^1\cup\clA^2$ with $\clQ:=\clQ^1\cup\clQ^2$.
% Now let $P_1,P_2$ be finite measures with $P_1\prec\clA^2$ and $\clA^1\prec P_2$.
% Assume $\clA^i$ is $P_i$-subadditive for both $i=1,2$
% and let $P=P_1 + P_2$.
% Then:
\begin{enumerate}
\item For $i=1,2$ and all base measures $\a\in\clQ^i_P$ we have  $\a\le P_i$,
\item If for all base measures $\a\in\clQ^2_{P_2}$ and $ \supp P_1  \northoG \supp \a$
there exists a base measure $\tilde \a\in\clQ^2_{P_2}(\supp P_1)$ with $\a\leq \tilde \a$ 
then $\clA^1\cup\clA^2$ is $P$-subadditive.
\end{enumerate}
\end{prop}

To illustrate   condition  \emph{(b)} consider 
clustering   bases $(\clA_{p,s},\clQ^{p,s},\orthoD)$
and  $(\clA_{p,t},\clQ^{p,t},\orthoD)$ for some $s<t$.
The condition specifies that any such base measure $\a$
intersecting $\supp P_1$ can be majorized by one which
supports $\supp P_1$.
Then all parts of $\supp P_1$ intersecting at least one
component of $\supp P_2$ have to be on the same niveau line of $P_2$.
Note that this is trivially satisfied if the $\supp P_1 \cap \supp P_1 = \emptyset$.
Recall that mixtures of the latter form have already been clustered in 
\cite{RinaldoWassermann-2010} by a  kernel smoothing approach. Clearly, our axiomatic approach makes
it possible to define clusterings for significantly more involved distributions
as the following two examples demonstrate.
 
% where only disjoint supports and only kernel-smoothened measures were considered
% so that the final probability measure was absolutely continuous to Lebesgue(?).

\begin{example}[Mixture of atoms and full measure]
Consider $\Omega=\dsR$.
Let $(\clA_0,\clQ_0,\orthoD)$  be the singletons with Dirac measures and consider
for any fixed $p\ge 1$ the clustering base $(\clA_{p,s},\clQ_{p,s},\orthoD)$.
%\fixedtodo{Notation $\clQ_1,\orthoD$ does not fit to previous notation, see also Sec. 4.1}
Both are $\northoD$-additive and stable and we have $\clQ_0\prec\clQ_{p,1}$.
% Trivially $\clA_0$ is $\northoD$-additive, and $\clA_1$ is by definition.
% and hence $P$-subadditive for all $P$.
% Furthermore 
Now consider the measures
\[
  P_0 {}:= \delta_0 + 2\delta_1 + \delta_2  \qquad  \mbox{ and } \qquad 
  P_1(dx) {}:= \sin^2(\tfrac{2x}\pi)\,\clH_1(dx) .
\]
Then the assumptions of Proposition \ref{prop.unions.are.good} are satisfied and the clustering of 
$P:= P_0+P_1$ is given by 
% Then $(P_0,\clA_0,\clQ_0,\orthoD),(P_1,\clA_1,\clQ_1,\orthoD)$ are compatible.
% Hence
\begin{align*}
  \cl(P)&{} = \cl(P_0)\cup\cl(P_1)
  = \set{ \{0\},(0,\tfrac12),(\tfrac12,1),\{1\},\{2\} }\, .
\end{align*}
\end{example}

Our last example combines Examples~\ref{example.2d.saddle.point} and~\ref{example.dimension.1.in.2}.
% For its formulation oberve that the niveau lines are given by $x\cdot y=\lambda-1$.

\begin{example}[Mixtures in dimension 2]\label{example.mixed.dimension.0.1.2}
Consider  $\Omega := [-1,1]^2$ and the densities $f_1$ and $f_2$ introduced in Examples 
\ref{example.dimension.1.in.2} and \ref{example.2d.saddle.point}, respectively.
Furthermore, consider the measures 
\begin{align*}
  P_2&{}:=f_2\,d\clH^2, &
  P_1&{}:=f_1\,d\clH^1
%   &
%   P_d0&{}:= \delta_{p_1}+\ldots+\delta_{p_n}.
\end{align*}
and the clustering bases $(\clA_{p,1},\clQ^{p,1},\orthoD)$ and $(\clA_{p',2},\clQ^{p',2},\orthoD)$ for some fixed $p,p'\ge1$.
As above $\clQ^{p,1}\prec\clQ^{p',2}$.
% Then with obvious choices for $\clA_0,\clA_1,\clA_2$ and $\clQ_i:=\clQ_{\clH^i}(\clA_i)$
% and using $\orthoD$ we see that everything is compatible.
And by Proposition~\ref{prop.unions.are.good} the clustering forest of $P=P_1+P_2$ is given by
\begin{align*}
  \cl_1(P_1) \cup \cl_2(P_2) &=
  \Bigg\{\begin{matrix} &
  [0,1]\times\{0\}, [0,\tfrac13]\times\{0\}, [\tfrac23,1]\times\{0\}, \\&
  g_1\big((0,1)\big),g_1\big((0.2,0.5)\big),g_1\big((0.5,0.8)\big), \\&
  g_2\big([0,1]\big),g_2\big([0,0.5)\big),g_2\big((0.5,1]\big), \\&
  [-1,1]^2, [-1,0)^2 , (0,1]^2
  \end{matrix}\Bigg\}
  &
  \begin{tikzpicture}[baseline=0cm,scale=1]
\fill[blue!20] (-1,-1) rectangle (1,1);
\fill[blue] (-1,-1) rectangle (-.0,-.0) (1,1) rectangle (.0,.0) ;
\tikzset{root1/.style={red!20,line width=6pt}}
\tikzset{child1/.style={red,line width=2pt}}
\draw[thin] (-1,-1) rectangle (1,1) ;  %% The box
\draw[root1] (0,0) -- (1,0) ;
\draw[child1] (0,-.0) -- +(.3,0) ;
\draw[child1] (1,.0) -- +(-.3,0) ;
\draw[domain=0:1,smooth,variable=\t,root1] plot ({exp(ln(2)*(2*\t-2))},{-exp(-ln(2)*2*\t)});
\draw[domain=0:.47,smooth,variable=\t,child1] plot ({exp(ln(2)*(2*\t-2))},{-exp(-ln(2)*2*\t)});
\draw[domain=.53:1,smooth,variable=\t,child1] plot ({exp(ln(2)*(2*\t-2))},{-exp(-ln(2)*2*\t)});
\draw[domain=0:1,smooth,variable=\t,root1] plot ({-exp(ln(3)*(2*\t-2))},{exp(-ln(3)*2*\t)});
\draw[domain=.15:.47,smooth,variable=\t,child1] plot ({-exp(ln(3)*(2*\t-2))},{exp(-ln(3)*2*\t)});
\draw[domain=.53:.85,smooth,variable=\t,child1] plot ({-exp(ln(3)*(2*\t-2))},{exp(-ln(3)*2*\t)});
%
%\pgfmathsetseed{123}
%\foreach \i in {1,2,3,4,5,6} \fill[green] (2*rnd-1,2*rnd-1) circle (2pt);
%\fill[green] (.4,0) circle (2pt);
%\fill[green] (-.4,) circle (2pt);
%
\end{tikzpicture}
\end{align*}
where $g_i\colon[0,1]\to\Omega$ are given by $g_1(t) = (-3^{2t-2},3^{-2t})$ and $g_2(t) = (2^{2t-2},-2^{-2t})$. Observe that $g_1$ and $g_2$ lie on niveau lines of $f_2$.
%See also Figure~\ref{figure.mixed.dimension.0.1.2}
\end{example}

\makeatletter{}\section{Proofs}\label{sec:proofs}

\makeatletter{}

\subsection{Proofs for Section \ref{section.additive.clustering}}

%$\orthoG$-connectedness usually hold even for $\bar\clA$-sets?
We begin with some simple properties  of separation relations.

\begin{lemma}\label{lemma.separation.base.2}
Let $\orthoG$ be an $\clA$-separation relation. 
% Recall that $\emptyset \notin \clA$ by definition. 
Then the following statements are true:
\begin{enumerate}
\item %If $A\orthoG A'$ and $A\cap A'\in\clK$ then $A\cap A'=\emptyset$.
  For all $B,B'\in \clB$ with $B\orthoG B'$ we have  $B\cap B'=\emptyset$.
\item Suppose that  $\orthoG$ is stable and  $(A_i)_{i\geq 1}\subset \clA$ is increasing.
% $\iseq\subset A$. 
For  $A:=\bigcup_n A_n$ and  all 
$B\in\clB$ we then have 
\[
  A_n\orthoG B \quad \text{ for all } n\geq 1 \iff A \orthoG B
\]
\item Let $A\in\clA$ and $\rep Bk\in\clB$ be closed. Then:
\[
  A\subset \rep[\orthocupG]Bk \then \exists! i\le k\colon A\subset B_i
\]
\item %\begriff{Uniqueness}: 
For all $\rep Ak\in\clA$ and all $\rep{A'}{k'}\in\clA$, we have
\[
  \rep[\orthocupG] Ak = \rep[\orthocupG]{A'}{k'} \then
  \{ \rep Ak \} = \{\rep{A'}{k'}\}.
\]
%where $\rep[\orthocupA] Ak$ denotes the union together with the condition that
%$A_i\orthoG A_j$ for all $i\ne j$.
% \item If $\clA'\subset\clA$ then
% $\orthoG$ is also an $\clA'$-separation relation.
\end{enumerate}
\end{lemma}

\begin{proofof}{Lemma \ref{lemma.separation.base.2}}
\ada a	Let us write $B_0:=B\cap B'$. Monotonicity and $B\orthoG B'$ implies $B_0\orthoG B'$ 
	and thus $B'\orthoG B_0$ by symmetry. Another application of the monotonicity gives $B_0\orthoG B_0$ 
	and the reflexivity thus shows $B\cap B'=B_0=\emptyset$.
	
\ada b ``$\Rightarrow$'' is stability and ``$\Leftarrow$'' follows from monotonicity.

\ada c  Existence of such an $i$ is $\clA$-connectedness.
  Now assume that there is an $j\neq i$ with $A\subset B_j$.
Then $\emptyset\ne A\subset B_i\cap B_j$
  contradicting $B_i\orthoG B_j$ by (a).
  
\ada d We write $F:=\{ \rep Ak \}$ and $F':=\{\rep{A'}{k'}\}$.
  By (c) we find an injection $I\colon F\to F'$ such that\ $A\subset I(A)$
  and hence $k\le k'$. Analogously, we find an injection
  $J\colon F'\to F$ such that\ $A\subset J(A)$, and we get $k=k'$.
  Consequently, $I$ and $J$ are bijections. Let us now fix an $A_i\in F$. For $A_j:=J\circ I(A_i)\in F$
  we then find 
  $A_i\subset I(A_i) \subset J(I(A_i))=A_j$. This implies $i=j$, since otherwise $A_i\subset A_j$ would contradict $A_i\orthoG A_j$
  by (a). Therefore we find $A_i=I(A_i)$ and the bijectivity of $I$ thus yields the assertion.
%  
%   
% \ada e  This is trivial from the definition.
\end{proofof}

\begin{proofof}{Proposition \ref{prop.indicators}}
We first need to check that the support is defined for all restrictions 
$\mu_{|C}:=\mu(\cdot\cap C)$ to sets $C\in\clB$ that satisfy $0<\mu(C)<\infty$.
To this end, we check that $\mu_{|C}$ is inner regular:
If $\Omega$ is a Radon space then there is nothing to prove since $\mu_{|C}$ is a finite measure.
% 
% 
% First we have to remark that $\clK(\mu)$ is well-defined:
% Let $C\in\clB$ with $0<\mu(C)<\infty$ and set $\mu_{|C}:=\mu(\cdot\cap C)$.
% Then $\mu_{|C}$ is a finite measure and we need to show that it is inner regular, so that its support is well-defined.
% If $\Omega$ is a Radon space then by definition of Radon Space the finite measure $\mu_{|C}$ is inner regular.
If $\Omega$ is not a Radon space, then the definition of $\clM^\infty_\Omega$ guarantees that $\mu$ is inner regular
and hence $\mu_{|C}$ is inner regular by Lemma~\ref{lemma.support.general}.

% Since every $C \in \clK(\mu)$ is the support of a measure, $\clK(\mu)$ consists of closed and non-empty sets.
Let us now verify that $(\clA,\clQ^{\mu,\clA},\orthoG)$ is a (stable) clustering base.
To this end, we first observe that each $Q_A\in\clQ^{\mu,\clA}$ is a probability measure by construction and since we have already seen that 
$\mu_{|C}$ is inner regular for all $C\in \clK(\mu)$ we conclude that $\clQ^{\mu,\clA}\subset \clM$.
Moreover, fittedness follows from $\clA\subset \clK(\mu)$.
% 
% 
% The $Q_A$ inherit the inner regularity of $\mu$ and are by construction probability measures.
% Fittedness follows from $\clA\subset\clK(\mu)$.
For flatness let $A,A'\in\clA$ with $A\subset A'$ and $Q_{A'}(A) \neq 0$. Then for all $B\in\clB$ we have
\[
  Q_A(B) 
  = \frac{\mu(B\cap A)}{\mu(A)}
%   = \frac{\mu(B\cap A)}{\mu(A\mid A')\cdot \mu(A')}
  = \frac{\mu(B\cap A\cap A')}{\mu(A\mid A')\cdot \mu(A')}
  = \frac{\mu(B\cap A\mid A')}{\mu(A\mid A')}
  = \frac{Q_{A'}(B\cap A)}{Q_{A'}(A)}
  .
  \qedhere
\]
\end{proofof}

\begin{proofof}{Lemma \ref{lemma.simple.measures.unique}}
Let $Q = \sum_{A\in F} \alpha_{A}Q_{A}$ and $Q = \sum_{A'\in F'} \alpha'_{A'}Q_{A'}$ be two representations of $Q\in \clQ$.
By  part (d) of Lemma~\ref{lemma.support.general} we then obtain
\[
  \supp Q = \supp\left(\sum_{A\in F} \alpha_AQ_A\right)
  = \bigcup_{A\in F} \supp Q_A = \bigcup_{A\in F} A\, ,
  = \Gr F
\]
and since we analogously  find  $\supp Q =  \Gr F'$, we conclude that $\Gr F=\Gr F'$.
The latter together with Lemma \ref{lemma.separation.base.2} gives $\max F = \max F'$.
%%Assume $F$ has more than one root, hence $F'$ has as well, and 
%We show that $\alpha_A=\alpha'_A$ for all roots $A\in \max\ F=\max F'$.
%So assume that $F$ and $F'$ are trees with the same root $A$ and $\alpha_A<\alpha'_A$.
%By  Lemma \ref{lemma.separation.base.2} we know that $A$ is not the union of its children $\rep Ak$ in $F$.
%The set $B:=A\setminus(\rep[\cup]Ak)$ is relatively open in $A=\supp Q_A$
To show that $\alpha_A=\alpha'_A$ for all roots $A\in \max\ F=\max F'$,
we pick a root $A\in \max\ F$  and assume that $\alpha_A<\alpha'_A$.
Now, if $A$ has no direct child, we set $B:=A$. Otherwise we 
define $B:=A\setminus(\rep[\cup]Ak)$, where the $A_k$ are the direct children of $A$ in $F$. 
Because of the definition of a direct child % if $k=1$ 
and part (d) of Lemma \ref{lemma.separation.base.2} we find
$\rep[\cup]Ak \subsetneqq A$ in the second case. In both cases we conclude that  
$B$ is non-empty and relatively open in $A=\supp Q_A$
and   
by Lemma~\ref{lemma.support.general} we obtain $Q_A(B)>0$. 
Consequently, our assumption  $\alpha_A<\alpha'_A$ yields $ \alpha_AQ_A(B) < \alpha'_A Q_A(B) \le Q(B)$.
However, our construction also gives
\[
 Q(B) = \sum_{A''\in F}\alpha_{A''}Q_{A''}(B)
= \alpha_AQ_A(B) + \sum_{A''\subsetneqq A}\alpha_{A''}Q_{A''}(B) + \sum_{A''\orthoG A}\alpha_{A''}Q_{A''}(B)
= \alpha_AQ_A(B)  \, ,
\]
i.e.~we have found a contradiction. Summing up, we already know that   $\max\ F=\max F'$ and 
$\alpha_A=\alpha'_A$ for all  $A\in \max\ F$. This yields
\[
   \sum_{A\in\max F}\alpha_A Q_A = \sum_{A'\in\max F'}\alpha'_{A'} Q_{A'}\, .
\]
Eliminating the roots gives the 
 forests $F_1:= F\setminus \max F$ and $F_1':= F'\setminus \max F'$
and
\[
   Q_1 
   := \sum_{A\in F_1} \alpha_{A}Q_{A} 
   = Q - \sum_{A\in\max F}\alpha_A Q_A 
   = Q - \sum_{A'\in\max F'}\alpha'_{A'} Q_{A'}
   = \sum_{A'\in F'_1} \alpha'_{A'}Q_{A'}\, ,
\]
i.e.~$Q_1$ has two representations based upon the reduced forests $F_1$ and $F_1'$. Applying the argument above recursively 
thus yields $F=F'$ and $\alpha'_A=\alpha'_{A}$ for all $A\in F$.
% % 
% % As a consequence, it follows
% % \begin{align*}
% % 0 < Q(B) &{}= \sum_{A''\in F}\alpha_{A''}Q_{A''}(B)
% % = \alpha_AQ_A(B) + \sum_{A''\subsetneqq A}\alpha_{A''}Q_{A''}(B) + \sum_{A''\orthoG A}\alpha_{A''}Q_{A''}(B)
% % \\&{}= \alpha_AQ_A(B) < \alpha'_A Q_A(B) \le Q'(B)
% % \end{align*}
% % which contradicts $Q=Q'$. Therefore, since $A$ was arbitrary chosen, $\alpha_A=\alpha'_A$ for all $A$ and hence:
% \[
%   \sum_{A\in\max F} Q\Rl A
%   = Q-\sum_{A\in\max F}\alpha_A Q_A
%   = Q-\sum_{A\in\max F'}\alpha'_A Q_A
%   = \sum_{A\in\max F'} Q\Rl A.
% \]
% We proceed recursively $F=F'$ and $\alpha'_A=\alpha'_{A'}$ for $A=A'$.
\end{proofof}

\begin{proofof}{Theorem \ref{thm.clustering.of.simple.measures}}
We first show that \eqref{def-uniq-add-clust} defines an additive clustering.
Since Axiom~\ref{axiom.base.clustering} is obviously satisfied, 
it suffices to check the two additivity axioms for $\clP:= \clS(\clA)$. We begin by establishing 
DisjointAdditivity. To this end, we pick 
 $\rep Qk\in \clS(\clA)$ with representing $\orthoG$-forests $F_i$
 such that  $\supp Q_i=\Gr F_i$ are mutually $\orthoG$-separated.
 For $A\in\max F_i$ and $A'\in\max F_j$ with $i\neq j$, we then have $A\orthoG A'$, and therefore
%  
% Since $Q_i$ is simple on $F_i$ and by assumption the $\supp Q_i=\Gr F_i$ are pairwise $\orthoG$-separated,
% also $A\in\max F_i$ and $A'\in\max F_j$ are $\orthoG$-separated and it follows that:
\[
  F := \rep[\cup] Fk
\]
is the representing $\orthoG$-forest of $Q:=\rep[+]Qk$. This gives $Q\in \clS(\clA)$ and 
\[
  \cl(Q) = s(F) = s(F_1)\cup\dots\cup s(F_k) = \cl(Q_1)\cup\dots\cup \cl(Q_k).
\]
To check BaseAdditivity we fix a $Q\in\clS(\clA)$ with representing $\orthoG$-forest $F$
and a base measure $\a = \ca Q_A$ with 
$\supp Q\subset \supp \a$.
For all $A'\in F$ we then have $A'\subset \Gr F=\supp Q\subset A$ and therefore
$F':=\{A\}\cup F$ is the representing $\orthoG$-forest of $\a+Q$.
This yields $\a+Q\in\clS(\clA)$ and
\[
  \cl(\a+Q) = s(F') = s\big( \{A\} \cup F \big)
  = s\big( \supp \a \cup \cl(Q) \big).
\]

Let us now show that every 
additive $\clA$-clustering $\cl: \clP \rightarrow \clF$
satisfies both $\clS(\clA) \subset \clP$ and \eqref{def-uniq-add-clust}.
To this end we pick a $Q\in \clS(\clA)$ with representing forest $F$
and show by induction over $|F|=n$ that both  $Q\in\clP$ and $\cl(Q)=s(F)$.
Clearly, for  $n=1$ this immediately follows from Axiom \ref{axiom.base.clustering}.
For the induction step we assume that for some $n\geq 2$ we have already proved
$Q'\in\clP$ and $\cl(Q')=s(F')$ for all 
$Q'\in\clS(\clA)$ with representing forest $F'$
of size $|F'|<n$. 

Let us first consider the case in which $F$ is a tree. Let $A$ be its root and $\alpha_A$ be corresponding 
coefficient in the representation of $Q$. Then $Q':=Q-\alpha_AQ_A$ is a simple measure with 
representing forest $F':=F\setminus A$ and since $|F'|=n-1$ we know 
$Q'\in\clP$ and $\cl(Q')=s(F')$ by the induction assumption.
By the axiom of BaseAdditivity we conclude that 
\[
  c(Q) = c(\alpha_AQ_A+Q') = s(\{A\}\cup c(Q')) = s(\{A\}\cup F')
  = s(F)\, ,
\]
where the last equality follows from the assumption that  $F$ is a tree with root $A$.

Now consider the case where $F$ is a  forest with $k\ge2$ roots $\rep Ak$. 
For $i\leq k$ we define  $Q_i:=Q\Rle{A_i}$.
Then all $Q_i$ are simple measures with
representing
forests $F_i := F\Rle{A_i}$ and we have $Q=Q_1+\dots+Q_k$.
Therefore, the induction assumption guarantees 
$Q_i\in\clP$ and $\cl(Q_i)=s(F_i)$.
Since $\supp Q_i = A_i$ and 
$A_i \orthoG A_j$ whenever $i\neq j$, the axiom of DisjointAdditivity then shows $Q\in \clP$ and 
% 
% 
% Clearly these roots are pairwise $\orthoG$-separated, i.e. $A_i \orthoG A_j$ if $i\neq j$.
% Let $\rep\a k$ be the levels of the roots and denote by $Q_i:=Q\Rle{A_i}$.
% Since the $\rep Ak$ are pairwise $\orthoG$-separated we have by the axiom of DisjointAdditivity and induction assumption:
\[
  \cl(Q) 
%   = \cl\left(\sum_i Q_i\right)
  = \cl(Q_1) \cup\dots\cup \cl(Q_k)
  =  s(F_1) \cup\dots\cup s(F_k)
  = s(F)\, .
  \qedhere
\]
\end{proofof}

\makeatletter{}\makeatletter{}\subsection{Proofs for Section \ref{section.continuous.clustering}}

\begin{proofof}{Lemma \ref{lemma.limit.is.forest}}
For the first assertion it suffices to 
 check 
 $\bar\clA$-connectedness. To this end, we fix an $A\in \bar\clA$ and 
 closed sets  $\rep Bk$ with $A\subset \rep[\orthocupG]Bk$. Let $(A_n)\subset \clA$ with 
  $A_n\nearrow A$. For all $n\geq 1$ part (c) of Lemma \ref{lemma.separation.base.2} then gives 
   exactly one $i(n)$ with $A_n\subset B_{i(n)}$. This uniqueness  together with 
    $A_n\subset A_{n+1}$ yields  $i(1)=i(2)=\dots$  and hence $A_n\subset B_{i(1)}$ for all $n$.
    We conclude that $A\subset B_{i(1)}$ by    part (b) of Lemma \ref{lemma.separation.base.2}.
    
%     are the same, and hence
%   $A_n\subset B_{i(1)}$ for all $n$ implying $A\subset B_{i(1)}$.
%   $\bar\clA$-uniqueness then follows from this.

For the second assertion we pick an isomonotone sequence $(F_n)\subset \clF_\clA$ and define $F_\infty:=\lim_n s(F_n)$.
Let us first show that $F_\infty$ is a $\orthoG$-forest. To this end, we pick $A,A'\in F_\infty$. 
By the 
construction of $F_\infty$ there then exist $A_1,A_1'\in s(F_1)$ such that for $A_n:= \zeta_n(A_1)$ and 
$A_n':= \zeta_n(A_1')$ we have 
 $A_n\nearrow A$ and $A'_n\nearrow A'$ 
Now, 
%  
%  be two nodes with the respective approximations.
% It is clear from the definition that $A,A'\in\bar\clA$.
% We show that $F_\infty$ is a forest.
 if $A_1\orthoG A_1'$ then $A_n\orthoG A_n'$ %by isomonotonicity
and thus 
 $A_m\orthoG A_n'$ for all $m,n$ by isomonotonicity. Using the stability of $\orthoG$ twice we first obtain 
  $A\orthoG A_n'$ for all $n$ and then  $A\orthoG A'$.
  If $A_1\not\orthoG A_1'$, we may assume  $A_1\subset A_1'$ since $s(F_1)$ is a $\orthoG$-forest.
  Isomonotonicity implies $A_n\subset A_n' \subset A'$ for all $n$ and hence 
  $A\subset A'$. Finally, $s(F_n)\le F_\infty$  is trivial.
\end{proofof}

\begin{proofof}{Proposition \ref{prop.additive.implies.subadditive}}
We first show  that $\clA$ is $P$-subadditive if 
$\clA$ is $\nPorthoP$-additive. To this end we fix 
$A,A'\in\clA$ with $A\nPorthoP A'$. Since 
$\clA$ is $\nPorthoP$-additive we find $B:=A\cup A'\in \clA$.
This yields 
\[
  Q_B(A)=\frac{\mu(A\cap B)}{\mu(B)} = \frac{\mu(A)}{\mu(B)}>0
\]
and analogously we obtain  $Q_B(A')>0$.
For $\ca Q_A,\ca' Q_{A'}\le P$ we can therefore assume
that  $\cb:=\frac{\ca}{Q_B(A)} < \frac{\ca'}{Q_B(A')}$. Setting $\b:=\cb Q_B$ we now obtain by the flatness assumption
\[
  \ca Q_A(\cdot) = \ca\cdot  \frac { Q_B(\cdot \cap A)}{Q_B(A)}
  = \b(\cdot\cap A) \le \b(\cdot).
\]
Now assume that $(\clA,\clQ^{\mu,\clA},\orthoG)$ is $P$-subadditive for all $P\ll\mu$.
Let $A,A'\in\clA$ with $A\nPortho\mu A'$.
Then we have $P:=Q_A+Q_A'\ll\mu$ and  $Q_A,Q_{A'}\le P$.
Since $\clA$ is $P$-subadditive 
there is a base measure $\b\le P$ with $A\cup A'\subset \supp \b\subset \supp P = A\cup A'$
by Lemma \ref{lemma.support.general}.
Consequently we obtain $A\cup A' = \supp \b\in \clA$.
\end{proofof}

\begin{lemma}\label{lemma.kinship.transitive}
Let $P\in \clM$ and  $(\clA,\clQ,\orthoG)$ be a  $P$-subadditive clustering base.
% Assume that the clustering base $(\clA,\clQ,\orthoG)$ is $P$-subadditive.
Then the kinship relation  $\sim_P$ is a symmetric and transitive relation on $\{B\in \clB\, |\, P(B)>0\}$ and 
an equivalence relation on the set $\{A\in \clA\, | \, \exists \alpha>0 \mbox{ such that  } \alpha Q_A \leq P    \}$.
Finally, 
% 
% on all $A\in\clA$ for
% which there is a base measure $\a\le P$ on $A$.
% 
% %Furthermore, let $\rep\a k\le P$ be base measures on $\rep Ak\in\clA$ that all are pairwise kin.
% %Then there is $\b\in \clQ_P(\rep[\cup]Ak)$ with $\b\ge \a_i$ for at least one of the $i\le k$.
% Furthermore, let 
for all finite sequences 
$\rep Ak\in\clA$ of sets that are  pairwise kin below $P$
there is $\b\in \clQ_P(\rep[\cup]Ak)$.
%Finally if $\rep\a k\le P$ are base measures on $\rep Ak\in\clA$ and
%$A_1\nPorthoP Ai$ for all $2\le i\le k$.
%Then there is $\b\in \clQ_P(\rep[\cup]Ak)$ with $\b\ge \a_i$ for at least one of the $1\le i\le k$.
\end{lemma}

\begin{proofof}{Lemma \ref{lemma.kinship.transitive}}
Symmetry is clear. Let $B_1\sim_P B_2$ and $B_2 \sim_P B_3$ be events with $P(B_i)>0$.
Then there are base measures $\c=\cc Q_C\in\clQ_P(B_1\cup B_2)$ and
$\c'=\cc'Q_{C'}\in\clQ_P(B_2\cup B_3)$ supporting them. This yields $B_2\subset C\cap C'$ and thus 
$P(C\cap C')\ge P(B_2)>0$. In other words, we have  $C\nPorthoP C'$, and by
subadditivity we conclude that there is a  $\b\in\clQ_P(C\cup C')$.
This gives $B_1\cup B_3\subset C\cup C' \subset \supp \b$, and therefore
$B_1\sim_P B_3$ at $\b$.
To show reflexivity on the specified subset of $\clA$, we fix 
an  $A \in \clA$ and an $\ca > 0$ such that $\a:=\ca Q_A\le P$. Then we have $\a \in \ca Q_P(\clA)$ and hence we obtain $A\sim_P A$.

%For any base set $A$ with a base measure $\a=\ca Q_A\le P$ we have $A\sim_P A$ at $\a$,
%and hence $\sim_P$ is reflexive on such $A$.

The last statement follows by induction over $k$, where the initial step $k=2$ is simply the definition of kinship.
Let us therefore 
assume the statement is true for some $k\ge2$.
Let $\rep A{k+1}\in\clA$ be pairwise kin.
By assumption there is a $\b\in\clQ_P(\rep[\cup]Ak)$.
Since this latter yields $A_1\subset\supp\b$ we find $A_1\sim_P\supp\b$
and by transitivity of $\sim_P$ we hence have $A_{k+1}\sim_P\supp\b$.
By definition there is thus a $\tilde\b\in\clQ_P(A_{k+1}\cup \supp\b)$
and since  this gives
$
  \rep[\cup]A{k+1} \subset A_{k+1} \cup\supp\b \subset \supp\tilde\b
%   \then \tilde\b\in\clQ_P(\rep[\cup]A{k+1}).
%   \qedhere
$
we find $\tilde\b\in\clQ_P(\rep[\cup]A{k+1})$.
\end{proofof}

\begin{lemma}\label{lemma.level}
%Assume $\clQ$ is flat.
Let $(\clA,\clQ,\orthoG)$ be a clustering base and 
$Q\in\clS(\clA)$ with representing  forest $F\in\clF_\clA$.
Then for all $A\in F$ %,  $A_0\in \max F$ and $B\subset A$  
we have
\[
    Q(\,\cdot\, \cap A) = \lambda_Q(A) + Q\Rl{A} \, .
\]
% 
% \begin{align}\label{lemma.level.h1}
%   Q(B)& 
%   = \sum_{A'\supset A} \alpha_{A'} Q_{A'}(B)
%   + \sum_{A'\subsetneqq A} \alpha_{A'} Q_{A'}(B)
%   + \sum_{A'\orthoG A_0} \alpha_{A'} Q_{A'}(B)\\ \label{lemma.level.h2}
%   Q(\,\cdot\, \cap A) & = \lambda_Q(A) + Q\Rl{A} 
% \end{align}
\end{lemma}

\begin{proofof}{Lemma \ref{lemma.level}}
Let $A_0\in \max F$ be the root with $A\subset A_0$. Then we can decompose $F$ into
$F = \{A'\in F:   A'\supset A\}  \dotcup \{ A'\in F:   A'\subsetneqq A \}  \dotcup \{ A'\in F:   A' \orthoG A \}$.
Moreover, flatness of $\clQ$ gives 
$Q_{A'}(\cdot \cap A) = Q_{A'}(A) \cdot Q_A(\cdot)$
for all $A'\in \clA$ with $A\subset A'$ while fittedness gives 
$Q_{A'} (A) = 0$ for all $A'\in \clA$ with $A'\orthoG A_0$ by the monotonicity of $\orthoG$, 
part (a) of Lemma \ref{lemma.separation.base.2},
and part (b) of Lemma \ref{lemma.support.general}.
For $B\in\clB$ we thus have
\begin{align*}
  Q(B\cap A)&{} 
  = \sum_{A'\supset A} \alpha_{A'} Q_{A'}(B\cap A) + \sum_{A'\subsetneqq A} \alpha_{A'} Q_{A'}(B\cap A) + \sum_{A'\orthoG A_0} \alpha_{A'} Q_{A'}(B\cap A)
  \\&{}= \sum_{A'\supset A} \alpha_{A'} Q_{A'}(A)  Q_A(B) + \sum_{A'\subsetneqq A} \alpha_{A'} Q_{A'}(B\cap A)\\
  &{}= \lambda_Q(A)(B) + Q\Rl{A}(B)\, ,
%  \qedhere  %%%%% Do not know why this does not work here...
\end{align*}
where  the last step  uses $Q_{A'}(B\cap A) = Q_{A'}(B)$ for $A'\subset A$, which follows from fittedness.
\end{proofof}

\begin{lemma}\label{lemma.ordering}
Let $(\clA,\clQ,\orthoG)$ be a  clustering base 
and  $\a,\b$ be base measures on $A,B\in\clA$ with $A\subset B$.
Then for all $C_0\in \clB$ with $\a(C_0\cap A)>0$ we have 
\[
   \b(\cdot \cap A) = \frac {\b(C_0 \cap A)}{\a(C_0 \cap A)}  \cdot \a(\cdot \cap A)\, .
\]
% 
% 
% If there is a $C_0\in \clB$ with $\b(C_0)< \a(C_0)$ then 
% $\b(\cdot \cap A)\le \a(\cdot \cap A)$.
\end{lemma}

\begin{proofof}{Lemma \ref{lemma.ordering}}
By assumption there are $\ca,\cb>0$ with $\a = \ca Q_A$ and $\b=\cb Q_B$.
% and for future use we note that $\ca = \a(A)$ and $\cb = \b(B)$ since $Q_A(A) = Q_B(B)=1$.
Moreover, flatness   guarantees $Q_B(\cdot \cap A) = Q_B(A)\cdot  Q_A(\cdot)$.
For all $C\in \clB$  we thus obtain 
\[
   \b(C\cap A) = \cb Q_B(C \cap A) 
   =  \cb  Q_B(A)\cdot  Q_A(C)
   =  \cb  Q_B(A)\cdot  Q_A(C\cap A)
   = \frac{\cb  Q_B(A) }{\ca} \a(C \cap A)\, .
\]
where in the second to last step we used $Q_A(\cdot)= Q_A(\cdot \cap A)$, which follows from  $A=\supp Q_A$.
For $C_0\in \clB$ with $\a(C_0\cap A)>0$ we thus find 
$\frac{\cb  Q_B(A) }{\ca}  = \frac {\b(C_0\cap A) }{\a(C_0 \cap A)}$
and inserting this in the previous formula gives the assertion.
% 
% 
% 
% and flatness 
% gives  $Q_B(A)=0$ or $Q_A(\cdot)=Q_B(\cdot \mid A)$.
% In the first case the assertion follows from 
% $\b(C\cap A) = \cb Q_B(C\cap A) \leq \cb Q_B(A) = 0\leq \a(C\cap A)$ for all $C\in\clB$. In the second case, we first note 
% that $\b(C_0\cap A) \leq \b(C_0) < \a(C_0) = \a(C_0\cap A)$, and hence we may assume that $C_0 \subset A$.
% 
% 
% For $C\subset A$ we further have 
% \[
%   \b(C) = \cb Q_B(C) = \cb\frac{Q_B(C)Q_B(A)}{Q_B(A)}
%   = \cb Q_B(C\mid A) Q_B(A) = \cb Q_A(C) Q_B(A) = \tfrac{\cb Q_B(A)}{\ca} \a(C)
% \]
% and $\a(C_0)<\b(C_0)$ means that $\frac{\ca Q_A(B)}{\cb} < 1$.
\end{proofof}

\begin{lemma}\label{lemma.simplicityI}
Let $(\clA,\clQ,\orthoG)$ be a  clustering base and $Q\in \clS(\clA)$ be a simple measure,
$\a$  be a base measures on some  $A \in \clA$, and $C\in \clB$.
Then the following statements are true:
\begin{enumerate}
   \item If  $\a\le Q$ then there is a level $\b$ in $Q$ with  $\a \leq \b$.
%   then there is a level $\b$ of $Q$ with $\a\le \b$.
  \item If  $\b(\cdot \cap C)\le \a(\cdot \cap C)$  
for all levels $\b$ of $Q$ then $Q(C)\le \a(C)$.
  \item For all $P\in \clM$ we have $Q\le P$ if and only if $\b\le P$ for all levels $\b$ in $Q$.
\end{enumerate}
\end{lemma}

\begin{proofof}{Lemma \ref{lemma.simplicityI}}
In the following we denote the representing forest of $Q$ by $F$.

\ada a 
By $\a\leq Q$ we find 
$A\subset \supp Q=\Gr F$. Since the roots $\max F$ form a finite $\orthoG$-disjoint union of closed sets of $\Gr F$,
the $\clA$-connectedness shows that $A$ is already contained in one of the roots, say $A_0\in \max F$.
\Use{Ortho:Connectedness:SuppMaxF}
For $F':=\{ A'\in F\mid A\subset A'\}$ we thus have $A_0\in F'$.
Moreover, $F'$ is a chain, since 
if there were $\orthoG$-disjoint $A',A''\in F'$
then $A$ would only  be contained in one of them  by Lemma \ref{lemma.separation.base.2}.% since $A\ne\emptyset$.
\Use{Ortho:Reflexivity}
Therefore there is a unique leaf $B:=\min F'\in F$ and thus $A\subset B$.
We %set $Q':=Q|_{F'}$ and 
denote the level of $B$ in $Q$ by $\b$.
% By the definition of levels we observe that $\b$ is also the level of $B$ in $Q$. Therefore 
Then 
it suffices to show 
 $\a\le \b$. To this end, let 
  $\{\rep Ck\}=\max F\Rl{B}$ be the direct children of $B$ in $F$.
% They are base sets and hence closed.
% By construction $A$ is not contained entirely in any of them.
\Use{Ortho:Connectedness:A}
By construction we know $A\not\subset C_i$ for all $i = 1,\dots,k$ and 
hence $\clA$-connectedness yields $A\not\subset \rep[\orthocupG] Ck$.
  Therefore $C_0:=A\setminus \bigcup_i C_i$ is non-empty 
   and relatively open in $A=\supp Q_A$. % by   part (d) of Lemma \ref{lemma.separation.base.2}.
  This gives $\a(C_0\cap A)  >0$
   by Lemma 
\ref{lemma.support.general}. Let us write  $\b := \lambda_Q(B)$ for the level of $B$ in $Q$.
Lemma  \ref{lemma.level} applied to the node $B\in F$ then gives
\[
   Q(C_0) = \b(C_0) + Q\Rl{B}(C) = \b(C_0) + \sum_{A'\in F:A'\subsetneqq B} \alpha_{A'} Q_{A'}(C_0) = \b(C_0)
\]
since for $A'\in F$ with  $A'\subsetneqq B$ we have $A'\subset \bigcup_i C_i$ and thus $\supp Q_{A'} \cap C_0= A'\cap C_0=\emptyset$.
Therefore, we find  $\a(C_0\cap A)= \a(C_0)\le Q(C_0)  =\b(C_0) = \b(C_0\cap B)$. By Lemma \ref{lemma.ordering} we conclude that 
$\b(\cdot \cap A) \geq  \a(\cdot \cap A)$. 
For $B'\in \clB$ the decomposition 
$B' = (B'\setminus A) \dotcup (B'\cap A)$ and the fact that $A = \supp \a \subset  \supp \b$ then yields the assertion.

\ada b 
For  $A\in F$ we define 
\[
   B_A:= A\setminus\bigcup_{A'\in F\colon A'\subsetneqq A} A'\, ,
\]
i.e.~$B_A$ is obtained by removing the strict descendants from $A$. From this description it is easy to see that 
$\{B_A:A\in F\}$ is a partition of $\Gr(F) = \supp Q$.
Hence we obtain
\begin{align}\nonumber
  Q(C) = \sum_{A\in F} Q(C\cap B_A)
  &{}= \sum_{A\in F} \sum_{A'\in F}\alpha_{A'}Q_{A'}(C\cap B_A)\\ \nonumber
  &{}= \sum_{A\in F}  \sum_{A'\supset A}\alpha_{A'}Q_{A'}(C\cap B_A)
           + \sum_{A\in F}\sum_{A'\subsetneqq A}\alpha_{A'}Q_{A'}(C\cap B_A) \\ \label{lemma.simplicityI-h1}
  &{}= \sum_{A\in F}  \lambda_{Q}(A) (C\cap B_A)\, ,
\end{align}
where   we used $Q_{A'} (C\cap B_A) = Q_{A'}(C\cap B_A\cap A)$ together with flatness applied to pairs
$A\subset A'$ as well as $A'\cap B_A=\emptyset$ applied to  pairs $A'\subsetneqq A$.  Our assumption now yields
% we now find 
  \begin{align*}
 Q(C)   \le \sum_{A\in F} \a(C\cap B_A)
  = \a(C\cap \supp Q)
  \le\a(C).
\end{align*}
\ada c Let $\b := \lambda_Q(B)$ be a level of $B$ in $Q$ with $\b\not\le P$. Then 
 there is  a $B'\in \clB$ with $\b(B')>P(B)$ and for $B'':= B'\cap \supp \b = B'\cap B$
 we find $Q(B'') \geq \a(B'') = \a(B')  >P(B') \geq P(B'')$.
Conversely, assume $\b\le P$ for all levels $\b$ in $Q$. 
By the decomposition \eqref{lemma.simplicityI-h1} we then obtain 
\[
  Q(C) 
  = \sum_{A\in F} \lambda_{Q}(A)(C\cap B_A)
  \le \sum_{A\in F} P(C\cap B_A)
  = P(C\cap \supp Q)
  \le P(C).\qedhere
\]
\end{proofof}

\begin{corollary}\label{kinship-inQ}
Let $(\clA,\clQ,\orthoG)$ be a  clustering base, $Q\in \clS(\clA)$ a simple measure with 
representing forest $F$ and $A_1,A_2\in F$. Then for all $\a\in \clQ_Q(A_1\cup A_2)$
there exists a level $\b$  in $Q$ such that 
 $A_1\cup A_2\subset B$ and $\a\leq \b$.
% and the level $\b$ of $B$ in $Q$ satisfies $\a\leq \b$ for all $\a\in \clQ_Q(A_1\cup A_2)$.
\end{corollary}

\begin{proofof}{Corollary \ref{kinship-inQ}}
 Let us fix an $\a\in \clQ_Q(A_1\cup A_2)$. Since $\a\leq Q$, Lemma  \ref{lemma.simplicityI}
 gives a level $\b$ in $Q$ with $\a\leq \b$. Setting $B:=\supp \b\in F$ then gives 
 $A_1\cup A_2 \subset \supp \a \subset B$. 
\end{proofof}

\begin{proofof}{Proposition \ref{prop.simple.measures.adapted.to.themselves}}
Let $Q$ be a simple measure and  $Q=\sum_{A\in F} \ca_A Q_A$ be its unique representation.
Moreover, let $A_1,A_2$ be direct siblings in $F$ and $\a_1,\a_2$ be the corresponding levels in $Q$.
Then $Q$-groundedness follows directly from Corollary \ref{kinship-inQ}.
To show that 
$A_1,A_2$ are $Q$-motivated and $Q$-fine, we fix an $\a\in  \clQ_Q(A_1\cup A_2)$.
Furthermore, let $\b$ be the level in $Q$ found by  Corollary \ref{kinship-inQ},
i.e.~we have $A_1\cup A_2\subset \supp \b=:B$ and $\a\leq \b\leq Q$.
% 
% 
% In order to show that 
% $A_1,A_2$ are $Q$-motivated and $Q$-fine,  it  suffices to  assume they belong to the same root, say $A_0\in \max F$, by 
% Corollary \ref{kinship-inQ}. 
Now let 
 $A_3,\dots,A_k\in F$ be the remaining  direct siblings of $A_1$ and $A_2$.
%  and $\a_3,\dots,\a_k$ be their levels in $Q$.
 Since $B$ is an ancestor of $A_1$ and $A_2$ it is also an ancestor of $A_3,\dots,A_k$
 and hence $A_1\cup \dots\cup A_k\subset B$. This immediately gives 
 $\b\in \clQ_Q(A_1\cup \dots\cup A_k)$ and we already know $\b\geq \a$. In other words, $A_1,A_2$ are $Q$-fine.
 Finally, observe that for $B\subset A'$ flatness gives 
 $Q_{A'}(B) Q_B(\cdot) = Q_{A'}(\cdot \cap B)$. Since $A_1\subset B$ we hence obtain
 \[
  \a(A_1)\le \b(A_1) 
  = \sum_{A'\supset   B} \alpha_{A'} Q_{A'}(B) Q_B(A_1)
  = \sum_{A'\supset   B} \alpha_{A'} Q_{A'}(A_1)
\]
and since $Q_{A_1}(A_1)=1$ we also find 
\[
 \a_1(A_1) = \sum_{A'\supset   A_1}\! \alpha_{A'} Q_{A'}(A_1) Q_{A_1}(A_1) 
 = \sum_{A'\supset   A_1}\! \alpha_{A'} Q_{A'}(A_1)  
 = \sum_{A'\supset   B}\! \alpha_{A'} Q_{A'}(A_1) + \alpha_{A_1}\, .
\]
Since $\ca_{A_1}>0$ we conclude that $\a(A_1) < (1-\epsilon_1) \a_1(A_1)$ for a suitable $\epsilon_1>0$. 
Analogously, we find an $\epsilon_2>0$ with  
$\a(A_2) < (1-\epsilon_2) \a_2(A_2)$ and taking $\ca := 1-\min\{\epsilon_1, \epsilon_2\}$
thus yields $Q$-motivation.
% 
% Let $\b\in\clQ_Q(C\cup C')$ be any base measure that supports $C$ and $C'$.
% Then by Simplicity~I there is $\tilde B\in F$ with level $\tilde \b \ge \b$.
% And this implies $\tilde B\supset C\cup C'$.
% We show adaptedness in three steps:
% \begin{description}
% \item[Motivated] By definition of level:
% \begin{align*}
%   \b(C) &{}\le \tilde\b(C) = \sum_{A\supset \tilde B} \alpha_A Q_A(C)
%   < \sum_{A\supset \tilde B} \alpha_A Q_A(C) + \alpha_C Q_C(C)
%   = \c(C)
% \end{align*}
% and similarily $\b(C')<\c'(C')$ and this yields $\b\not\ge(1-\epsilon)\c$ and $\b\not\ge(1-\epsilon)\c'$
% for some $\epsilon>0$.
% % \item[Fine] Let $C''$ be another direct sibling of $C$ and $C'$ and
% % let $\c''\le Q$ be the corresponding level.
% % Since $C''$ is a direct sibling of $C$ and $C'$ and
% % $\tilde B\in F$ is their parent therefore $C''\subset \tilde B$ as well.
% % So $\tilde\b$ already majorizes $\b$ and supports $\tilde B\supset C\cup C'\cup C''$.
% \qedhere
% \end{description}
\end{proofof}

\makeatletter{}%!TEX root = article.tex

\subsubsection{Proof of Theorem \ref{thm.uniqueness}}

% 
% The proof of Theorem~\ref{thm.uniqueness} uses the next two results.
% Firstly the uniqueness of roots in adapted approximations as a natural consequence
% of Lemma~\ref{lemma.kinship.roots}. Secondly the analogue for the children.
% 
% Both parts use the same kind of binary relation:
% the restriction of the $\nPorthoP$-relation to the nodes of a forest on the left and the right.
% More precisely, let $\rep Ak$ and $\rep{A'}{k'}$ be two sequences of events of positive $P$-probability.
% Then $\nPorthoP$ induces a binary relation $\sim$ on those sets.
% We call it injective (or right-unique) if for any right set there is at most one left set,
% and surjective if every right set has a left set.
% If both hold in both directions, then this induces a bijection between both sets.

\begin{lemma}\label{lemma.kinship.roots}
%Assume $Q_A(A)>0$ for all base sets $A\in\clA$.
Let $(\clA,\clQ,\orthoG)$ be a  clustering base,
$P\in\clM_\Omega$, and $Q,Q'\le P$ be simple measures on finite forests $F$ and $F'$.
If all roots in both $F$ and $F'$ are $P$-grounded,
then any root in one tree can only be kin below $P$ to at most one root in the other tree.
\end{lemma}

\begin{proofof}{Lemma \ref{lemma.kinship.roots}}
Let us assume the converse, i.e.~we have 
% Let us assume that there exists a root $A$ of $F$ that is below $P$ and two roots $B$ and $B'$ of $F'$,i.e. 
an $A\in \max F$ and $B,B'\in \max F'$ such that $A\sim_P B$ and $A\sim_P B'$.
Let $\a,\b,\b'$ be the respective summands in the simple measures $Q$ and $Q'$.
Then $0<\a(A)\le Q(A) \le P(A)$ and analogously $P(B),P(B')>0$.
Then by transitivity of $\sim_P$
established in Lemma \ref{lemma.kinship.transitive} we have $B\sim_P B'$ and by groundedness there has to be
a parent for both in $F'$, so they would not be roots.
\end{proofof}

%\bigskip

%Next we show some properties of monotone convergence for measures.

\begin{prop}\label{prop.unique.roots}
Let $(\clA,\clQ,\orthoG)$ be a stable clustering base and $P\in \clM$ such that 
 $\clA$ is $P$-subadditive.
Let $(Q_n,F_n)\uparrow P$, where all forests $F_n$ have $k$ roots
$A_n^1,\dots,A_n^k$, which, in addition, are  assumed to be $P$-grounded.
Then $A^i:=\bigcup_n A^i_n$ are unique under all such approximations up to a $P$-null set.
%They are pairwise disjoint.
\end{prop}

\begin{proofof}{Proposition \ref{prop.unique.roots}}
The $\repx{A^\x}k$ are pairwise $\orthoG$-disjoint\Use{Ortho:Nope?}
by Lemma \ref{lemma.limit.is.forest}
and by Lemma~\ref{lemma.support} they partition $\supp P$ up to a $P$-null set, 
i.e.~$P(\supp P\setminus\bigcup_i A^i) = 0$.
Therefore any  $B\in\clB$  with $P(B)>0$ intersects
at least one of the $A_i$.
Moreover, we have  $0<Q_1(A^i_1)\le P(A^i_1)\le P(A^i)$, i.e.~$P(A^i)>0$.
Now let $(Q_n',F_n') \uparrow P$ be another 
approximation of the 
assumed type
with roots $B_n^i$ and 
limit roots $B^1,\dots,B^{k'}$.
Clearly, our preliminary considerations also hold for these limit roots.
Now 
consider the binary relation $i\sim j$, which is defined to hold iff 
 $A^i\nPorthoP B^j$.

Since $P(A^i)>0$ there has to be a $B^j$ with $P(A^i\cap B^j)>0$,
so for all $i\le k$ there is a $j\le k'$ with $i\sim j$.
Then, since $A_n^i\cap B_n^j\uparrow A^i\cap B^j$, there is an $n\geq 1$ with 
 $P(A^i_n\cap B^j_n)>0$.
By $P$-subadditivity of $\clA$ we conclude that $A^i_n$ and $B^j_n$
 are kin below $P$, and 
 Lemma~\ref{lemma.kinship.roots} shows that this can only happen for at most one $j\le k'$.
 Consequently, we have  $k\le k'$ and $\sim$ defines an injection $i\mapsto j(i)$.
The same argument also holds in the other direction and we see that $k=k'$ and
that $i\sim j$ defines a bijection. Clearly,  we may  assume that $i\sim j$ iff $i=j$.
Then  $P(A^i\cap B^j)>0$ if and only if $i=j$, and since both sets of roots 
partition $\supp P$ up to a $P$-null set, we conclude that $P(A^i\triangle B^i)=0$.
% 
% only $P$-positive intersection with $A^i$ and not the other roots $B^i\setminus A^i$
% is a $P$-null set, and $A^i\setminus B^i$ as well, so $P(A^i\triangle B^i)=0$.
% Thus the set of roots is unique up to a $P$-null set.
\end{proofof}

\begin{lemma}\label{lemma.connected.to.motivated.siblings}
Let $(\clA,\clQ,\orthoG)$ be a   clustering base and $P\in \clM$
such that 
 $\clA$ is $P$-subadditive.
Moreover, let
% Assume $(\clA,\clQ,\orthoG)$ is $P$-subadditive.
 $\rep\a k\le P$ be base measures on $\rep Ak\in\clA$ such that 
$A_1\nPorthoP A_i$ for all $2\le i\le k$.
Then there is $\b\in \clQ_P(\rep[\cup]Ak)$ and an $\a_i$ such that 
 $\b\ge \a_i$, and if 
 $k\ge3$ and the $\a_2,\dots,\a_k$ satisfy the motivation implication \eqref{motiv-implicat} pairwise,
%  are pairwise $P$-motivated 
 then $\b\ge\a_1$.
\end{lemma}

\begin{proofof}{Lemma \ref{lemma.connected.to.motivated.siblings}}
The proof of the first assertion is based on induction.
% This follows by induction over $k\ge 2$.
For $k=2$ the assertion is  $P$-subadditivity. Now assume that the statement is true for $k$.
Then there is a $\b\in\clQ_P(\rep[\cup]A{k})$ and an $i_0\le k$ with $\b\ge \a_{i_0}$.
The assumed  $A_1\nPorthoP A_{k+1}$ thus yields
\[
  P(A_{k+1}\cap \supp \b) \ge P(A_{k+1}\cap A_1) > 0\, ,
\]
and hence $P$-subadditivity gives a $\tilde b\in\clQ_P(A_{k+1}\cup \supp \b)$
with $\tilde\b\ge\a_{k+1}$ or $\tilde\b\ge\b\ge\a_{i_0}$.
For the second assertion observe that $\b\in \clQ_P(A_i\cap A_j)$ for all $i,j$ and hence 
\eqref{motiv-implicat} implies $\b\not\ge\a_i$ for $i\ge 2$.
% If the $\a_2,\dots,\a_k$ are pairwise $P$-motivated $\b\not\ge\a_i$ for $i\ge 2$ and hence $\b\ge \a_1$
\end{proofof}

\begin{lemma}\label{lemma.strict.motivated}
%Assume $(\clA,\clQ)$ is flat and fulfills ordering.
Let $(\clA,\clQ,\orthoG)$ be a   clustering base
and $Q\le P$ be a simple and $P$-adapted measure with representing forest $F$.
Let $C^1,\ldots,C^k\in F$ be direct siblings for some $k\geq 2$.
% 
% We fix an $C^1\in F$ and consider all its direct siblings 
% $C^2,\ldots,C^k\in F$.
% If $k\geq 2$, 
Then 
 there exists an $\epsilon > 0$ such that:
\begin{enumerate}
\item For all  $\a\in\clQ_P(C^1\cup\ldots\cup C^k)$ and $i\le k$ we have $\a(C^i) \le (1-\epsilon) \cdot Q(C^i)$.
% \[
%   \a(C^i) \le Q(C^i)\cdot (1-\epsilon) \qquad\forall i\le k.
% \]
\item Assume that $\clA$ is $P$-subadditive and that $\a\leq P$ is a simple measure with  
$\supp \a\nPorthoP C^i$ for at least two $i\le k$.  Then for all 
$i\le k$ we have $\a(C^i) \le (1-\epsilon) \cdot Q(C^i)$.
% \[
%   \a(C^i) \le Q(C^i)\cdot (1-\epsilon) \qquad\forall i\le k.
% \]

\item If $\clA$ is $P$-subadditive and  $Q'\le P$ is a simple measure with representing forest $F'$ 
such that 
% 
% Let $Q'\le P$ be simple on $F'$ and assume 
there is an $i\le k$ with the property that for all $B\in F'$ we have
\[
  B\nPorthoP C^i \then \exists j\ne i\colon B\nPorthoP C^j.
\]
Then  $Q'(\cdot\cap C^i) \le (1-\epsilon)\,Q(\cdot \cap C^i)$ holds true.
% and even $Q'(D)\le (1-\epsilon) Q(D)$ for all events $D\subset C^i$.
\end{enumerate}
\end{lemma}

\begin{proofof}{Lemma \ref{lemma.strict.motivated}}
Let $\c_1,\dots,\c_k$ be the levels of $C^1,\ldots,C^k$ in $Q$.
Since $Q$ is adapted,  \eqref{motiv-implicat} holds for some 
$\ca\in(0,1)$.
% such that the levels $\ca \c_i$ are still motivated. 
We define $\epsilon:=1-\ca$.

\ada a
We fix an  $\a\in\clQ_P(C^1\cup\ldots\cup C^k)$, an $i \le k$, and a
 $j\le k$ with $j\ne i$.
Let $\c_i,\c_j$ be the levels of $C^i$ and $C^j$ in $Q$.
Since $\ca \c_i$ and $\ca \c_j$ are motivated, we have $\a\not \ge \ca \c_i$ and $\a\not \ge \ca \c_j$.
Hence, there is a $C_0\in \clB$ with $\a(C_0)< \ca \c_i(C_0)$ and thus also 
$\a(C_0 \cap C^i) < \ca \c_i(C_0\cap C^i)$.
Lemma~\ref{lemma.ordering} then yields $\a(\cdot \cap C^i)\le\ca\c_i(\cdot \cap C^i)$ and 
the definition of levels gives
% and we get by definition of level:
\[
  \a(C^i) \le \ca \c_i(C^i) = \ca Q(C^i) = (1-\epsilon) Q(C^i).
\]
% Since $i\le k$ was arbitrary this holds for all $i\le k$.
\ada b We may assume $\supp \a\nPorthoP C^1$ and $\supp \a\nPorthoP C^2$.
By the second part of  Lemma~\ref{lemma.connected.to.motivated.siblings}
applied to     $\supp \a,C^1,C^2$
there is an
$\a'\in\clQ_P(\supp \a \cup C^1\cup C^2)\subset \clQ_P( C^1\cup C^2)$ with  $\a'\geq \a$, and  
since $Q$ is $P$-fine, we may actually assume that $\a' \in\clQ_P(C^1\cup\ldots\cup C^k)$. 
Now part
(a) yields $\a'(C^i) \le (1-\epsilon) \cdot Q(C^i)$ for all $i=1,\dots,k$.
% By the first part of the proof then:
% \[
%   \a(C^1) \le \a'(C^1) \le (1-\epsilon) Q(C^1).
% %  \qedhere
% \]

\ada c We may assume $i=1$. 
Our first goal is to show %$Q'(C^i) \le (1-\epsilon)\,Q(C^i)$
% that for 
% .
% To this end it suffices to show for 
% all levels $\b$ in $Q'$ that 
\begin{align}\label{lemma.strict.motivated-h1}
 \b(\cdot \cap C^1)\le (1-\epsilon)\c_1(\cdot \cap C^1) 
\end{align}
for all levels $\b$ in $Q'$,
% 
% 
% $\b(\cdot \cap C^1)\le (1-\epsilon)\c_1(\cdot \cap C^1)$  
% for all levels $\b$ in $Q$, 
% since part (b) of Lemma \ref {lemma.simplicityI} then gives 
%  $Q'(C^1) \le (1-\epsilon)\c_1(C^1) \le (1-\epsilon) Q(C^1)$.
To this end,  we fix a level $\b$ in $Q'$ and write $B:= \supp \b$.
% 
% We show that $\a\le (1-\epsilon)\c_1$ on $C^1$ for all levels $\a$ of $Q'$.
% By Simplicity~II we then have that $Q'(C^1) \le (1-\epsilon)\c_1(C_1)
% \le (1-\epsilon) P(C_1)$ since $\c_1\le P$.
If $P(B\cap C^1)=0$, then \eqref{lemma.strict.motivated-h1} follows from 
% Let $A\in F'$ and $\a$ be its level in $Q'$.
% If $A\PorthoP C^1$ we have:
\begin{align*}
  \b(C^1) &{}=  \b(B\cap C^1) \le P(B\cap C^1)=0\, .
\end{align*}
In the other case  we have $B\nPorthoP C^1$ and our assumption gives a $j\neq 1$ with $B\nPorthoP C^j$.
By the second part of Lemma~\ref{lemma.connected.to.motivated.siblings} 
we find an  $\a\in\clQ_P(B\cup C^1\cup C^j)\subset \clQ_P(C^1\cup C^j)$
with $\a\ge \b$, and by (a) we thus obtain
$\a(C^1) \leq (1-\epsilon)\,Q(C^1) = (1-\epsilon)\c_1(C^1)$. Now, Lemma \ref{lemma.ordering} 
gives  $\a(\cdot\cap C^1) \leq  (1-\epsilon)\c_1(\cdot \cap C^1)$ and hence \eqref{lemma.strict.motivated-h1} follows.
% 
% we conclude that $\b(\cdot\cap C^1) \leq  (1-\epsilon)\c_1(\cdot \cap C^1)$
% 
% 
% Moreover, since $(1-\epsilon)\c_1$ and $(1-\epsilon)\c_j$ are motivated, we have 
% there is a $C_0\in \clB$ with $\a(C_0)< \ca \c_i(C_0)$ and thus also 
% $\a(C_0 \cap C^i) < \ca \c_i(C_0\cap C^i)$.
% 
% we have $(1-\epsilon)\c_1\not\le \a'$ on $C^1$ and by Lemma~\ref{lemma.ordering} therefore:
% \[
%   Q'(C^1) = \b(C^1) \le \a'(C^1) \le (1-\epsilon)\c_1(C^1)\le (1-\epsilon)\,Q(C^1).
% \qedhere
% \]

With the help of \eqref{lemma.strict.motivated-h1}  we now conclude by part (b) of Lemma \ref {lemma.simplicityI}
that $Q'(\cdot \cap C^1) \leq  (1-\epsilon)\c_1(\cdot \cap C^1)$
and using  $\c_1(\cdot \cap C^1) \leq Q(\cdot \cap C^1)$ we thus obtain the assertion.
% % % 
% % % % \leq  (1-\epsilon)Q(D \cap C^1) $.
% % % % 
% % % 
% % % Let us now fix a $D\in \clB$ and a level $\b$ of $Q'$.
% % % By \eqref{lemma.strict.motivated-h1} we then find 
% % % \[
% % %    \b(D \cap C^1)\le (1-\epsilon)\c_1(D \cap C^1)  \leq  (1-\epsilon)Q(D \cap C^1)  \, ,
% % % \]
% % % and hence the assertion follows by part (b) of Lemma \ref {lemma.simplicityI}.
% % % % 
% % % % Finally let $D\subset C^1$ be an event.
% % % % For all levels $\b$ of $Q'$ and for all events $E$ set $E':=E\cap D\subset C^1$.
% % % % Then by the above result we have
% % % % \[
% % % %   \b(E \cap D) = \b(E')=\b(E'\cap C^1)
% % % %   \le (1-\epsilon)\c_1(E' \cap C^1) = (1-\epsilon)\c_1(E') = (1-\epsilon) \c_1(E\cap D).
% % % % \]
% % % % So by part (b) of Lemma \ref {lemma.simplicityI} we have
% % % % $Q'(D) \le (1-\epsilon)\c_1(D)\le (1-\epsilon)Q(D)$.
\end{proofof}

\begin{lemma}\label{lemma.children.1}
Let $(\clA,\clQ,\orthoG)$ be a   clustering base 
and $P\in \clM$
such that 
 $\clA$ is $P$-subadditive.
Moreover, 
% and $Q_A(A)>0$ for all $A\in\clA$.
let $Q,Q'\le P$ be simple $P$-adapted measures on $F,F'$, and 
$S\in s(F)$ and $S'\in s(F')$ be two nodes that   have children in $s(F)$ and $s(F')$, respectively.
Let 
% 
% and denote those direct children by
\[
  \{C^1,\ldots,C^k\} = \max s(F)\Rl{S} \qquad \tand \qquad 
  \{D^1,\ldots,D^{k'}\} = \max s(F')\Rl{S'} 
\]
be their direct children and 
% Assume that these all are motivated and fine.
% 
% \qquad k,k'\ge 2.
% 
consider the   relation  
$i\sim j:\Leftrightarrow C^i\nPorthoP D^j$.
Then we have $k,k'\geq 2$ and if
  $\sim$ is left-total, i.e.\ for every $i\le k$ there is a $j\le k'$ with $i\sim j$,
then it is right-unique, i.e.~for every $i\le k$ there is at most one $j\le k'$ with $i\sim j$.
\end{lemma}

\begin{proofof}{Lemma \ref{lemma.children.1}}
The definition of the structure of a forest gives $k,k'\geq 2$.
Moreover, we note that 
 $P(A)\ge Q(A)>0$ for all $A\in F$ 
 and $P(A)\ge Q'(A)>0$ for all $A\in F'$.
%  and the same holds for $Q'$ and $F'$.
Now assume that $\sim$ is not right-unique, 
say
$1\sim j$ and $1\sim j'$ for some $j\ne j'$.
Applying  $P$-subadditivity twice we then find a $\b\in\clQ_P(C^1\cup D^j\cup D^{j'})$ 
with $\b\geq \c_1$ or $\b\geq \d_j$ or $\b\geq \d_{j'}$, where $\c_1$, $\d_j$, and $\d_{j'}$ are the 
corresponding levels.  
% that supports all three
% and is longer than at least one of the three levels $\c_1,\d_j,\d_{j'}$.
Since $\d^j,\d^{j'}$ are motivated we conclude that  $\b\ge \c_1$.
Now, because of  $\clQ_P(C^1\cup D^j\cup D^{j'}) \subset \clQ_P(D^j\cup D^{j'})$ and $P$-fineness of $Q'$
there is a $\b'\in\clQ_P(\rep[\cup] D{k'})$ with $\b'\ge \b$. % and the latter yields $C^1\subset \supp \b\subset \b'$.
Now pick a direct sibling of $C^1$, say $C^2$. Then 
 there is a $j''$ with $2\sim j''$, and since 
   $B':=\supp \b' \supset \rep[\cup]D{k'}$ this implies $P(B'\cap C^2)\ge P(D^{j''}\cap C^i)>0$.
By $P$-subadditivity we hence find a 
 $\b''\in\clQ_P(B'\cup C^2)\subset \clQ_P(C^1\cup C^2)$ with $\b''\ge\b'$ or $\b''\ge\c_2$.
 Clearly, $\b''\ge \c_2$ violates the fact that $C^1, C^2$ are motivated, and thus 
   $\b''\ge \b'$.  However, we have shown  $\b'\geq \b\ge \c_1$, and thus 
   $\b''\ge\c_1$. Since this again violates the fact that $C^1, C^2$ are motivated, we have found a contradiction.
%    
%    we obtain $\b''\geq \c_1$ by  $\b\ge \c_1$, 
%  
%  
% If $\b''\ge \b'\ge \c_1$ the two sibling levels $\c_1,\c_i$ are not motivated,
% and in the other case too.
% Therefore this cannot happen, and we see that $1\sim j$ for at most one $j$.
\end{proofof}

\begin{proofof}{Theorem \ref{thm.uniqueness}}
We prove the theorem by induction over the generations in the forests. 
For a finite forest $F$, we define 
 $s_0(F):=\max F$ and
\[
  s_{N+1}(F) := s_N(F) \cup \set{ A\in s(F) \mid \text{$A$ is a direct child of
  a leaf in $s_N(F)$} }.
\]
We will now show by induction over $N$ that there is a graph-isomorphism
$\zeta_N\colon s_N(F_\infty)\to s_N(F'_\infty)$ with
$P(A\triangle \zeta_N(A))=0$ for all $A\in s_N(F_\infty)$.
For   $N=0$ this has already been shown in  Proposition~\ref{prop.unique.roots}.
Let us therefore  assume that the statement is true for some  $N\geq 0$.
Let us fix an $S\in\min s_N(F_\infty)$ and let $S':=\zeta_N(S)\in\min s_N(F'_\infty)$
be the corresponding node.
We have to show that both have the same number of direct children in $s_{N+1}(\cdot)$ and
that these children are equal up to $P$-null sets. By induction this then finishes the proof.

Since $S\in s_N(F_\infty)\subset s(F_\infty)$, the node $S$ has either no children or at least $2$.
Now, if both $S$ and $S'$ have no direct children then we are finished.
Hence we can assume that  $S$ has direct children $C^1,\ldots,C^k$ for some $k\ge2$, i.e.
\[
  \max(F_\infty\Rl S) = \{ C^1,\ldots,C^k \}.
\]
Let $S_n,C^1_n,\ldots,C^k_n\in s(F_n)$ and $S'_n\in s(F'_n)$ be the nodes
that correspond to $S,C^1,\ldots,C^k$, and $S'$, respectively.
Since $P(S\triangle S')=0$ we then obtain for all $i\le k$
\begin{align*}
P(S'\cap C^i) = P(S\cap C^i) = P(C^i)\ge Q_1(C^i) \ge Q_1(C^i_1)>0\, ,
\end{align*}
that is $S'\nPorthoP C^i$ for all $i\le k$.
Since $S'=\bigcup_n S'_n$ and $C^i=\bigcup_n C^i_n$
this can only happen if 
$S'_n\nPorthoP C^i_n$ for all sufficiently large $n$.
We therefore may assume without loss of generality that 
\begin{equation}\label{ingo-h1}
   S'_1\nPorthoP C^i_n \qquad \qquad \mbox{for all $i\le k$ and all $n\geq 1$.}
\end{equation}
% 
%   $S'_1\nPorthoP C^i_n$ for all $i\le k$ and all $n\geq 1$.
Let us now investigate the   structure of $F'_n\Rle{S'_n}$.
To this end, we will seek a  kind of anchor $B'_n\in F'_n\Rle{S'_n}$, which
will turn out later to be the direct parent of the yet to find
$\zeta_{N+1}(C^i)\in F_\infty'$.
We define this anchor by 
% For $n\in\dsN$ let $B'_n\in F'_n$ be the minimal set that $P$-intersects all $C^i_1$:
\[
  B'_n := \min \{ B\in F'_n \mid B\nPorthoP C^i_1  \mbox{ for all } i=1,\dots,k\}.% \subset S'_n.
\]
This minimum is unique.
Indeed, let $\tilde B'_n$ be any other minimum
with\ $\tilde B_n'\nPorthoP C^i_1$ for all $i\le k$.
Since both are minima, none is contained in the other
and because $F'_n$ is a forest this means $B'_n\orthoG \tilde B'_n$.
Let $\b'_n$ and $\tilde\b'_n$ be their levels in $Q_n'$.
Since $Q_n'$ is $P$-adapted, these two levels are motivated.
This means that there can be no base measure majorizing one  of them and supporting
$B'_n\cup\tilde B'_n$.
On the other hand, by the second part of Lemma \ref{lemma.connected.to.motivated.siblings}
there exists a $\b''_n \in \clQ_p(B'_n\cup C^1_1\cup\dots\cup C^k_1)$ with
$\b''_n\ge\b'_n$.
% 
% On the other hand by $P$-subadditivity, there is a
% $\b''_n$ supporting $B'_n\cup C^1_1\cup\dots\cup C^k_1$ and majorizing $\b'_n$ or one of these levels.
% As all levels in $Q_n$ are motivated this can only mean $\b''_n\ge\b'_n$.
Now because of $P(\tilde B'_n\cap\supp\b''_n) \ge P(\tilde B'_n\cap C_1^1) > 0$ and $P$-subadditivity there exists
a base measure majorizing $\tilde\b'_n\ge\b'_n$ or $\b''_n$ and supporting
$\tilde B'_n\cap\supp\b''_n$.
This contradicts the motivatedness of $\b'_n$ and $\tilde\b'_n$ and
hence the minimum $B'_n$ is unique.
% % % % % % % % % % % % % % % % % % % % % % % % % % % % % % % % % % % % 
% 
% I dont understand this part: why can we find \b_n'' ???  FIXED and CHECKED
% Also: why do we have B'_n\subset S'_n as said below?     FIXED and CHECKED
%%% 
%
% I hope, it is clear now?
% 
% % % % % % % % % % % % % % % % % % % % % % % % % % % % % % % % % % % % % % 

Since $B'_n$ is the unique minimum among all $B\in F'_n$ with $B\nPorthoP C_1^i$ for all $i$,
we also have $B'_n\subset B$ for all such $B$ and hence $B'_n\subset S'_n$ by %\in s(F'_n)$ by 
\eqref{ingo-h1}.
The major difficulty in handling $B'_n$ though is that it may jump around as a function of $n$:
Indeed we may have  $B'_n\in F'_n\setminus s(F'_n)$ and therefore
the monotonicity $s(F'_n)\le s(F'_{n+1})$ says nothing about $B'_n$.
In particular, we have in general  $B'_n\not\subset B'_{n+1}$.
%It is not even clear a priori how deep inside $s(F'_n)$ it is located:
%there could be a child $C'\in s(F'_\infty)

Let us now   enumerate the set $\min F'_n\Rl{B'_n}$ of direct children of $B_n'$ 
 by $D_n^1,\ldots,D_n^{k_n}$,
where $k_n\geq 0$.
%:
%\[
%  \set{ D_n^1,\ldots,D_n^{k_n} } = \min \big(F'_n\Rl{B'_n}\big).
%\]
Again these $D^i_n$ can jump around as a function of $n$.
The number $k_n$ specifies different cases: we have  $B'_n\in\min F'_n$, i.e.~$B_n'$ is a leaf, iff $k_n=0$;
on the other hand $D_n^i\in s(F'_n)$ iff $k_n\ge2$.
Next we  show that for all $i\le k$ and all sufficiently large $n$ 
there is an index $j(i,n)\in \left\{1,\dots,k_n\right\}$ with 
\begin{equation}\label{ingo-h3}
   C^i_1\nPorthoP D_n^{j(i,n)}\, .
\end{equation}
Note that this in particular implies $k_n\ge1$ for sufficiently large $n$.
To this end we fix an $i\le k$. Suppose that 
$C^i_1\PorthoP (D_{n_m}^1\cup\dots\cup D_{n_m}^{k_{n_m}})$ for infinitely many $n_1,n_2,\dots$.
By construction $B'_{n_m}$ is the smallest element of $F'_{n_m}$ that $\PorthoP$-intersects $C^i_1$.
More precisely, %in this subsequence %for all $m\ge 1 and n \rightarrow n_m$
for any $A\in F'_{n_m}$ with $A\nPorthoP C^i_1$
we have $A\supset B'_{n_m}$ and therefore $A\nPorthoP C^{i'}_1$ for all such $A$ and all $i'\le k$.
Hence, all $Q'_{n_m}$ in this subsequence fulfill the conditions of the last statement
in Lemma~\ref{lemma.strict.motivated} and we get an $\epsilon > 0$ such that for all such $n_m$
\begin{align}
  Q'_{n_m}(C^i_1)\le (1-\epsilon) Q_1(C^i_1)\le (1-\epsilon) P(C^i_1)
  \label{eq.thm.0}
\end{align}
which contradicts $Q_{n_m}'(C^i_1)\uparrow P(C^i_1)$ since $P(C^i_1)>0$.

Therefore for all $i\le k$ and all sufficiently large  $n$
there is an index $j(i,n)$ such that \eqref{ingo-h3} holds.
Clearly, we may thus assume that there is such an $j(i,n)$ for all $n\geq 1$. 
% , say for all $n\ge1$.
Since $j(i,n)\in \left\{1,\dots,k_n\right\}$ we conclude that $k_n\geq 1$ for all $n\geq 1$.
% 
% Now for all $n$ this means $k_n\ge 1$, else there could be no $1\le j(i,n)\le k_n$.
Moreover, $k_n=1$ is impossible, since $k_n=1$ yields $j(i,n)=1$,  
% And we even have $k_n\ge2$, else $j(i,n)=1$
and this would mean, that $C^i_1\nPorthoP D^1_n$
for all $i\le k$ contradicting that $B'_n$ is the minimal set in $F'_n$ having this property.
Consequently $B'_n$ has the direct children $D^1_n,\ldots,D^{k_n}_n$ where $k_n \geq 2$ for all $n\geq1$.
%Since they are direct siblings they all appear in $s(F'_1)\le s(F'_2)\le\ldots\le F'_\infty$
%in some generation inside $S'_n$
%and hence $k_n=k'$ for some $k'\ge 2$.
%Furthermore $C^i_1\nPorthoP D^j_1$ implies $C^i\nPorthoP D^j$ at least for $j=j(i,1)$.

So far we  have seen that   $D^1_n,\ldots,D^{k_n}_n\in s(F'_n)$ are inside $S'_n$.
Therefore $S'_n$ is not a leaf, and hence $S'\notin\min F'_\infty$ as well.
But still for infinitely many $n$ these $D^j_n$ might not be the direct children of $S'_n$.
Let us therefore denote 
the direct children of $S'_n\in s(F'_n)$
by $E^1_n,\ldots,E^{k'}_n\in s(F'_n)$, where we pick a numbering such that 
$E^i_n\subset E^i_{n+1}$ and by the definition of the structure of a forest we have $k'\ge 2$.
%We now have to show that these are the same as the $D^1_n,\ldots,D^{k_1}_n$.

For an arbitrary but fixed $n$ we now 
show  $\{D^1_n,\ldots,D^{k_n}_n\} = \{E^1_n,\ldots,E^{k'}_n\}$.
To this let us assume the converse.
% So assume the $E_n^j$ are not the $D^j_n$.
Since the $E_n^j$ are the direct children of $S'_n$ in the structure $s(F'_n)$
there is a  $j_n\le k'$ with $D_n^j\subset E_n^{j_n}$ for all $j$, and since 
 $B'_n$ is the direct parent of the $D^j_n$ we conclude that  $B'_n\subset E_n^{j_n}$.
Therefore we have $C^i_1\nPorthoP E_n^{j_n}$ for all $i\leq k$.
%So the level of $E_n^{j_n}$ can be majorized by a base measure $\e_n$ supporting all $C^i_1$.
%and therefore if this happens for infinitely many $n$ 
% and we show that for all $i\le k$
%Then $i\sim j_n$ for all $i\le k$ because $P(C^i_1\cap E_n^{j_n})\ge P(C^i_1\cap B'_n)>0$.
%So for all $i$ there is at least a $j$ with $i\sim j$, namely $j=j_n$.
Since $Q_1$ and $Q_n'$ are adapted we can use Lemma~\ref{lemma.children.1} to 
 see that for all $i\le k$ we have $C^i_1\PorthoP E_n^{j}$ for all $j\ne j_n$.
Let us fix a $j\neq j_n$.
Our goal is to show 
\[
   Q_m(E_n^j)<(1-\epsilon)Q'_n(E_n^j)\, ,
\]
for all sufficiently large $m\geq n$, since this inequality contradicts the assumed convergence
of $Q_m(E_n^j)$ to $P(E_n^j) \ge Q_n'(E_n^j)>0$.
By part (c) of Lemma~\ref{lemma.strict.motivated} with $Q_n'$ as $Q$ and $Q_m$ as $Q'$ 
it suffices to show that for all $A\in F_m$ and all sufficiently large $m\geq n$ we have
\begin{equation}\label{ingo-h2}
   A\nPorthoP E^j_n\then A\nPorthoP E^{j_n}_n\, .
\end{equation}
To this end, we fix an $A\in F_m$ with $A\nPorthoP E^j_n$.
Then 
we first observe that for all $m\geq n$ we have 
$P(A\cap S'_m)\ge P(A\cap S'_n)\ge P(A\cap E^j_n)>0$.
Moreover, the  induction assumption ensures $P(S\triangle S')=0$ and 
since $S_m\nearrow S$ and $S'_m\nearrow S'$, we conclude that $P(A\cap S_m)>0$
for all sufficiently large $m$.
% let us fix an
% 
% To use part (c) of Lemma~\ref{lemma.strict.motivated} with $Q_n'$ as $Q$ and $Q_m$ as $Q'$ 
% we show that for any $A\in F_m$ we have
% $A\nPorthoP E^j_n\then A\nPorthoP E^{j_n}_n$
% and then we get $Q_m(E_n^j)<(1-\epsilon)Q'_n(E_n^j)$
% for all $m$ which contradicts convergence.
% Indeed, let 
% 
% $A\in F_m$ with $A\nPorthoP E^j_n$.
% This implies $A\nPorthoP S'_n$ and by induction assumption hence
% $P(A\cap S_m) = P(A\cap S'_m)\ge P(A\cap S'_n)\ge P(A\cap E^j_n)>0$.
Now,  $C_m^1\cup\dots\cup C_m^k$ are direct siblings and hence we either have  
% 
% Since $F_m$ is a forest we thus find $A\subset S_m$,
% and hence we have either 
$C_m^1\cup\dots\cup C_m^k \subset A$
or $A\subset C_m^{i_0}$ for exactly one $i_0\le k$.
% 
% or for one and only one $i_A\le k$ we have $A\subset C_m^i$.
In the first case we get
\[
  P(A\cap E^{j_n}_n) \ge P(C_m^1\cap E^{j_n}_n) \ge P(C_1^1\cap E^{j_n}_n) > 0
\]
by the already established $C^i_1\nPorthoP E_n^{j_n}$ for all $i\leq k$.
The second case is impossible, since it contradicts adaptedness. Indeed, 
$A\subset C_m^{i_0}$ implies $C_m^{i_0}\nPorthoP E_n^{j}$ and 
by the already established $C^i_1\nPorthoP E_n^{j_n}$ for all $i\leq k$, we also know
$C_m^{i_0}\nPorthoP E_n^{j_n}$.
By the second part of Lemma \ref{lemma.connected.to.motivated.siblings}
we therefore find a $\tilde\c \in \clQ_P( C_m^{i_0}\cup E_n^j\cup  E_n^{j_n})$
with $\tilde \c \geq \c_m^{i_0}$, where $\c_m^{i_0}$ is the level of $C_m^{i_0}$ in $Q_m$.
Now fix any $i\le k$ with $i\ne i_0$ and observe that we have 
$P(C_m^i\cap \supp\tilde\c) \ge P(C_m^i\cap E^{j_n}_n)\ge P(C_1^i\cap E^{j_n}_n)>0$,
and hence $P$-subadditivity yields a $\c''\in \clQ_P(C_m^i\cup \supp\tilde\c)$
with $\c''\geq \c^i_m$ or $\c''\geq \tilde \c\geq \c_m^{i_0}$, where 
 $\c_m^{i}$ is the level of $C_m^{i}$ in $Q_m$. Since $\c''\in \clQ_P(C_m^i\cup \supp\tilde\c) \subset 
 \clQ_P(C_m^i\cup C_m^{i_0})$, we have thus found a contradiction to the 
 fact that the direct siblings $C_m^i$ and $C_m^{i_0}$ are $P$-motivated.

% The second case contradicts adaptedness: In fact this means that $C_m^{i_0}\nPorthoP E_n^{j_n}$
% and $C_m^{i_0}\nPorthoP E_n^{j}$ 
% 
% so by $P$-subadditivity and fineness of $E_n^1,\ldots,E_n^{k'}$
% the level $\c_m^{i_0}$ of $C_m^{i_0}$ in $Q_m$ can be majorized by a $\tilde\c$ supporting
% $C_m^{i_0}\cup E_n^1\cup\ldots\cup E_n^{k'}$.
% Since $P(C_m^i\cap \supp\tilde\c) \ge P(C_m^i\cap E^{j_n}_n)\ge P(C_1^i\cap E^{j_n}_n)>0$
% for all $i\le k$ this contradicts motivatedness of $C_m^{i_0}$ and all its siblings.
% % % % % % % % % % % % % % % % % % % % % % % % % % % % % % % % % % % % 
% 
% I dont understand the implication for A\in F
% Also: I did not check the rest of the proof
%
%
% Is it clearer when we say A\in F_1? We can also say, that the roles in Lemma 42 are reversed...
% 
% % % % % % % % % % % % % % % % % % % % % % % % % % % % % % % % % % % % % % 

So far we have shown 
 $\{D^1_n,\ldots,D^{k_n}_n\} = \{E^1_n,\ldots,E^{k'}_n\}$ and $k_n=k'$ for all $n$.
Without loss of generality we may thus  assume that $D^j_n=E^j_n$ for all $n$ and all $j\le k'$.
In particular, this means that the direct children of $S_n'$ in $s(F_n')$ equal the 
direct children of $B_n'$ in $F_n'$.
Let us write 
%  
%  
% Remark that we now know that $B'_n$ is the direct parent of the  of the first children of $S'_n$.
% Without loss of generality we assume that $D^j_n=E^j_n$ for all $n$ and all $j\le k'$ and we let
\[
  D^j:=\bigcup_{n\geq 1} D^j_n, \qquad\qquad  j=1,\dots , k'
\]
and  $i\sim j$ iff $C^i_1\nPorthoP D_1^j$.
We have seen around \eqref{ingo-h3}
that for all $i\le k$ there is at least one $j\le k_1=k'$ with $i\sim  j$, namely $j(i,1)$.
By  Lemma~\ref{lemma.children.1} we then conclude that $j(i,1)$ is the only index $j\leq k'$ satisfying
$i\sim j$.
% 
% Now, $Q_1$ and $Q_1'$ are adapted and using Lemma~\ref{lemma.children.1}
% we see that for all $i$ there is only $j(i,1)$ s.t.\ $i\sim j$.
By reversing the roles of $C^i_1$ and $D^j_1$, which is possible since $D^j_1 = E^j_1$ is a direct children of $S'_n$ in 
$s(F'_n)$,
we can further see that for all $j$ there is an index $i$ with $i\sim j$
and again by Lemma~\ref{lemma.children.1} we conclude that there is at most one  $i$ with $i\sim j$.
Consequently,  $i\sim j$ defines a bijection between $\{C^1_1,\ldots,C^k_1\}$ and $\{D^1_1,\dots,D^{k'}_1\}$
and hence we have $k=k'$. Moreover, we may assume without loss of generality that 
$i\sim j$ iff $i=j$. From the latter we obtain 
% 
% Hence $k=k'$.
% W.l.o.g.\ we assume $i\sim j$ iff $i=j$.
% From this also follows 
$C^i_1\nPorthoP D^j_1$ iff $i=j$.
% 
% andd hence $C^i\nPorthoP D^i$ for all $i\le k$.

To generalize the latter, we fix $n,m\geq 1$ and write $i\sim j$ iff 
$C^i_n\nPorthoP D_m^j$. Since we have $P(C^i_n\cap D^i_m)\ge P(C^i_1\cap D^i_1)>0$,
we conclude that $i\sim i$, and by Lemma~\ref{lemma.children.1} we again see that $i\sim j$ is 
false for $i\neq j$. This yields 
 $C^i_n\nPorthoP D^j_m$ iff $i=j$
 and by taking the limits, we find  $C^i\nPorthoP D^j$ iff $i=j$.

% 
% for all $i$ there is a $j$ with $i\sim j$ and for all $j$ there is 
% an $i$ with $i\sim j$.
% 
% 
% Now observe that 
% for all $n,m$ we have $P(C^i_n\cap D^i_m)\ge P(C^i_1\cap D^i_1)>0$
% 
% 
% 
% and because $Q_n$ and $Q_m'$ are adapted we get using the same argument based on
% Lemma~\ref{lemma.children.1}
% 
% the same kind of bijection between
% $\{\repx{C^\x_n}k\}$ and $\{\repx{D^\x_m}k\}$.
% This gives for all $n,m$ that if $i\ne j$ then $C^i_n\nPorthoP D^j_m$ and
% hence $P(C^i\cap D^j)=\lim_n P(C^i_n\cap D^j_n) = 0$.
% On the other hand we have $P(C^i\cap D^i)\ge P(C^i_n\cap D^j_m)\ge P(C^i_1\cap D^j_1)>0$.
% All in all we have $C^i_n\nPorthoP D^j_m$ iff $i=j$, and $C^i\nPorthoP D^j$ iff $i=j$
% and by taking the limits, we find 

Next we show that $P(C^i\triangle D^i)=0$ for all $i\le k$.
Clearly, it suffices to consider the case $i=1$.
% 
% , by showing it for $i=1$.
To this end assume that  $R:=C^1\setminus D^1$ satisfies $P(R)>0$.
For $R_n:=R\cap C^1_n=C^1_n\setminus D^1$,
we then have 
 $R_n\uparrow R$ since $C^1_n\uparrow C^1$ and $R\subset C^1$.
Consequently, $0<P(R)=P(R\cap C^1)$ implies $P(R_n)>0$ for all sufficiently large $n$.
%and hence there is $m_0\ge n_0$ with $Q_{m}(R_n)>0$ for all $m\ge m_0$ and $n\ge n_0$.
%So $Q_n(R_n)>0$ for all $n\ge m_0$.
On the other hand, we have $P(R\cap D^1)=0$ by the definition of $R$ and $P(R\cap D^j)\le P(C^1\cap D^j)=0$
for all $j\neq 1$ as we have shown above.

We next show that $Q'_m(R_n)=Q'_m\Rge{B'_m}(R_n)$.
To this end it suffices to show that for any $A\in F'_m$ with $A\notin F'_m\Rge{B'_m}$ we have $Q'_m(A\cap R_n) \le P(A\cap R_n)=0$.
Let us thus fix an $A\in F'_m$ with $A\notin F'_m\Rge{B'_m}$. Then we either have 
$A\subsetneqq B'_m$ or $A\orthoG B'_m$.
In the first case there is $j\le k$ with $A\subset D_m^j$ which means, as shown above, that  $P(A\cap R_n)\le P(D_m^j\cap R_n)=0$.
In the second case, by definition of structure, we even have $A\orthoG S'_m$.
So there is a $A_m'\in s(F'_m)$ with $A\subset A_m'$ and $A_m'\orthoG S'_m$
and by isomonotonicity of the structure there is $A'\in F'_\infty$ with $A_m'\subset A'$ and $A'\orthoG S'$.
Hence by induction assumption $P(A\cap R_n)\le P(A\cap S_n) \le P(A\cap S) \le P(A'\cap S) = P(A'\cap S')=0$.

Using $P(C^i \cap D^i)>0$ we now observe that 
$Q'_m\Rge{B'_m}$ fulfills the conditions of part (c) of Lemma~\ref{lemma.strict.motivated} 
for $C^1$ and $C^2$ and by $R_n\subset C^1_n$ we thus obtain 
\[
Q'_m(R_n)=Q'_m\Rge{B'_m}(R_n)\le (1-\epsilon)Q_n(R_n)\le (1-\epsilon) P(R_n).
\]
%We fix an $n\ge m_0$ for the moment s.t.\ $Q'_n(R_{m_0}\cap S'_n)>0$.
%Let $\b'_n$ be the level of $B'_n$ in $Q'_n$.
%Since $B'_n\nPorthoP C^j_{m_0}$, $j\le k$, there is $\tilde\b'\in Q_P(B'_n\cup C^1_{m_0}\cup\dots\cup C^k_{m_0})$
%with $\tilde\b'\ge\b'_n$ and set $\tilde B':=\supp\tilde\b'$.
%%From Simplicity~ that $Q'_n(E)\le \tilde\b'_n(E)$
%Remember that $A\in F'_n$ with $B'_n\subset A\subset S'_n$ form a chain and
%that $\tilde C:=C^1_{m_0}\cup\dots\cup C^k_{m_0}\subset S_{m_0}$
%For all $E\subset\tilde B'\cap S'_n\setminus (D^1_n\cup\dots\cup D^k_n)$ one can show that
%\begin{align*}
%  Q'_n(E) \le \tilde\b'_n(E)
%\end{align*}
%On the other hand from Lemma~\ref{lemma.strict.motivated} and Lemma \ref{lemma.ordering}
%it follows that for all $E\subset C^1_{m_0}$:
%\[
%  Q'_n(E) \le\tilde\b'_n(E) \le (1-\epsilon) \c^1_{m_0}(E) \le (1-\epsilon) Q_{m_0}(E) \le (1-\epsilon) P(E)
%\]
This contradicts $0<P(R_n) = \lim_\toi m Q'_m(R_n)$.
So we can assume $P(R_n)=0$ for all $n$ and therefore $P(R) = \lim_\toi n P(R_n) =0$.
By reversing roles we thus find $P(D^1\triangle C^1)=P(C^1\setminus D^1)+P(D^1\setminus C^1)=0$
and therefore the children are indeed the same up to $P$-null sets.

Finally, we are able to finish the induction: To this end we extend $\zeta_N$ to the map
$\zeta_{N+1}\colon s_{N+1}(F_\infty) \to s_{N+1}(F'_\infty)$ by setting, for  every leaf $S\in \min s_N(F_\infty)$,
\[
 \zeta_{N+1}(C^i) := D^i
\]
where $\repx{C^\x}k\in s_{N+1}(F_\infty)$ are the direct children of $S$ and 
$\repx{D^\x}k\in s_{N+1}(F'_\infty)$ are the nodes we have found during our above construction.
Clearly, our construction shows that $\zeta_{N+1}$ is  a graph isomorphism satisfying 
% 
% 
% for every leaf $S\in \min s_N(F_\infty)$ the image of its children $\repx{C^\x}k\in s_{N+1}(F_\infty)$
% to $\zeta_{N+1}(C^i) := D^i$ for all $i\le k$.
% Then we have 
$P(A\triangle \zeta_{N+1}(A))=0$ for all $A\in s_{N+1}(F_\infty)$.
\end{proofof}

\makeatletter{}\subsubsection{Proof of Theorem \ref{thm.extension.axioms}}

\begin{lemma}\label{lemma.disjoint.simple.adapted}\Use{Ortho:DisjointSupport:P}
Let $(\clA,\clQ,\orthoG)$ be a   clustering base, 
$\rep Pk\in\clM$ with $\supp P_i\orthoG\supp P_j$ for all $i\ne j$, and 
 $Q_i\le P_i$ be simple measures with representing forests $F_i$. 
We define $P:=\rep[+]Pk$, $Q:=\rep[+]Qk$, and $F:=F_1\cup\dots\cup F_k$.
Then we have:
% 
% Then $Q$ is a simple measure and $F$ is its representing 
% $\orthoG$-forest.
% % 
% % Moreover, let  $Q_i\le P_i$ be simple measures with representing forests $F_i$. Then 
% % $F:=F_1\cup\dots\cup F_k$ is a $\orthoG$-forest.
% Moreover, if 
% % 
% % %Assume $\clA$ is fitted and let
% % and $Q_i\in\clS$ be simple on forests $F_i$, $i\le k$.
% % If $Q_i\le P_i$ for all $i\le k$ then $F:=F_1\cup\dots\cup F_k$ is a $\orthoG$-forest.
% % \Use{Ortho:Forest}
% $\clA$ is $P_i$-subadditive and $Q_i$ is $P_i$-adapted   for all $i\le k$,
% then the following statements hold:
% 
% 
% Let $Q:=\rep[+]Qk$ and $P:=\rep[+]Pk$. Then $Q$ is simple on $F$ and:
\begin{enumerate}
\item The measure $Q$ is  simple  and $F$ is its representing 
$\orthoG$-forest.
\item For all base measures $\a\le P $ there exists exactly one $i$ with $\a\le P_{i}$.
% \[
%   \a\le P \then \exists! i(\a)\le k \colon \a\le P_{i(\a)}.
% \]
\item If $\clA$ is $P_i$-subadditive for all $i\leq k$, then $\clA$ is $P$-subadditive.
\item if  $Q_i$ is $P_i$-adapted   for all $i\le k$, then $Q$ is adapted to $P$.
\end{enumerate}
\end{lemma}

\begin{proofof}{Lemma \ref{lemma.disjoint.simple.adapted}}
%Since $\clA$ is fitted $\Gr F_i=\supp Q_i$.
\Use{Ortho:DisjointSupport:P}\Use{Ortho:DisjointGround}
\ada a
Since $Q_i\le P_i\le P$ we have $\Gr F_i=\supp Q_i\subset \supp P_i$.
% 
% Let $i\ne j$. Then $\supp P_i\orthoG \supp P_j$ and 
By the monotonicity of $\orthoG$ we then obtain
$\Gr F_i\orthoG \Gr F_j$ for  $i\ne j$.
From this we  obtain the assertion.
% 
% 
% and another application of  monotonicity  yields 
% $A\orthoG A'$  for all $A\in F_i$ and $A'\in F_j$ we have
% \Use{Ortho:Monotonicity:A}
% $A\orthoG A'$ and hence $F$ is indeed a $\orthoG$-forest.
% Now for the other statements:

\ada b Let $\a\le P$ be a base measure on $A\in\clA$. Then we have 
 $A=\supp \a\subset \supp P = \bigcup_i \supp P_i$.
%The supports are disjoint and $A$ is connected
By $\clA$-connectedness there thus exists a $i$ with 
% 
% $\clA$-connectedness\Use{Ortho:Connectedness}
% there is $i(A)\le k$ with 
% 
$A\subset \supp P_{i}$. For $B\in \clB$ we then find 
$\a(B) = \a(B\cap \supp P_{i}) \leq P(B\cap \supp P_{i}) = P_i(B\cap \supp P_{i}) = P_i(B)$.
Moreover, for $j\neq i$ we have $\a(A) > 0$ and $P_j(A) = 0$ and thus $i$ is unique.
% 
% and since the $\supp P_i$ are pairwise $\orthoG$-disjoint we have for all $B\subset A\subset \supp P_{i(A)}$:
% \[
%   \a(B) \le P_1(B)+\dots+P_k(B) = P_{i(A)}(B).
% \]

\ada c  
Let $\a,\a'\le P$ be base measures on base sets $A,A'$ with  $A\nPorthoP A'$.
Since  $A\orthoG A'$ implies  $A\orthoD A'$, we have 
 $A\northoG A'$. By (b) we find unique indices $i,i'$ with $\a\leq P_i$ and $\a'\leq P_{i'}$. 
 This implies 
 $A\subset\supp P_i$ and $A'\subset \supp P_j$, and hence  we have $\supp P_i\northoG\supp P_{i'}$
 by monotonicity. This gives $i=i'$, i.e.~$\a,\a'\le P_i$.
 Since $\clA$ is $P_i$-subadditive
there now is an $\tilde\a \in\clQ_{P_i}(A\cup A')$ with $\tilde\a\ge\a$ or $\tilde \a\ge\a'$, and 
since $\tilde\a\le P_i\le P$ we obtain the assertion.

\ada d From (b) we conclude $\clQ_P(A_1\cup A_2)=\emptyset$ for all roots $A_1\in F_i$ and $A_2\in F_j$ and all $i\neq j$.
This can be used to infer the groundedness and fineness of $Q$ from the 
groundedness and fineness of the $Q_i$.
% 
% groundedness of the $F_i$ we obtain the groundedness of $F$
% 
% 
% % 
% % $F$ is grounded since there is by (b) no base measure $\a$ supporting any
% % root of $F_i$ with any root of $\bigcup_{j\ne i}F_j$.
% % No pair of roots is below P by (b) and all other pairs of direct siblings are contained in some $F_i$. 
% Moreover, 
% 
% Since $Q_i$ is adapted 
% $F$ also inherits fineness of $F_1$ and $F_2$ because direct siblings
% only supported inside either of those.
Now let $\a,\a'\le P$ be the levels of some direct siblings $A,A'\in F$ in  $Q$ %with $A\orthoG A'$.
and $\b\in\clQ_P(A\cup A')$ be any base measure.
By (b) there is a unique $i$ with $\b\le P_i$, and hence $\a,\a'\le P_i$ as well.
Therefore $Q$ inherits strict motivation from $Q_i$.
\end{proofof}

\begin{lemma}\label{lemma.base.added.adapted}
Let $(\clA,\clQ,\orthoG)$ be a   clustering base, 
 $P\in\clM$,   
 $\a$ be a base measure on $A\in\clA$ with $\supp P\subset A$, and
%Assume $\clA$ is fitted and let
$Q\leq P$ be a simple   measure with representing forest $F$.  
We define  $P':=\a+P$, $Q':=\a+Q$, and 
 $F':=\{A\}\cup F$. Then the following statements hold:
% 
% 
% 
% Assume furthermore that $\clA$ is $P$-subadditive and $Q$ is adapted to $P$.
% Let $Q':=\a+Q$ and $P':=\a+P$. Obviously $Q'$ is simple on $F'$.
% Then:
\begin{enumerate}
\item The measure $Q'$ is simple and $F'$ is its representing $\orthoG$-forest.
\item Let $\a'\le P'$ be a base measure on $A'$.
  Then either $\a'\le \a$ or there is an $\ca\in(0,1)$ such that
  $
    \a'(\cdot \cap A') =  \a(\cdot \cap A')+\ca\a'(\cdot \cap A')
  $.
\item If  $\clA$ is $P$-subadditive then $\clA$ is $P'$-subadditive.
\item If $Q$ is $P$-adapted, then $Q'$ is $P'$-adapted.
\end{enumerate}
\end{lemma}

\begin{proofof}{Lemma \ref{lemma.base.added.adapted}}
\ada a We have $\Gr F=\supp Q\subset\supp P\subset A$
and hence $F'$ is a $\orthoG$-forest, which is obviously representing $Q$.
% Now for the other statements:

\ada b Let us assume that $\a'\not\le\a$, i.e.~there is  a $C_0\in\clB$ with $\a'(C_0)>\a(C_0)$ and thus we find
$\a'(C_0\cap A') = \a'(C_0)> \a(C_0)\geq \a(C_0\cap A')$. 
In addition, we have $A'=\supp\a'\subset\supp\a= A$, and therefore 
  Lemma \ref{lemma.ordering} shows 
$\a(\cdot \cap A')=\gamma \a'(\cdot \cap A')$, where $\gamma:= \frac {\a(C_0\cap A')}{\a'(C_0\cap A')}<1$.
Setting $\ca := 1-\gamma$ yields the assertion.

% 
% 
% Since  
% % Let $\a'\le P'$ be a base measure on the base set $A'$.
%  $A'=\supp\a'\subset\supp\a= A$, the 
%  second part of Lemma~\ref{lemma.flatness} then gives the dsired $\a$.

\ada c 
Let $\a_1,\a_2\le P'$ be base measures on sets $A_1,A_2\in \clA$ with $A_1\nPorthoPp A_2$.
Since $\supp P'=A$, we have  $A_1\cup A_2\subset A$, and thus $\a\in\clQ_{P'}(A_1\cup A_2)$.
Clearly, if $\a\geq \a_1$ or $\a\geq \a_2$, there is nothing left to prove, and
hence we assume  $\a_1\not\le\a$ and $\a_2\not\le\a$.
Then (b) gives   $\ca_i\in(0,1)$ with $\a_i(\cdot \cap A_i)=\a(\cdot \cap A_i)+\ca_i\a_i(\cdot \cap A_i)$.
We conclude that 
$ \a(\cdot \cap A_i)+\ca_i\a_i(\cdot \cap A_i) 
= \a_i(\cdot \cap A_i) \leq P'(\cdot \cap A_i) = \a(\cdot \cap A_i) + P(\cdot \cap A_i)$, 
and thus $\ca_i \a_i = \ca_i\a_i(\cdot \cap A_i) \leq P(\cdot \cap A_i) \leq P$.
Since $\clA$ is  $P$-subadditive, we thus find  
an $\tilde \a\in\clQ_{P}(A_1\cup A_2)$ with say $\tilde\a\ge\ca_1\a_1$.
For  $\tilde A:= \supp\tilde\a$
we then have 
\[
 \tilde\a' := \a(\cdot \cap \tilde A) +\tilde\a(\cdot \cap \tilde A)
 \geq \a(\cdot \cap \tilde A) +\ca_1\a_1(\cdot \cap \tilde A)
 \ge \a(\cdot \cap   A_1) +\ca_1\a_1(\cdot \cap  A_1) = \a_1\, ,
\]
where we used  $\supp\a_1 = A_1\subset \tilde A$.
% we conclude that 
% $\tilde\a' \ge \a(\cdot \cap   A_1) +\ca_1\a_1(\cdot \cap  A_1)
% = \a_1(\cdot\cap A_1) = \a_1$.
Moreover $\tilde A= \supp\tilde\a  \subset \supp P \subset A$, together with flatness of $\clQ$
shows that $\tilde\a'$ is a base measure, and we also have
$\tilde \a' \leq \a +\tilde \a \leq \a + P=P'$.
Finally we observe that $A_1\cup A_2 \subset \tilde A = \supp \tilde \a'$, and hence 
$\tilde\a'\in \clQ_{P'}(A_1\cup A_2)$.

% % 
% % Since $\tilde A:= \supp\tilde\a  \subset \supp P \subset A$, we can now define a simple measure 
% % by $Q':= \a+\tilde \a$. Let  $\tilde\a'$ be the level of $\tilde A$ in $Q'$.
% % Then the definition of levels yields
% % \[
% %   \tilde\a'= \a(\cdot \cap \tilde A) +\tilde\a \ge \a+\ca_1\a_1=\R{A_1} \a_1
% % \]
% % 
% % we then have 
% % \[
% %  \tilde\a' := \a(\cdot \cap \tilde A) +\tilde\a(\cdot \cap \tilde A)
% %  \geq \a(\cdot \cap \tilde A) +\ca_1\a_1(\cdot \cap \tilde A) \, ,
% % \]
% % and by $A_1\subset \tilde A$
% % we conclude that 
% % $\tilde\a' \ge \a(\cdot \cap \tilde A_1) +\ca_1\a_1(\cdot \cap \tilde A_1)
% % = \a_1(\cdot\cap A_1) = \a_1$.
% % 
% % % = \a_1(\cdot \cap \tilde A)
% % define $\tilde\a'= \a(\cdot \cap \tilde A) +\tilde\a$.
% % 
% % Now 
% % % 
% % % $\tilde\a\le P$
% % % on $\tilde A\supset A_1\cup A_2$ with say $\tilde\a\ge\ca_1\a_1$.
% % let $\tilde\a'$ be the level of $\tilde A:= \supp\tilde\a$ in $\a + \tilde\a$.
% % Then the definition of levels yields
% % \[
% %   \tilde\a'= \a(\cdot \cap \tilde A) +\tilde\a \ge \a+\ca_1\a_1=\R{A_1} \a_1
% % \]
% % which gives us $\tilde\a'\ge\R{\tilde A\cap A_1}\a_1$,
% % but since $\supp\a_1=A_1\subset\tilde A$ this implies $\tilde\a'\ge\a_1$ everywhere.

\ada d 
Clearly, 
$F'$ is grounded because it is a tree.
Now let $\rep Ak\in F'$, $k\ge2$ be direct siblings and 
 $\a_i'$ be their levels in $Q'$.
Since $A$ is the only root it has no siblings, so for all $i$ we have $A_i\in F$. Moreover, 
the levels $\a_i$ of $A_i$ in $Q$  are $P$-motivated
and $P$-fine since $Q$ is $P$-adapted.
% Note that this construction immediately gives $\a_i'=\a(\cdot\cap A_i)+\a_i\not\le\a$.
Now let $\b\in\clQ_{P'}(A_1\cup A_2)$ and  $B:=\supp \b$.

To check that $Q'$ is $P'$-fine, we first observe that in the case $\b\le\a$ there is nothing to prove since
$\a\in\clQ_{P'}(\rep[\cup]Ak)$ by construction. In the remaining case $\b\not\le\a$ we find a $\cb>0$
with $\b(\cdot \cap B)=\a(\cdot \cap B)+\cb\b(\cdot \cap B)$ by (b), and by $P$-fineness of $Q$, there exists 
a  $\tilde\b\in\clQ_P(\rep[\cup]Ak)$ with $\tilde \b\geq \cb\b$.
Since $\supp\tilde \b\subset \supp P\subset \supp\a$ we see that $\a+\tilde\b$ is a simple measure, and hence 
we can consider the level
 $\tilde\b'$   of $\supp\tilde\b$ in   $\a+\tilde\b$. Since  $\tilde\b'\le \a+\tilde\b\leq \a+P\leq P'$, we then obtain 
$\tilde\b'\in\clQ_{P'}(\rep[\cup]Ak)$ and for $C\in \clB$ we also have 
  \[
    \b(C) = \b(C\cap B) = \a(C\cap B)+\cb\b(C\cap B) \le \a(C\cap B)+\tilde\b(C\cap B)
    = \tilde\b'(C\cap B) \le \tilde\b'(C).
  \]
To check that $Q'$ is strictly $P'$-motivated we fix 
  the constant $\ca\in (0,1)$ appearing in the  strict $P$-motivation of $Q$.
 Then there are $\tilde \ca_i\in (0,1)$  such that  $\a(\cdot\cap A_i)+\ca\a_i = \tilde\ca_i\a'_i$.
%  and obviously $\tilde\ca_i<1$. 
 We set $\tilde\ca:=\max\{\tilde\ca_1,\tilde\ca_2 \}\in(0,1)$ and obtain
 $\a(\cdot\cap A_i)+\alpha\a_i\le \tilde\ca\a_i'$ for both $i=1,2$.
Let us first consider the case $\b\le\a$.
Since our construction yields $\a_i'=\a(\cdot\cap A_i)+   \tilde \a_i\not\le\a$, there is 
a $C_0\in \clB$ with $\a_i'(C_0) > \a(C_0)$.
This implies $\tilde \ca \a_i'(C_0) \geq \a(C_0\cap A_i) + \ca \a_i(C_0) > \a(C_0\cap A_i)\geq \b(C_0\cap A_i)$,
i.e.~$\b\not\geq \tilde\ca\a'_i$.
 Consequently, it remains to 
 consider the case $\b\not\le\a$.
%  Let  $\ca\in (0,1)$ be the constant appearing in the  strict $P$-motivation of $Q$.
%  Moreover, there are $\tilde \ca_i\in (0,1)$  such that  $\a(\cdot\cap A_i)+\ca\a_i = \tilde\ca_i\a'_i$.
% %  and obviously $\tilde\ca_i<1$. 
%  We set $\tilde\ca:=\max\{\tilde\ca_1,\tilde\ca_2\}\in(0,1)$ and obtain
%  $\a(\cdot\cap A_i)+\alpha\a_i\le \tilde\ca\a_i'$ for both $i=1,2$ uniformely.
 By (b) and $\supp \b \subset \supp P' = A$
 there is a
  $\cb\in(0,1]$ with $\b(\cdot\cap B)=\a(\cdot\cap B)+\cb\b(\cdot\cap B)$. Then 
  \[
    \cb\b= \cb\b(\cdot\cap B)=\b(\cdot\cap B)-\a(\cdot\cap B)\le P'(\cdot\cap B)-\a(\cdot\cap B)= P(\cdot\cap B)\leq P \, ,
  \]
  and since $\cb\b\in\clQ_{P}(A_1\cup A_2)$ we obtain $\cb\b\not\ge \ca  \a_i$ for $i=1,2$.
  Hence there is an event $C_0\subset\supp\b$ with $\cb\b(C_0)< \ca  \a_i(C_0)$, which
  yields $\b(C_0\cap A_i)= \a(C_0\cap A_i\cap B)+\cb\b (C_0\cap A_i) < \a(C_0\cap A)+ \ca\a_i(C_0\cap A_i) \le\tilde\ca \a'_i(C_0\cap A_i)$,
  i.e.~$\b\not\geq \tilde \ca\a'_i$.
%\fixedtodo{Philipp: we only check motivation but not strict motivation!?}
%   
%   This gives $\b = \a(\cdot\cap B)+\cb\b \not\ge \a(\cdot\cap B)+\ca\a_i$. Using 
%   $\a(\cdot\cap B)+\ca\a_i  \leq \a(\cdot\cap B)+\a_i  = \a(\cdot\cap A_i)+\a_i =  \a'_i(\cdot\cap B)\leq \a'_i$
%   
% and by $P$-motivation of $Q$ we have $\cb\b\not\le\ca_i\a_i$ for $i=1,2$ and hence:
%   \[
%     \b = \a+\cb\b \not\le \a+\ca_i\a_i = \a_i.
%     \qedhere
%   \]
\end{proofof}

\begin{proofof}{Theorem \ref{thm.extension.axioms}}
For a $P\in \bar\clS(\clA)$ and a $P$-adapted isomonotone sequence
$(Q_n,F_n)\nearrow P$ we 
define $c_{\clA}(P) :=_P \lim_{n\to \infty}s(F_n)$, which is possible by 
Theorem \ref{thm.uniqueness}. By Proposition \ref{prop.simple.measures.adapted.to.themselves}
we then now that $c_\clA(Q) = c(Q)$ for all $Q\in \clQ$, and hence 
$c_\clA$ satisfies the Axiom of BaseMeasureClustering. Furthermore, $c_\clA$ is obviously 
structured and scale-invariant, and continuity follows from Theorem~\ref{thm.uniqueness}.

To check that $c_\clA$ is disjoint-additive, we fix 
$\rep Pk\in\clP_\clA$ with pairwise $\orthoG$-disjoint supports
and let $(Q_n^i,F_n^i)\nearrow P_i$ be $P_i$-adapted isomonotone sequences of simple measures.
We set $Q_n:= Q_n^1 +\dots+Q_n^k$ and $P:=\rep[+]Pk$.
By Lemma~\ref{lemma.disjoint.simple.adapted} $Q_n$ is simple on $F_n:=F_n^1\cup \dots\cup F_n^k$ and 
 $P$-adapted,  and  $\clA$ is $P$-subadditive.
Moreover, we  have $Q_n\nearrow P$ and $s(F_n) = \bigcup_i s(F_n^i)$
inherits monotonicity as well. Therefore $(Q_n,F_n)\nearrow P$ is $P$-adapted
and $\lim s(F_n) = \bigcup_i \lim s(F_n^i)$ implies disjoint-additive.

To check BaseAdditivity we fix a $P\in \clP_\clA$ and a base measure $\a$ with $\supp P\subset \a$.
Moreover, let 
  $(Q_n,F_n)\nearrow P$ be a $P$-adapted sequence.
Let $Q_n':=\a+Q_n$ and $P':=\a+P$.
Then by Lemma~\ref{lemma.base.added.adapted} $Q_n'$ is simple on $F_n':=\{A\}\cup F_n$ and $P'$-adapted,
and $\clA$ is $P'$-subadditive.
Furthermore we have $(Q_n',F_n')\nearrow P'$ and therefore we find
% 
% $Q_n'\uparrow P':=P+\a$ and trivially $s(F_1')\le s(F_2')\le\dots$.
% Hence $(Q_n',F_n')\uparrow P'$ adapted and therefore 
$P'\in\clP_\clA$ and
  $\lim s(F'_n) = s(\{A\} \cup \lim s(F_n))$.

% \begin{description.defi}
% % \item[\DisjointAdditivity]
% % Let $\rep Pk\in\clP_\clA$ be measures with pairwise $\orthoG$-disjoint supports
% % and let $(Q_n^i,F_n^i)\uparrow P^i$ be adapted isomonotone sequences of simple measures.
% % We set $Q_n:=\sum_i Q_n^i$ for all $n$ and $P:=\rep[+]Pk$.
% % By Lemma~\ref{lemma.disjoint.simple.adapted} $Q_n$ is simple on $F_n:=F_n^1\cup F_n^2$
% % and adapted to $P$ and $P\in\clP_\clA$.
% % By additivity of measures we still have $Q_n\uparrow P$ and $s(F_n) = \bigcup_i s(F_n^i)$
% % inherits monotonicity as well. Therefore $(Q_n,F_n)\uparrow P$ adapted
% % and $\lim s(F_n) = \bigcup_i \lim s(F_n^i)$ implies the statement.
% 
% \item[\BaseAdditivity]
% Let $(Q_n,F_n)\uparrow P$ adapted.
% Let $Q_n':=\a+Q_n$ and $F_n':=F_n\cup\{A\}$.
% Then by Lemma~\ref{lemma.base.added.adapted} $Q_n'$ is simple on $F_n'$ and adapted
% and $\clA$ is $P'$-subadditive.
% Furthermore $Q_n'\uparrow P':=P+\a$ and trivially $s(F_1')\le s(F_2')\le\dots$.
% Hence $(Q_n',F_n')\uparrow P'$ adapted and therefore $P'\in\clP_\clA$ and
% we have $\lim s(F'_n) = s(\{A\} \cup \lim s(F_n)$.
% \end{description.defi}

For the uniqueness we finally observe that 
 Theorem~\ref{thm.clustering.of.simple.measures} together with the Axioms of Additivity 
 shows equality on $\clS(\clA)$ and 
 the Axiom of Continuity in combination with Theorem \ref{thm.uniqueness}
 extends this equality to $\clP_\clA$.
% 
% 
% By Theorem~\ref{thm.clustering.of.simple.measures} any continuous clustering $\tilde\cl$
% extends the clustering $Q\mapsto s(F)$ of simple measures, and hence $\clS\subset\tilde\clP$.
% For any $P\in\clP_\clA$ there is $(Q_n,F_n)\uparrow P$ and by \Continuity\ $P\in\tilde\clP$
% and $\tilde\cl(P) = \lim_n s(F_n) = \cl_\clA(P)$.
\end{proofof}

\makeatletter{}%!TEX root = article.tex

\subsubsection{Proof of Theorem \ref{thm.clustering.density.level.sets}}

\begin{lemma}\label{lemma.density.subadditivity.rich}
Let $\mu\in\clM_\Omega^\infty$, and consider
$(\clA,\clQ^{\mu,\clA},\orthoG)$.% the stable clustering based described in Proposition  \ref{prop.indicators}.
\begin{enumerate}
\item If $A,A'\in\clA$ with $A\subset A'$ $\mu$-a.s. then $A\subset A'$.
\item Let $P\in\clM_\Omega$ such that $\clA$ is $P$-subadditive and
$P$ has a $\mu$-density $f$ that is of $\clAll$-type 
with a dense subset $\Lambda$ such that $s(F_{f,\Lambda})$ is finite.
For all $\lambda\in\Lambda$ and all $\rep Ak\in\clA$ with $\rep[\cup]Ak\subset\{f>\lambda\}$ $\mu$-a.s.
there is $B\in\clA$ with $\rep[\cup]Ak\subset B$ pointwise and $B\subset\{f>\lambda\}$ $\mu$-a.s.
\end{enumerate}
\end{lemma}

\begin{proofof}{Lemma \ref{lemma.density.subadditivity.rich}}
\ada a Let $A,A'\in\clA$ with $A\subset A'$ $\mu$-a.s.\ and let $x\in A$.
Now $B:=A\setminus A'$ is relative open in $A$ and if it is non-empty then $\mu(B)>0$
since $A$ is a support set. Since by assumption $\mu(B)=0$ we have $B=\emptyset$.

\ada b %By (a) we have $\rep[\cup]Ak\subset B:=\{f>\lambda\}$ pointwise.
Since $H:=\{f>\lambda\}\in\bar\clA$ there is an increasing sequence $C_n\nearrow H$ of base sets.
Let $d\b_n:= \lambda 1_{B_n}d\,\mu\in\clQ_P$.
For all $i\le k$ eventually $B_n\nPortho\mu A_i$,
so there is a $n$ s.t.\ $B_n$ is connected to all of them.
By $P$-subadditivity between $\b_n$ and $\repx{\lambda 1_{A_{\x}}\,d\mu}k$ there is
$d\c=\lambda'1_C\,d\mu\in\clQ_P$ that supports all of them and majorizes at least one of them.
Hence $\lambda\le\lambda'$ and thus $\rep[\cup]Ak\subset C\subset \{f>\lambda'\}\subset\{f>\lambda\}$ $\mu$-a.s.
By (a) we are finished.
\end{proofof}

\begin{lemma}\label{lemma.constructing.chains}
Let $f$ be a density of $\clAll$-type, set $P:=f\,d\mu$ and assume $\clA$ is $P$-subadditive and $F_{f,\Lambda}$ is a chain.
For all $k\ge0$ and all $n\in\dsN$ let $B_n=\rep[\cup]Ck$ be a (possibly empty) union of base sets $\rep Ck\in\clA$
with $B_n\subset \set{f>\lambda}$ for all $\lambda\in\Lambda$.
Then $P:=f\,d\mu\in\clP$ and there is $(Q_n,F_n)\nearrow P$ adapted
where for all $n$ $F_n$ is a chain and $B_n\subset\min F_n$.
\end{lemma}
\begin{proofof}{Lemma \ref{lemma.constructing.chains}}
Let $(\lambda_n)_n\subset \Lambda$ be a dense countable subset with $\lambda_n<\rho$
and set $\Lambda_n := \{\lambda_1,\ldots,\lambda_n\}$, $\Lambda_\infty:=\bigcup_n \Lambda_n$.
Remark that $\max\Lambda_n<\rho$ for all $n$, $|\Lambda_n|=n$ and $\Lambda_1\subset \Lambda_2\subset \dots$.
For very $n$ we enumerate the $n$ elements of $\Lambda_n$ by $\lambda(1,n)<\ldots<\lambda(n,n)$.
For every $\lambda\in \Lambda_\infty$ we let $n_\lambda:=\min\{n\mid \lambda\in\Lambda_n\}\in\dsN$.

Since $f$ is of $\clAll$-type, $H(\lambda):=\{f>\lambda\}\in\bar\clA$ for $\lambda\in\Lambda$.
Therefore there is $A_{\lambda,n}\in\clA$ s.t.\ $A_{\lambda,n}\uparrow H(\lambda)$, where $n\ge0$.
We would like to use these $A_{\lambda,n}$ to construct $Q_n$, but they need to be made compatible
in order that $(Q_n,F_n)_n$ becomes isomonotone.
%By $P$-subadditivity we even can assume $B_n\subset A_{\lambda,n}$.
Hence we construct by induction a family of sets $A(\lambda,n)\in\clA$, $\lambda \in \Lambda_n$, $n\in\dsN$
with the following properties:
\[
  A_{\lambda,n} \cup A(\lambda(i+1,n),n) \cup A(\lambda,n-1)\cup B_n 
  \subset A(\lambda,n) \subset H(\lambda)\dotcup N(\lambda,n),
  \qquad \mu(N(\lambda,n))=0.
\]
Here $A(\lambda(i+1,n),n)$ is thought as empty if $i=n$ and similarly $A(\lambda,n-1)=\emptyset$
if $n=1$ or $\lambda\notin\Lambda_{n-1}$. All of these involved sets $C$ are base sets with $C\subset H(\lambda)$
and hence by Lemma~\ref{lemma.density.subadditivity.rich}
there is such an $A(n,\lambda)$.
Since $A_{\lambda,n}\nearrow_n H(\lambda)$ we then also have $A(\lambda,n_\lambda+n)\uparrow H(\lambda)$.

%Now assume that $A(\lambda',n')$ already is defined according to the above properties for all $n'<n$ and $\lambda'\in\Lambda_{n'}$ and for all 
%with $n'<n$ or $\lambda'>\lambda$.
%Let
%\[
%  C_\lambda := \begin{cases}
%    A_{\lambda,n} & n=n_\lambda, \\
%    A(\lambda,n-1) & n>n_\lambda \\
%  \end{cases}
%\]
%In both cases $C_\lambda\in\clA$ and $B_n\subset C_\lambda\subset H(\lambda)$
%either by definition or by induction assumption.
%Further let
%\[
%  D_\lambda:=\bigcup_{\lambda'\in\Lambda_n\colon \lambda'>\lambda} A(\lambda',n)
%  = \max\{A(\lambda',n)\mid \lambda'\in\Lambda_n, \lambda'>\lambda\}
%\]
%We have $D_\lambda\subset H(\lambda)$ by assumption and it is empty if $\lambda = \max\Lambda_n$
%and  $D_\lambda\in\clA$ else.
%Since $A_{\lambda,i}\uparrow H(\lambda)$ by the previous lemma (used twice when $D_\lambda\ne\emptyset$)
%By $P$-subadditivity there is
%\[
%  A(\lambda,n) \in\clA\colon A_{\lambda,n-n_\lambda} \cup C_\lambda \cup D_\lambda \big)
%  \subset A(\lambda,n)\subset H(\lambda).
%\]
%This gives $A(\lambda,n)\in\clA$, $B_n\subset C_\lambda\subset A(\lambda,n)\subset H(\lambda)$ and
%$A_{\lambda',n}\subset D_\lambda\subset A(\lambda,n)$
%for all $\lambda'\in\Lambda_n$ with $\lambda'>\lambda$.
%Furthermore, if $\lambda\in\Lambda_{n-1}$ then $n>n_\lambda$ and therefore
%$A(\lambda,n-1) = C_\lambda \subset A(\lambda,n)$.
%Lastly $A_{\lambda,n} \subset A(\lambda,n)$ for $n\ge n_{\lambda}$.
%So we have constructed such a sequence.
%
Now for all $n$ consider the chain $F_n:=\{A(\lambda,n)\mid \lambda\in\Lambda_n\}\subset \clA$
and the simple measure $Q_n$ on $F_n$ given by:
\[
  h_n := \sum_{i=1}^n \big(\lambda(i,n)-\lambda(i-1,n)\big) \cdot 1_{A(\lambda(i,n),n)}
  = \sum_{\lambda\in\Lambda_n} \lambda\cdot
  1_{A(\lambda,n)\setminus \bigcup_{\lambda'>\lambda}A(\lambda',n)}
  \qquad (\lambda(0,n):=0)
\]
Let $x\in B$. Let 
\[
  \Lambda_n(x) := \{ \lambda \in\Lambda_n\mid x\in A(\lambda,n)\}
\]
Then $h_n(x) = \max \Lambda_n(x)$. And if $x\in A(\lambda,n)$ then $x\in A(\lambda,n+1)$
so $\Lambda_n(x)\subset \Lambda_{n+1}(x)$ and we have:
\[
  h_n(x) = \max \Lambda_n(x)
  \le \max \Lambda_{n+1}(x) = h_{n+1}(x)
\]
Furthermore if $\lambda\in\Lambda_n(x)$ then $x\in A(\lambda,n)\subset H(\lambda)$
implying $h(x)>\lambda$.
Therefore $h_1\le h_2\le\dots\le h$.

On the other hand for all $\epsilon >0$, since $\Lambda_\infty$ is dense, there is a $n$ and
$\lambda\in\Lambda_n$ with $h(x)-\epsilon\le \lambda<h(x))$. Then $x\in H(\lambda)$
and therefore for $n$ big enough $x\in A(\lambda,n)$ and then:
\[
  h(x)\ge h_n(x)\ge \lambda\ge h(x)-\epsilon.
\]
This means $h_n(x)\uparrow h(x)$ for all $x\in B$ so we have $h_n\uparrow h$ pointwise
and by monotone convergence $(Q_n,F_n)\uparrow P_0$.
\end{proofof}

\begin{proofof}{Theorem \ref{thm.clustering.density.level.sets}}
Let $f$ be a density as supposed and set $F:=s(F_{f,\Lambda})$.
By assumption $F$ is finite.
If $|F|=1$ then $F_{f,\Lambda}$ is a chain and the Theorem follows from Lemma~\ref{lemma.constructing.chains}
using $B_n=\emptyset$, $n\in\dsN$, in the notation of the lemma.
Hence we can now assume $|F|>1$.
We prove by induction over $|F|$ that $f\,d\mu\in\bar\clS(\clA)$ and $c(f\,d\mu)=_\mu s(F_{f,\Lambda})$
and assume that this is true for all $f'$ with level forests $|s(F_{f',\Lambda'}|<|F|$.
For readability we first handle the case that $F$ is not a tree.

Assume that $F$ has two or more roots $\rep Ak$ with $k=k(0)$.
Denote by $f_i:=f\R{A_i}$ the corresponding densities, hence $f=\rep[+]fk$,
and set $F_i:=s(F_{f_i,\Lambda})=F\Rle{A_i}$ and $P_i:=f_i\,d\mu$.
We cannot use \DisjointAdditivity, because separation of the $A_i$ does not imply separation of the supports.
Hence we have to construct a $P$-adapted isomonotone sequence $(Q_n,F_n)\nearrow P$.
Since $F=\rep[\dotcup]Fk$ we have $|F_i|<|F|$ and hence
by induction assumption for all $i\le k$ we have $\cl(P_i)=F_i$,
and there is an isomonotone $P_i$-adapted sequence $(Q_{i,n},F_{i,n})\nearrow P_i$.
For $Q_n:=Q_{1,n}+\ldots+Q_{k,n}$ and $F_n:=F_{1,n}\cup\ldots\cup F_{k,n}$
it is clear that $(Q_n,F_n)\nearrow P$ is isomonotone.
Let $\b\in\clQ_P$ and $B:=\supp\b$. We show that this is $\nPortho\mu$-connected to exactly one $A_i$.
There is $\cb>0$ s.t.\ $d\b=\cb 1_Bd\mu$ and $\cb 1_B\le f$ $\mu$-a.s.
Now let $\lambda \in \Lambda$ with $\lambda<\cb$ and $\lambda<\inf\set{\lambda'\in\Lambda\mid k(\lambda')\ne k(0)}$.
Because for all $\lambda\in\Lambda$ also the closures of clusters are $\orthoG$-separated we have
\begin{align*}
  B&{}\subset \overline{H_f(\lambda)} = \repx[\orthocupG]{\overline{B_\x(\lambda))}}k.
\end{align*}
By connectedness there is a unique $i\le k$ with $B\subset \overline{B_i(\lambda)}$
and by monotonicity $B\orthoG\overline{B_j(\lambda)}$ for all $i\ne j$.
Since this holds for all $\lambda\in\Lambda$ small enough
and $\Lambda$ is dense, this means that $B$ is $\nPortho\mu$-connected to exactly $i$.
Using this, $P$-adaptedness of $Q_n$ is inherited from $P_i$-adaptedness of $Q_{i,n}$.
Therefore $P=\lim_n Q_n\in\clP$ and $\cl(P)=F$.

Now assume that $F$ is a tree. Since $|F|>1$ there are direct children $\rep Ak$ of the root in the structured forest $F$
with $k\ge2$.
Let $\rho:=\inf\{\lambda\in\Lambda\mid k(\lambda)\ne1\}$.
Since $F$ is a tree, $\rho>0$.
Let $f_0(\omega):=\min\{\rho,f(\omega)\}$ and $f'(\omega):=\max\{0,f(\omega)-\rho\}$ for all $\omega\in\Omega$,
and set $dP_0:=f_0\,d\mu$ and $dP':=f'\,d\mu$.
Then $P=P_0+P'$ is split into a \emph{podest} corresponding to the root and its chain and
the density corresponding to the children.
We set $\Lambda':=\set{\lambda-\rho\mid \lambda\in\Lambda, \lambda> \rho}$.
Then $|F_{f',\Lambda'}|=|F|-1$ and by induction assumption there is $(Q_n',F_n')\uparrow P'$ adapted.
Set $B_n:=\Gr F_n'$ and $B:=\bigcup B_n$.
Then by Lemma~\ref{lemma.constructing.chains} there is $(Q_n,F_n)\nearrow P_0$ adapted,
which is given by a density $h_n$.

Now there might be a gap $\epsilon_n:=\rho-\sup h_n>0$.
By construction $\epsilon_n\to0$ but to be precise we let
\begin{align*}
  \tilde Q_n&{}:= Q_n' + \sum_{A\in\max F_n'} \epsilon_n\cdot 1_A\,d\mu.
\end{align*}
This is still a simple measure on $F'_n$
and therefore $(Q_n+\tilde Q_n,F_n\cup F'_n)\nearrow P$. We have to show $P$-adapted:
\begin{description}
\item[Grounded:] Is fulfilled, since we consider trees at the moment.
\item[Fine:] Let $\rep Ck\in F_n\cup F'_n$ be direct siblings. Then $\rep Ck\in F'_n$ because $F_n$ is a chain.
If they are contained in one of the roots of $F'_n$ fineness is inherited from adaptedness of $Q_n'$.
Else they are the roots of $F'_n$.
Let $\a=\ca 1_Ad\mu \in\clQ_P$ be a basic measure that $\PorthoP$-intersects say $C_1$ and $C_2$.
Then is clear that $\ca\le \rho$ and by $P$-subadditivity fineness is granted.
\item[Motivated] Let $C,C'\in F_n\cup F'_n$ be direct siblings. Then again $C,C'\in F'_n$.
If they are contained in one of the roots of $F'_n$ motivatedness is inherited from adaptedness of $Q_n'$.
Else they are the roots of $F'_n$. Let $\a=\ca 1_Ad\mu\in\clQ_P$ be a base measure that supports $C_1\cup C_2$.
Again it is clear that $\ca\le \rho$ and hence it cannot majorize neither the level of $C$ nor the one of $C'$.
\qedhere
\end{description}
\end{proofof}

\begin{proofof}{Proposition \ref{prop.clustering.density.level.sets.continuous}}
Since $f$ is continuous, all $H_f(\lambda)$ are open and it is the disjoint
union of its open connected components.
We show any connected component contains at least one of the $\rep{\hat{x}}k$.
To this end let $\lambda_0\ge0$ and $B_0$ be a connected component of $H_f(\lambda_0)$
(then $B_0\ne\emptyset$).
Because $\Omega$ is compact, so is the closure $\bar B_0$, and hence the maximum of $f$ on $\bar B_0$
is attained at some $y_0\in\bar B_0$.
Since there is $y_1\in B_0$ we have $f(y_0)\ge f(y_1)>\lambda$ we have $y_0\in H_f(\lambda)$.
Now $H_f(\lambda)$ is an open set, so $y_0$ is an inner point of this open set,
and we know $y_0\in\bar B_0$, therefore $y_0\in B_0$.
Therefore $y_0\in B_0$ is a local maximum.

Hence for all $\lambda$ there are at most $k$ components and $f$ is of $(\clA,\clQ^{\mu,\clA},\orthoD)$-type.
The generalized structure $\tilde s(F_f)$ is finite, since there are only $k$ leaves.

Now, fix for the moment a local maximum $\hat{x}_i$.
Since $\hat x_i$ is a local maximum, there is $\epsilon_0$ s.t.\ $f(y)\le f(\hat x_i)$ for all $y$ with $d(y,\hat x_i)<\epsilon_0$.
For all $\epsilon\in(0,\epsilon_0)$ consider the sphere
\begin{align*}
  S_\epsilon(\lambda)&{}:=\set{y\in\Omega\colon f(y)\ge\lambda \tand d(y,\hat x_i)=\epsilon}.
\end{align*}
Since $\Omega$ is compact and $S_\epsilon(\lambda)$ is closed, it is also compact.
So as $\lambda\uparrow f(\hat x_i)$ the $S_\epsilon(\lambda)$ is a monotone decreasing sequence of compact sets.
Assume that all $S_\epsilon(\lambda)$ were non-empty:
Let $y_n\in S_{\epsilon_0/(n+1)}(\lambda)$ then $(y_n)_n$ is a sequence in the compact set $S_{\epsilon_0/2}(\lambda)$,
hence there would be a subsequence converging to some $y_\epsilon$.
This subsequence eventually is in every $S_{\epsilon_0/(n+1)}$ and
hence $y_\epsilon\in\bigcap_{\lambda<f(\lambda)} S_\epsilon(\lambda)$, so this would be non-empty.
This means that $f(y_\epsilon)\ge f(\hat x_i)$.
On the other hand, since $\epsilon<\epsilon_0$ we have $f(y_\epsilon)\ge f(\hat x_i)$.
Therefore all $y_\epsilon$ are local maxima, yielding a contradiction to the assumption that there are only finitely many.
Hence for all $\epsilon$, $S_\epsilon(\lambda)=\emptyset$ for all $\lambda\in(\lambda_\epsilon,f(\hat x_i)$.
From this follows, that all local maxima have from some point on their own leaf in $F_f$.
Therefore there is a bijection $\psi\colon\{\rep{\hat{x}}k\}\to\min\cl(P)$
s.t.\ $\hat{x}_i\in \psi(\hat{x}_i)$.

Lastly, we need to show that also the closures of the connected components are separated,
to verify the conditions of Theorem~\ref{thm.clustering.density.level.sets}.
We are allowed to exclude a finite set of levels, in this case the levels $\rep\lambda m$
at which $\lambda\mapsto k(\lambda)\in\dsN$ changes.
Consider $0\le\lambda_0<\lambda_1$ s.t.\ for all $\lambda\in(\lambda_0,\lambda_1)$ $k(\lambda)$ stays constant.
Set $\tilde\lambda:=\frac{\lambda_0+\lambda}2\in(\lambda_0,\lambda)$.
Now let $A,A'$ be connected components of $H_f(\lambda)$ and let $B,B'$ be the connected components of $H_f(\tilde\lambda)$
with $A\subset B$ and $A'\subset B'$.
First we show $\bar A\subset B$: let $y_0\in\bar A$. Then there is
$(y_n)\subset A$ with $y_n\to y_0$. Because $f$ is continuous we have
\[
  \lambda< f(y_n) \to f(y_0)\ge \lambda > \tilde \lambda
\]
and hence $y_0\in B$.
Similarly we have $\bar A'\subset B'$ and $B\orthoD B'$ implies $\bar A\orthoD \bar A'$.
\end{proofof}

\makeatletter{}\subsection{Proofs for Section \ref{section.examples}}

\begin{lemma}\label{lemma.intersection.connectivity}
Let $A,A'$ be closed, non-empty, and (path-)connected. Then:
\[
   \text{$A\cup A'$ is (path-)connected} \iff A\northoD A'.
\]
Therefore any finite or countable union $\rep[\cup]Ak$, $k\le\infty$ of such sets
is connected iff the graph induced by the intersection relation is connected.
\end{lemma}

\begin{proofof}{Lemma \ref{lemma.intersection.connectivity}}
Topological connectivity means that $A\cup A'$ cannot be written as
disjoint union of closed non-empty sets.
Hence, if $A\cup A'$ is connected, then this union cannot be disjoint.
On the other hand if $x\in A\cap A'\ne\emptyset$ and
$A\cup A' = B\cup B'$ with non-empty closed sets then $x\in B$ or $x\in B'$.
Say $x\in B$, then still $B'$ has to intersect $A$ or $A'$, say $B'\cap A\ne\emptyset$.
Then both $B,B'$ intersect $A$ and both $C:=B\cap A$ and $C':=B'\cap A$ are closed and non-empty.
But since $A=C\cup C'$ is connected there is $y\in C\cap C'\subset B\cap B'$ and
therefore $B\cup B'$ is not a disjoint union.

For path-connectivity: If $x\in A\cap A'\ne\emptyset$ then for all $y\in A\cup A'$
there is a path connecting $x$ to $y$, so $A\cup A'$ is path-connected.
On the other hand, if $A\cup A'$ is path connected then for any $x\in A$ and $x'\in A'$
there is a continuous path $f\colon [0,1]\to A\cup A'$
connecting $x$ to $x'$. Then $B:=f\1(A)$ and $B':=f\1(A')$ are closed and non-empty,
and $B\cup B'=[0,1]$. Since $[0,1]$ is topologically connected there is $y\in B\cap B'$
and so  $f(y) \in A\cap A'$.
\end{proofof}

\begin{proofof}{Example \ref{example.separation.relations}}
Reflexivity and monotonicity are trivial for all the three relations.
%\begin{description.defi}
\emph{Disjointness}: Stability is trivial and connectedness follows from Lemma~\ref{lemma.intersection.connectivity} and from the observation:
\begin{align*}
  A&{}\subset \rep[\orthocupD] Bk
  &\Rightarrow\qquad A &{}= %\orthobigcupD_i (A\cap B_i)
  \repx[\orthocupD]{(A\cap B_\x)}k
\end{align*}
%%%%%%
\emph{$\tau$-separation}: Connectedness follows from the definition of $\tau$-connectedness.
For stability let $A_n\uparrow_n A$ and $A_n\orthoT B$ for $n\in\dsN$ and observe
\begin{align*}
d(A,B) = \sup_{x\in A} d(x,B) = \sup_{n\in\dsN} \sup_{x\in A_n} d(x,B)
= \sup_{n\in\dsN} d(A_n,B)\ge\tau.
\end{align*}
%%%%%%
\emph{Linear Separation}: Connectedness follows from the condition on $\clA$
since $A\subset \rep[\orthocupL]Bk$ implies $A=\rep[\orthocupL]{A\cap B}k$.
To prove stability let $A_n\uparrow_n A$ and $A_n\orthoL B$ for $n\in\dsN$.
Observe that
$$v\mapsto\sup\{\alpha\in\dsR\mid \<v\mid a>\le \alpha \forall a\in A\}$$
is continuous and the same holds for the upper bound for the $\alpha$.
Hence for each $n$ and any vector $v\in H$ with $\<v\mid v>=1$
there is a compact, possibly empty interval $I_n(v)$
of $\alpha$ fulfilling the separation along $v$.
Since by assumption the unit sphere is compact
so is the semi-direct product $I_n:=\{(v,\alpha)\mid \alpha\in I_n(v)\}$.
Since $I_n\ne\emptyset$ and $I_n\supset I_{n+1}$ is
a monotone limit of non-empty compact sets,
the limit $\bigcap_n I_n$ is non-empty.
%\qedhere
%\end{description.defi}
\end{proofof}

\begin{lemma}\label{lemma.intersection.graph.support.sets}
Let $\mu\in\clM_\Omega^\infty$.
If $\clC\subset\clK(\mu)$ then $\dsC_{\PorthoG}(\clC)\subset \clK(\mu)$.
\end{lemma}

\begin{proofof}{Lemma \ref{lemma.intersection.graph.support.sets}}
Let $A=\rep[\cup]Ck\in\dsC_{\PorthoG}(\clC)$ then:
\[
  \supp 1_A\,d\mu = \supp(1_{C_1}+\dots+1_{C_k})\,d\mu
  = \rep[\cup]Ck = A.
  \qedhere
\]
\end{proofof}

\begin{lemma}\label{lemma.intersection-additivity.of.dsC}
Let $\clC\subset\clB$ be a class of non-empty closed sets.
We assume the following generalized stability:
If $B\in\clB$ and $\rep Ak\in\clC$ form a connected subgraph of $\clG_{\PorthoG}(\clC)$:
%$A:=\rep[\cup]Ak\in\dsC_{\PorthoG}(\clC)$
%then for all $B\in\clB$:
\[
  A_i\PorthoG B \quad\forall i\le k \then A\PorthoG B.
\]
Then $\dsC_{\PorthoG}(\clC)$ is $\PorthoG$-intersection additive.
Furthermore the monotone closure $\overline{\dsC_{\PorthoG}(\clC)}$ is
$$\bar\dsC_{\PorthoG}(\clC):= \{\, \iseq\cup C \mid \iseq,C\in \clC \text{ and the graph
      $\clG_{\PorthoG}(\{\iseq,C\})$ is connected} \,\}$$
\end{lemma}

\begin{proofof}{Lemma \ref{lemma.intersection-additivity.of.dsC}}
Let $A=\rep[\cup]Cn,A'=\rep[\cup]{C'}{n'}\in\dsC(\clC)$ with $A\nPorthoG A'$.
If for all $j\le n'$ we had $C'_j\PorthoG A$ then by assumption $A'\PorthoG A$
and therefore there has to be $j\le n'$ with $C'_j\nPorthoG A$.
By the same argument there then is $i\le n$ with $C_i\nPorthoG C_j$.
Therefore the intersection graph on $\rep Cn,\rep{C'}{n'}$ is connected and
\[
  A\cup A' = \rep[\cup]Cn \cup \rep[\cup]{C'}{n'} \in \dsC(\clC).
\]

Let $B\in \overline{\dsC(\clC)}$ and $\iseq,A\in\dsC(\clC)$ with $A_n\uparrow B$.
Then for all $n$ we have $A_n = C_{n1}\cup\ldots\cup C_{nk(n)}$ with $C_{nj}\in\clC$ and their
intersection graph is connected.
Since $A_n\subset A_{n+1}$ for all $C_{nj}$ there is $j'$ with $C{nj}\subset C_{(n+1),j'}$
which even gives $C_{nj}\nPorthoG C_{(n+1)j'}$.
Hence, the family $\{ C_{nj} \}_{n,j}$ being countable can be enumerated $\iseq,{\tilde C}$
s.t.\ for all $m$ there is $i(m)<m$ with $C_m\nPorthoG C_{i(m)}$.
Therefore for all $m$, the intersection graph on $\rep{\tilde C}m$ is connected
and hence
\[
  \tilde A_m := \rep[\cup]{\tilde C}m \in \dsC(\clC).
\]
And we see that $\bigcup_m \tilde A_m\in\bar\dsC(\clC)$ and therefore
\[
  B = \bigcup_n A_n = \bigcup_{nj} C_{nj} = \bigcup_m \tilde C_m \in \bar\dsC(\clC).
\]

Now let $B\in\bar\dsC(\clC)$ and $B=\bigcup_n C_n$ with $C_n\in\clC$ and
s.t.\ the intersection graph on $\iseq,C$ is connected.
By Zorn's Lemma it has a spanning tree.
Since there are at most countable many nodes, one can assume
that this tree is locally countable and therefore there is
an enumeration of the nodes $C_{n(1)},C_{n(2)},\ldots$ s.t.\ they
form a connected subgraph for all $m$.
Then the intersection graph on $C_{n(1)},\ldots,C_{n(m)}$
is connected for all $m$ and therefore
$A_m:=C_{n(1)}\cup\ldots\cup C_{n(m)}\in\dsC(\clC)$.
$A_m\in\dsC_i(\clC)\uparrow B$ is monotone and we have $B=\bigcup A_m\in\overline{\dsC_i(\clC)}$.
\end{proofof}

\begin{prop}\label{prop.separation.extended}
Let $\clC\subset \clB$ be a class of non-empty, closed events and
$\orthoG$ a $\clC$-separation relation.
We assume the following generalized countable stability:
If $B\in\clB$ and $\iseq,A\in\clC$ form a connected subgraph of $\clG_{\orthoG}(\clC)$:
\[
  A_n\orthoG B \quad\forall n \then \bigcup_n A_n\orthoG B.
\]
Then $\orthoG$ is a $\dsC_{\orthoG}(\clC)$-separation relation.
\end{prop}

\begin{proofof}{Proposition \ref{prop.separation.extended}}
Set $\tilde\clA:=\dsC_{\orthoG}$.
The assumption assures $\tilde\clA$-stability.
We have to show $\tilde\clA$-connectedness.
So let $A\in\tilde\clA$ and $\rep Bk\in\clB$ closed with:
\[
  A \subset \rep[\orthocupG] Bk.
\]
By definition of $\dsC$ there are $\rep Cn\in\clC$ with $A=\rep[\cup]Cn$
and s.t.\ the $\orthoG$-intersection graph on $\{\rep Cn\}$ is connected.
For all $j\le n$ we have $C_j\subset A\subset \rep[\cup]Bk$
and by $\clC$-connectedness there is $i(j)\le k$ with $C_j\subset B_{i(j)}$.
Now, whenever $i(j)\ne i(j')$ since $B_{i(j)}\northoG B_{i(j')}$
we have by monotonicity $C_j\northoG C_{j'}$.
\Use{Ortho:Monotonicity:A}
So whenever there is an edge between $C_j$ and $C_{j'}$ then $i(j)=i(j')$.
This means that $i(\cdot)$ is constant on connected components of the graph,
and hence on the whole graph.
\end{proofof}

\begin{prop}\label{prop.separation.main}
Let $\clC\subset \clB$ be a class of non-empty, closed events and
$\orthoG$ a $\clC$-separation relation with the following alternative $\dsC_{\orthoG}(\clC)$-stability:
For all $\iseq,A\in\clC$ and $B\in\clB$:
\begin{gather}
  \text{$\clG_{\PorthoG}(\{\iseq,A\})$ is connected and for all $n$}\colon
  A_n\orthoG B
  \then \bigcup_n A_n\orthoG B.
  \label{eq.stability.alternative}
\end{gather}
Then $\orthoG$ is a $\dsC_{\orthoG}(\clC)$-separation relation and
$\dsC_{\orthoG}(\clC)$ is $\orthoG$-intersection additive.

Assume furthermore $\PorthoG$ is a weaker relation ($B\orthoG B' \then B\PorthoG B'$).
Then $\orthoG$ is a $\dsC_{\PorthoG}(\clC)$-separation relation and
$\dsC_{\PorthoG}(\clC)$ is $\PorthoG$-intersection additive.
\end{prop}

\begin{proofof}{Proposition \ref{prop.separation.main}}
The first part is a corollary of Lemma~\ref{lemma.intersection-additivity.of.dsC}
and Proposition~\ref{prop.separation.extended}.
For the second part observe $\dsC_{\PorthoG}(\clC)\subset\dsC_{\orthoG}(\clC)$.
hence $\orthoG$ is also a $\dsC_{\PorthoG}(\clC)$-separation relation.
But now $\dsC_{\PorthoG}(\clC)$ is only $\PorthoG$-intersection additive.
\end{proofof}%

\begin{proofof}{Proposition \ref{prop.neighborhood.base}}
First if $A_n\uparrow B\in\bar\clA$ then for all $x,x'\in B$ there is $n$ with $x,x'\in A_n$
and since $A_n$ is path-connected there is a path connecting $x$ and $x'$ in $A_n\subset B$,
so they are connected also in $B$.

Let $O$ be open and path-connected.
Let $(A_n)_n\subset \clA'$ be the subsequence of all $A\in\clA'$ with $A\subset O$.
Since $O$ is open and $\clA'$ a neighborhood base $O=\bigcup_n A_n$.
Consider the graph on the $(A_n)_n$ given by the intersection relation.
Then by Zorn's Lemma there is a spanning tree, and we can assume
that it is locally at most countable.
Therefore there is an enumeration $\iseq,{A'}$
such that $\{\rep{A'}n\}$ is a connected sub-graph for all $n$.
By intersection-additivity hence $\tilde A_n:=\rep[\cup]{A'}n\in\clA$
and $\tilde A_n\uparrow O$.
\end{proofof}

\begin{lemma}\label{lemma.clustering.indicator.functions}
Let $\mu\in\clM^\infty_\Omega$ and assume there is a $B\in\clK(\mu)$ with $dP=1_B\,d\mu$.
Assume that $(\clA,\clQ^{\mu,\clA},\orthoA)$ is a $P$-subadditive stable clustering base and $(Q_n,F_n)\uparrow P$ is adapted. Then
$s(F_n)=\{\rep{A^n}{k}\}$ consists only of roots
and can be ordered in such a way that $A_i^1\subset A_i^2\subset\ldots$.
The limit forest $F_\infty$ then consists of the $k$ pairwise $\orthoA$-separated sets:
\[
  B_i := \bigcup_{n\geq 1} A_i^n\, ,
\]
there is a $\mu$-null set $N\in\clB$ with
\begin{equation}
  B = \rep[\orthocupA]Bk\orthocupD N.
\end{equation}
\end{lemma}

\begin{proofof}{Lemma \ref{lemma.clustering.indicator.functions}}
Once we have shown that all $s(F_n)$ only consists of their roots,
the rest is a direct consequence of the isomonotonicity, and the fact
that there is a $\mu$-null set $N$ s.t.:
\[
  B = \supp P = N\orthocupD \bigcup_n \supp Q_n = \rep[\orthocupA]Bk\orthocupG N.
\]
Now let $A,A'\in F_n$ be direct siblings and denote by $\a=,\a'\le P$ their levels in $Q_n$.
Then there are $\alpha,\alpha'>0$ with $\a=\alpha 1_A\,d\mu$ and $\a'=\alpha'1_{A'}\,d\mu$.
Now, $\a,\a'\le P$ implies $\alpha 1_A,\alpha' 1_{A'}\le 1_B$ ($\mu$-a.s.) and hence $\alpha,\alpha'\le 1$.
Assume they have a common root $A_0\in \max F_n$, i.e.\ $A\cup A'\subset A_0 \subset B$.
Then $\alpha 1_A,\alpha' 1_{a'}\le 1_{A_0}\le 1_B$ ($\mu$-a.s.) and hence they cannot be motivated.
%
%
%The link to the the previous section is performed naturally via
%$dP=1_C\,d\mu$ and $dQ_A:=1_A\,d\mu$ for all $A\in\clA$.
%Furthermore the notions introduced in the first section reduce in the following ways:
%\begin{itemize}
%\item Firstly, by simple arguments as in Lemma~\ref{lemma.support.general} and because
%$A,C\in\clK$
%\[
%  Q_A \le P \iff \mu(A\setminus B)=0 \iff A\subset B.
%\]
%
%\item Remember that we are interested in approximating $P$ by simple measures $Q_n$ on forests $F_n$.
%Now, in the present setting, if $A\subset A'\subset B$ then $\ca Q_A+\ca'Q_{A'}\le P$
%iff $\ca+\ca'\le 1$ and we even have
%\[
%  \ca Q_A+\ca'Q_{A'}\le  Q_{A'}\le P.
%\]
%So $Q_{A}$ is a better approximation with a simpler forest, namely just a root.
%This is why we consider in this section only forests that consist just of their roots.
%
%\item %$\iseq\le F$ holds iff:% for all $n$:% and all $A_i^n\in F_n$:
%Now let $F,F'$ be such forests only consisting of their roots. Then
%\[
%  F\le F'\iff
%  \forall A\in F\colon \exists ! A'\in F' \colon A\subset A'.
%\]
%\item Let $\iseq\le F$ be an isomonotone sequence of forests consisting only of roots.
%Then there is a $k\in\dsN$ and each forest can be enumerated as $F_n=\{\rep{A^n}k\}$ 
%in such a way that $A_i^n\subset A_i^{n+1}$ for all $n\in\dsN$ and $i\le k$.
%Furthermore $F_\infty:=\lim_n F_n$ consists of $k$ pairwise
%disjoint sets $F_\infty = \{\rep Bk\}$ with $B_i:=\bigcup_n A_i^n$ for all $i\le k$.
%
%To conclude $(Q_n,F_n)\uparrow P$ then indeed means that there is a $\mu$-null set $N$ with:
%\[
%  B = \rep[\dotcup]Bk \dotcup N.
%\]
%\end{itemize}
\end{proofof}

% 
% \begin{lemma}\label{lemma.topological.connectedness.monotone.limit}
% Assume all $A\in\clA$ are topologically connected.
% Then all $B\in\bar\clA$ are topologically connected.
% \end{lemma}
% 
% \begin{proofof}{Lemma \ref{lemma.topological.connectedness.monotone.limit}}
% Let $\clA\ni A_n\uparrow B\in\bar\clA$.
% Let $C_1,C_2\subset B$ be closed in the relative topology and both non-empty
% with $C_1\cup C_2=B$.
% If we can produce a $x\in C_1\cap C_2$ then $B$ is topologically connected.
% 
% Because $C_1\cup C_2=B=\bigcup_n A_n$ there is $n$ s.t.\ $A_n$ intersects both $C_i$.
% Let $\tilde C_i:=A\cap C_i$.
% Then both $\tilde C_i$ are closed in the relative topology and non-empty.
% Since $A_n$ is topologically connected there is
% \[
%   x\in\tilde C_1\cap\tilde C_2\subset C_1\cap C_2.
%   \qedhere
% \]
% \end{proofof}

\begin{proofof}{Lemma \ref{lemma.bilipschitz}}
The Hausdorff-dimension is calculated in \cite[Corollary 2.4]{Falconer-1993}.
Proposition~2.2 therein gives for all events $B\subset C$ and $B'\subset C'$:
\begin{align*}
  \clH^s(\phi(B)) &{}\le c_2^s\clH^s(B)
  \tand
  \clH^s(\phi\1(B')) \le c_1^s\clH^s(B').
\end{align*}
We show that $C'$ is a $\clH^s$-support set.
Let $B'\subset C'$ be any relatively open set and set $B:=\phi\1(B')\subset C$.
Then $B\subset C$ is open because $\phi$ is a homeomorphism.
And since $C$ is a support set we have $0<\clH^s(B)<\infty$. This gives
\begin{align*}
  0 < \clH^s(B) = \clH^s(\phi\1(B')) \le c_1^s\clH^s(B') \tand
  \clH^s(B') = \clH^s(\phi(B)) \le c_2^s\clH^s(B) < \infty.
\end{align*}
Therefore $C'$ is a $\clH^s$-support set.
\end{proofof}

\begin{proofof}{Proposition \ref{prop.base.sets.aggregated}}
The proof is split into four steps:
\ada a We first show that for all $A\in\clA$ there is a unique index $i(A)$ with $A\in \clA^{i(A)}$.
To this end, we fix an 
 $A\in\clA$. Then there is $i\le m$ with $A\in\clA^i$.
Let $\mu\in\clQ^i$ be the corresponding base measure with $\supp \mu=A$.
Let $j\le m$ and $\mu'\in\clQ^j$ be another measure with $\supp \mu'=A$.
Then $\mu(A)=1$ and $\mu'(A)=1$.
If $j>i$ then by assumption $\mu\prec \mu'$ and this would give $\mu'(A)=0$.
If $j<i$ we have $\mu'\prec \mu$ and this would give $\mu(A)=0$.
So $i=j$.

\ada b Next we show that for all $A,A'\in\clA$ with  $A\subset A'$ we have  $i(A)\le i(A')$.
To this end we first observe that 
% Let $A\in\clA^i$ and $A'\in\clA^j$ with $A\subset A'$.
% Now 
$A=A\cap A'=\supp Q_{A}\cap\supp Q_{A'}$.
If we had $i>j$ then $Q_{A'}\in\clQ^j\prec \clQ^i\ni Q_{A}$ and
since $Q_{A'}(A)\le Q_{A'}(A')=1<\infty$ we would have $Q_A(A)=0$.
Therefore $i\le j$.

\ada c Now we show that  $\orthoG$ is a stable $\clA$-separation relation. Clearly, it suffices to 
 show $\clA$-stability and $\clA$-connectedness.
The former follows since $i(A_n)$ is monotone if $\iseq\subset A$ by (b)
and hence eventually is constant.
For the latter let $A\in\clA^i$ and $\rep Bk\in\clB$ closed with
$A\subset \rep[\orthocupG]Bk$. Then since $\orthoG$ is an $\clA^i$-separation relation there
is $j\le k$ with $A\subset B_j$.

\ada d Finally, we show that $(\clA,\clQ,\orthoG)$ is a stable clustering base. To this end observe that
fittedness is inherited from the individual clustering bases.
Let $A\in\clA^i$ and $A'\in\clA^j$ with $A\subset A'$. Then $i\le j$ by (b).
If $i=j$ then flatness follows from flatness of $\clA^i$.
If $i<j$ then by assumption $Q_A\prec Q_{A'}$ and because $Q_A(A)=1<\infty$
we have $Q_{A'}(A) = 0$.
%\qedhere
\end{proofof}%

\begin{proofof}{Proposition \ref{prop.unions.are.good}}
\ada a Let $\a\le P$ be a base measure on $A\in\clA^i$.
If $i=1$ then $Q_A(A\cap\supp P_2) \le Q_A(A)=1$ and by $\clA^1\prec P_2$ we have $Q_A\prec P_2$
and hence $P_2(A\cap\supp P_2) = P_2(A)=0$. Now for all events $C\in A^c$
therefore $\a(C)=0\le P_1(C)$ and for all $C\subset A$:
\[
  \a(C) \le P(C) = \ca_1 P_1(C)+\ca_2 P_2(C) = \ca_1P_1(C).
\]

Now if $i=2$ then by assumption $P_1\prec \a$ and since $0<P_1(A\cap\supp P_1)<\infty$
we therefore have $\a(A\cap\supp P_1)\le \a(\supp P_1)=0$ and for all events
$C\subset \Omega\setminus\supp P_1$ we have $\a(C)\le P(C) = \ca_2 P_2(C)$
and for all events $C\subset \supp P_1$:
\[
  \a(C) \le \a(\supp P_1)=0 \le P_1(C).
\]
\ada b Let $\a,\a'\le P$ be base measures on $A\in\clA^i$ and $A\in\clA^j$ with $A\northoA A'$.
By the previous statement we then already have $\a\le \ca_i P_i$ and $\a'\le \ca_j P_j$.
Now, if $i=j$ then by $P_i$-subadditivity of $\clA^i$
there is a base measure $\b\le P_i\le P$ on $B\in\clA^i$ with $B\supset A\cup A'$.

Now if $i\ne j$ consider say $i=2$ and $j=1$.
Since $A\cap \supp P_2 \supset A\cap A'\ne\emptyset$ by assumption $\a$ can be majorized by a base measure
$\tilde \a\le P_2$ on $\tilde A\in\clA^2$ with $\supp P_1\subset\tilde A$ and $\tilde \a\ge \a$.
The latter also gives $A\subset \tilde A$ and
hence $\tilde a$ supports $A$ and $\supp P_1\supset \supp \a'$ and $\tilde \a\ge \a$.
\end{proofof}

\acks{This work has been supported by DFG Grant STE 1074/2-1. We thank the reviewers and editors for their helpful comments.}

%\clearpage
\appendix

\makeatletter{}\section{Appendix: Measure and Integration Theoretic Tools}

\makeatletter{}% \section{General Lemmata and Proofs}
% \label{section.proofs}

%\subsection{Auxiliary Results Related to Supports of Measures}
\label{app.support}

Throughout this subsection,  $\Omega$ is a Hausdorff space and $\clB$ is its 
Borel $\sigma$-algebra. 
Recall 
  that a measure $\mu$ on $\clB$ is inner regular iff for all $A\in\clB$ we have 
\[
  \mu(A) = \sup\set{\mu(K)\mid K\subset A \text{ is compact}} \, .
\]
A Radon space is a topological space such that  all finite measures are inner regular.
\citet[Theorem 8.6.14]{Cohn-2013} gives several examples of such spaces such as 
\emph{a)} Polish spaces, i.e.~separable spaces whose topology can be described by a complete metric, 
\emph{b)} open and closed subsets of Polish spaces, and 
\emph{c)} Banach spaces equipped with their weak topology.
In particular all separable Banach spaces equipped with their norm topology are Polish spaces and 
infinite dimensional spaces equipped with the weak topology are not Polish spaces but still they are Radon spaces. 
% \begin{itemize}
% \item Polish spaces: separable and the topology is equivalent to a complete metric.
% E.g.\ any separable Banach space with the norm-topology.
% Furthermore any of open or closed subset of a Polish space.
% Therefore $(0,1)$ with the standard topology is a Polish space even though it is not complete in the standard metric.
% \item Banach spaces with its weak topology.
% If it is infinite dimensional, then the topology is not metrizable, so these are examples
% of Radon spaces that are not Polish spaces.
% \end{itemize}
Furthermore Hausdorff measures, which are considered in Section~\ref{sub.examples.hausdorff}, 
are inner regular \citep[Cor.~2.10.23]{Federer}.
For any inner regular measure $\mu$ we define the \begriff{support} by 
% 
% \begin{defi}\label{defi.support}
% Let $\Omega$ be a Hausdorff space and $\mu$ be an inner regular measure on $(\Omega,\clB)$ where
% $\clB$ contains the Borel $\sigma$-algebra.
% Then the \begriff{support} of $\mu$ is:
\[
  \supp \mu
  := \Omega\setminus \bigcup\set{ O \subset \Omega\mid \text{$O$ is open and $\mu(O)=0$} }.
\]
% \end{defi}
By definition the support is closed and hence measurable. 
The following lemma collects some more basic facts about the support that are used throughout this paper.

\begin{lemma}\label{lemma.support.general}
Let $\mu$ be an inner regular measure and $A\in \clB$. Then we have:
\begin{enumerate}
% \item $\supp \mu$ is closed and hence measurable.
\item If  $A\orthoD\supp \mu$, then we have $\mu(A)=0$.
\item If $\emptyset\ne A\subset \supp\mu$ is relatively open in $\supp \mu$, then $\mu(A)>0$.
\item If $\mu'$ is another inner regular measures  and $\alpha,\alpha'>0$ then 
\[
  \supp(\alpha\mu+\alpha'\mu') = \supp(\mu) \cup \supp(\mu')
\]
\item The restriction $\mu_{|A}$ of $\mu$ to $A$ defined by $\mu_{|A}(B)=\mu(B\cap A)$ is 
    an inner regular measure and $\supp\mu_{|A} \subset \overline{A\cap \supp \mu}$.
\end{enumerate}
If $\mu$ is not inner regular, (d) also holds
provided that  $\Omega$ is a Radon space and $\mu(A)<\infty$.
\end{lemma}

\begin{proofof}{Lemma \ref{lemma.support.general}}
% \ada a Unions of open sets are open and hence the support is closed.% as it's complement.
\ada a We show that $A:=\Omega \setminus \supp \mu$ is a $\mu$-null set.
Let $K\subset A$ be any compact set.
By definition $A$ is the union of all open sets $O\subset \Omega$ with $\mu(O)=0$.
So those sets form an open cover of $A$ and therefore of $K$. 
Since $K$ is compact there exists a finite sub-cover $\left\{O_1,\dots,O_n\right\}$ of $K$.
By $\sigma$-subadditivity of $\mu$ we find
\[
  \mu(K) \le \sum_{i=1}^n \mu(O_i) = 0,
\]
and since this holds for all such compact $K\subset A$ we have by inner regularity
\[
  \mu(A) = \sup_{K\subset A} \mu(K) = 0.
\]

\ada b By assumption there 
% is Let $A\subset \supp \mu$ be relatively open in $\supp \mu$.
% Then there is 
an open   $O\subset\Omega$ with $\emptyset\neq A=O\cap \supp \mu$.
Now $O\cap \supp \mu\neq \emptyset$ implies $\mu(O)>0$.
Moreover, we have  the partition $O=A \cup (O\setminus \supp \mu)$ and since $O\setminus \supp \mu$
is open, we know $\mu(O\setminus \supp\mu) = 0$, and hence we conclude that $\mu(O) = \mu(A)$.

\ada c This follows from the fact that for all open $O\subset \Omega$ we have 
\begin{align*}
  (\alpha\mu+\alpha'\mu')(O)&{}= \alpha\mu(O)+\alpha'\mu'(O) =0 \iff \mu(O) = 0 \tand \mu'(O)=0.
\end{align*}

\ada d The measure $\mu_{|A}$ is inner regular since for $B\in\clB$ we have 
\begin{align*}
  \mu'(B) %= \mu(B\cap A)
  = \sup\set{\mu(K') \mid K'\subset B\cap A\text{ is compact}}
%   \\&{}= \sup\set{\mu'(K') \mid K'\subset B\cap A\text{ is compact}}
  &\le \sup\set{\mu'(K') \mid K'\subset B\text{ is compact}}
 \\&{} \le \mu'(B).
\end{align*}
Now observe that $X\setminus\overline{A\cap \supp \mu} \subset X\setminus (A\cap \supp \mu) = (X\setminus A) \cup (X\setminus \supp\mu)$.
For the  open set   $O:=X\setminus\overline{A\cap \supp \mu}$ we thus find 
\[
\mu_{|A}(O) \leq \mu_{|A}(X\setminus A) + \mu_{|A}(X\setminus \supp\mu) \leq \mu(X\setminus \supp\mu)  = 0. \qedhere
\]
% 
% For the second statement observe that for any open set $O\subset A^c$ we have $\mu'(O)=0$
% and hence for all open sets:
% \begin{align*}
%   O \subset (\overline{\supp \mu \cap A})^c &{}\iff
%   O \subset (\supp \mu \cap A)^c \iff
%   O \subset (\supp \mu)^c \cup A^c
%   \\ \then \mu'(O) = 0
%   &{}\iff O\subset (\supp\mu')^c
% %  \iff \mu'(O) = 0
% %  \iff O \subset (\supp \mu \cap A)^c
% %  \qedhere
% \end{align*}
\end{proofof} 
\makeatletter{}

\begin{lemma}\label{lemma.measures.compare}
Let $Q,Q'$ be $\sigma$-finite measures.
\begin{enumerate}
\item If $Q$ and $Q'$ have densities $h,h'$ with respect to some measure $\mu$ then
  \[ Q\le Q' \iff h\le h' \quad\text{$\mu$-a.s.} \]
\item If $Q\le Q'$ then $Q$ is absolutely continuous with respect to  $Q'$, i.e. 
  $Q$ has a density function $h$ with respect to $Q'$, $dQ = h\,dQ'$ such that:
	\[ h(x)  =
	\begin{cases}
	\in [0,1] &\text{ if $x \in \supp Q'$}\\
	0 &\text{ else}
	\end{cases}
  \]
%\item $Q\le Q'$ iff there is a measure $\mu$ and measurable functions
%$0\le h\le h'$ $\mu$-almost surely, such that $dQ = h\,d\mu$ and $dQ'=h'\,d\mu$ and $h'(x)=0$ for all $x\notin\supp Q'$.
\end{enumerate}
\end{lemma}

\begin{proofof}{Lemma \ref{lemma.measures.compare}}
\ada a "`$\Leftarrow$"' a direct calculation gives 
\[
  Q(B)= \int_B h\,d\mu \leq \int_B h'\,d\mu = Q'(B).
\]
and monotonicity of the integral.\\
For "`$\Rightarrow$"' assume  $\mu(\{x:h(x)>h'(x)\}) > 0$, then
\[
\int_{h>h'}hd\mu = Q(\{h>h'\}) \leq Q'(\{h>h'\}) = \int_{h>h'}h'd\mu < \int_{h>h'}hd\mu,
\]
where the last inequality holds since we assume $\mu(\{x:h(x)>h'(x)\} > 0$ and again the monotonicity of the integral. Through this contradiction implies the statement.

\ada b $Q\le Q'$ means every $Q'$-null set is a $Q$-null set.
Furthermore since $Q'$ is $\sigma$-finite $Q$ is $\sigma$-finite as well.
So we can use Radon-Nikodym theorem and there is a $h\ge 0$
s.t.\ $dQ=h\,dQ'$. Since the complement of $\supp Q'$ is a $Q'$-null set,
we can assume $h(x)=0$ on this complement.

We have to show that $h\le 1$ a.s. Let
\[
  E_n := \{ h\ge 1+\tfrac1n\} \qqtext{and} E:=\{ h>1\}.
\]
Then $E_n\uparrow E$ and we have
\[
  Q'(E_n) \ge Q(E_n) = \int_{E_n} h\,dQ' \ge (1+\tfrac1n)\cdot Q'(E_n),
\]
which implies $Q'(E_n)=0$ for all $n$. Therefore $Q'(E)=\lim_n Q'(E_n)=0$.
%\item `$\Rightarrow$' is clear by the previous statement as we can set $\mu=Q'$ and
%  `$\Leftarrow$' follows from:
%\[
%  Q(B) = \int_B h\,d\mu \le \int_B h'\,d\mu = Q'(B) \qquad\forall B\in\clB.
%\qedhere
%\]
%\end{enumerate}
\end{proofof}

\begin{lemma}\label{lemma.monotone.convergence.1}\label{lemma.support}
\begin{enumerate}
\item Let $Q_n\uparrow P$, $A:=\supp P$ and $B := \bigcup_n \supp Q_n$.
Then $B\subset A$ and $P(B\setminus A)=0$.

\item Assume $Q$ is a finite measure and $Q_1\le Q_2\le\ldots\le Q$ and
let the densities $h_n:= \frac{dQ_n}{dQ}$.
Then $h_1\le h_2\le \ldots\le 1$ $Q$-a.s.
Furthermore, the following are equivalent:
\begin{enumerate}\renewcommand{\labelenumii}{(\roman{enumii})}
\item $Q_n\uparrow Q$
\item $h_n\uparrow 1$ $Q$-a.s.
\item $h_n\uparrow 1$ in $L^1$.
\end{enumerate}
\end{enumerate}
\end{lemma}

\begin{proofof}{Lemma \ref{lemma.monotone.convergence.1}}
\begin{enumerate}
\item Since $Q_n\le P$ we have $\supp Q_n\subset A$ and therefore $B\subset A$.
Because of $(A\setminus B) \cap \supp Q_n = \emptyset$ and the convergence we have for all $n$
\[
  P(A\setminus B) = \lim_\toi n Q_n(A\setminus B) = 0.
\]

\item By the previous lemma we have $h_1\le h_2\le \cdot\le 1$ $Q$-a.s.

(i) $\Rightarrow$ (ii): Since $(h_n)_n$ is monotone $Q$-a.s.
it converges $Q$-a.s.\ to a limit $h\le 1$. Let
\[
  E_n := \{ h\le 1-\tfrac1n\} \qqtext{and} E:=\{ h<1\}.
\]
Then $E_n\uparrow E$ and we have by the monotone convergence theorem:
\[
%  Q(E_n) = \int_{E_n} 1\,dQ
  Q_m(E_n) = \int_{E_n} h_m\,dQ \xrightarrow[\toi m]{} \int_{E_n} h\,dQ
  \le (1-\tfrac1n)\,Q(E_n)
\]
But since $Q_m(E_n)\uparrow_m Q(E_n)$ this means
that $Q(E_n)=0$ for all $n$ and therefore $Q(E)=\lim_n Q(E_n)=0$.

(ii) $\Rightarrow$ (iii): This follows from monotone convergence, because $1\in L^1(Q)$.

(iii) $\Rightarrow$ (i): For all $B\in\clB$:
\[
  Q(B) - Q_n(B) = \int_B |1-h_n|\,dQ \le \int |1-h_n|\,dQ \to 0
\]
because of $h_n\to 1$ in $L^1$.
\qedhere
\end{enumerate}
\end{proofof}

%\nocite{*}
%\printbibliography
%\small  %%% to save some lines
\bibliography{clustering}

\end{document}